\documentclass[journal]{IEEEtran}
\ifCLASSINFOpdf
\else
\fi
\usepackage{graphicx}
\usepackage{graphics}
\usepackage{float}
\graphicspath{{./figures/}}

\usepackage{booktabs}

\usepackage{tikz}
\usetikzlibrary{positioning,decorations.pathreplacing,arrows.meta}
\tikzstyle{EmptyBlock}=[rectangle, draw=blue!80,fill=blue!20, node distance=6em, text centered, minimum height=3em, minimum width=9em, line width=1.5pt, opacity=0]
\tikzstyle{BlueBlock}=[rectangle, draw=blue!80,fill=blue!20, node distance=6em, text centered, minimum height=3em, minimum width=9em, line width=1.5pt]
\tikzstyle{OrangeBlock}=[rectangle, draw=red!80,fill=red!20, node distance=12em,  align=center, text width=0.25*\columnwidth,minimum height=1cm, line width=1.5pt]
\tikzstyle{EndBlock}=[rectangle, draw=red!0,fill=red!0, node distance=12em,  align=left, text width=0.25*\columnwidth,minimum height=1cm]
\tikzstyle{Arrow} = [-{Latex[length=2mm,width=2mm]}, line width=0.4mm]
\tikzstyle{ArrowText} = [sloped, anchor=center,above]

\usepackage{amsmath}
\usepackage{amssymb}  % assumes amsmath package installed
\newcommand{\argmin}{\mathop{\mathrm{argmin}}}
\newcommand{\argmax}{\mathop{\mathrm{argmax}}}

\usepackage{subcaption}
\usepackage{multirow}

\usepackage{bm}
\usepackage{algorithm}
\usepackage{algorithmicx}
\usepackage{algpseudocode}

\usepackage[font={small}]{caption}

\usepackage{textcomp}
\usepackage{lipsum}
\usepackage{siunitx}
\usepackage{soul}

\begin{document}
\title{Virtual Maps for Autonomous Exploration of Cluttered Underwater Environments}
%


\author{Jinkun~Wang,
        Fanfei Chen,
        Yewei Huang, 
        John McConnell,
        Tixiao~Shan,
       
       %Timothy Osedach,~\IEEEmembership{Member,~IEEE,}
       %Stephane Vannuffelen,
        and~Brendan~Englot,~\IEEEmembership{Senior Member,~IEEE}% <-this % stops a space
\vspace{-3mm}
\thanks{This work was supported by a grant from Schlumberger Technology Corporation, and by the National Science Foundation, Grants IIS-1652064 and IIS-1723996. \textit{(Corresponding author: Brendan Englot.)}

Y. Huang, J. McConnell, and B. Englot are with the Department of Mechanical Engineering, Stevens Institute of Technology, Hoboken, NJ, USA (e-mail: \{yhuang85, jmcconn1, benglot\}@stevens.edu). %*Asterisk indicates that these authors contributed equally to this work. 

J. Wang was previously with Stevens Institute of Technology and is now with Amazon Robotics, Louisville, CO, USA. (email: me.jkunw@gmail.com).

F. Chen was previously with Stevens Institute of Technology and is now with Exyn Technologies, Philadelphia, PA, USA. (email: frankycff@gmail.com).

T. Shan was previously with Stevens Institute of Technology and is now with SRI International, Princeton, NJ, USA. (email: tixiao.shan@sri.com).

%T. Osedach and S. Vannuffelen are with Schlumberger-Doll Research Center, Cambridge, MA, USA (e-mail: \{tosedach, svannuffelen\}@slb.com).

}
}

\maketitle
% \copyrightnotice

\begin{abstract}
We consider the problem of autonomous mobile
robot exploration in an unknown environment, taking into
account a robot’s coverage rate, map uncertainty, and state
estimation uncertainty. This paper presents a novel exploration framework for underwater robots operating in cluttered environments, built upon simultaneous localization and mapping (SLAM) with imaging sonar. The proposed system comprises path generation, place recognition forecasting, belief propagation and utility evaluation using a \textit{virtual map}, which estimates the uncertainty associated with map cells throughout a robot's workspace. We evaluate the performance of this framework in simulated experiments, showing that our algorithm maintains a high coverage rate during exploration while also maintaining low mapping and localization error. The real-world applicability of our framework is also demonstrated on an underwater remotely operated vehicle (ROV) exploring a harbor environment.

%We build on our previous work on expectation maximization (EM) exploration, which explicitly models unknown landmarks as latent variables and predicts their expected uncertainty, to resolve the lack of landmark state in denser instances of SLAM.

\end{abstract}


% \fi
\IEEEpeerreviewmaketitle

%\section{Introduction}
%\IEEEPARstart{T}{his} demo file is intended to serve as a ``starter file''
%for IEEE journal papers produced under \LaTeX\ using
%IEEEtran.cls version 1.8b and later.
%% You must have at least 2 lines in the paragraph with the drop letter
%% (should never be an issue)
%I wish you the best of success.

%%%%%%%%%%%%%%%%%%%%%%%%%%%%%%%%%%%%%%%%%%%%%%%%%%%%%%%%%%%%%%%%%%%%%%%%%%%%%%%%
\section{Introduction}
\label{sec:introduction}

\IEEEPARstart{S}{imultaneous} localization and mapping (SLAM) has been well studied in theory and applied successfully on real sensing platforms for state estimation and map-building using
data collected passively \cite{Cadena2016}. However, it’s still a challenge for an autonomous vehicle to actively map an unknown environment, properly managing the trade-off between exploration speed and state estimation quality. Improving the capability of autonomous exploration is beneficial for many robot mapping tasks, especially in scenarios where teleoperation is limited or infeasible due to constrained communication, e.g., in unknown subsea environments with underwater robots.

The autonomous exploration problem is generally solved
in three stages: path generation, utility evaluation and execution. First, we identify candidate waypoints or generate a sequence of actions to follow, which is typically achieved by enumerating frontiers or by employing sampling-based path
planning methods. The selected path is usually straightforward to execute using feedback controllers, thus leaving us
with a fundamental problem of designing a utility function
to measure path optimality. Essentially, it should capture the
exploration-exploitation dilemma, i.e., a balancing of visiting unknown areas to reduce map uncertainty, and revisiting known areas to seek better localization and map accuracy.

%A receding horizon next-best-view planner achieved real-time 3D exploration on a micro aerial vehicle \cite{Bircher2016}, and the selected path is optimized to minimize the expected localization and mapping uncertainty \cite{Papachristos2017}. %-> Introduce these references later

Predicting the impact of future actions on system uncertainty remains an open problem \cite{Cadena2016}, and this is particularly true for unobserved landmarks or other environmental features that may provide useful relative measurements. Our previous work on expectation-maximization (EM) exploration \cite{Wang2017} introduced the concept of a \textit{virtual map} composed of virtual landmarks acting as proxy for a real feature-based map, on which we are able to predict the uncertainty resulting from future sensing actions. Since every virtual landmark is deeply connected with robot poses that can observe it, the metrics for exploration and localization are unified as the determinant of the virtual landmarks' error covariance matrix. 
In follow-on work \cite{Wang2019exp}, we considered a more realistic dense navigation scenario, in which exploration with lidar-equipped ground robots must rely on pose SLAM that does not incorporate enumerated features explicity.
%the exploration problem is limited to rely on pose SLAM that does not incorporate features explicitly, such as LeGO-LOAM proposed in \cite{Shan2018}. 
In that scenario, SegMap \cite{Dube2017, Dube2017-2} provided a convenient utility for place recognition. Map segments served as landmarks, and loop-closures from segment matching allowed us to perform active localization efficiently. 

In this present paper, we extend the EM exploration algorithms employed in past works \cite{Wang2017,Wang2019exp} to the real-world application that originally inspired this work - the exploration of cluttered underwater environments by a robot equipped with a multibeam imaging sonar. In doing so, this paper provides the following contributions:
%We improved our previous work on EM exploration to accommodate dense observations using segment-aided LiDAR mapping.
\begin{itemize} 
\item A substantially more detailed exposition of processes for belief propagation (over candidate robot actions, and over virtual landmarks) that support the efficient implementation of EM exploration in real-time, at large scales,  (these topics are discussed only briefly in \cite{Wang2019exp});
\item An architecture for robust underwater SLAM and occupancy mapping with imaging sonar capable of supporting repeated real-time planning and decision making over their evolving estimates in cluttered environments;
\item A thorough evaluation of our framework's performance in simulated environments, rigorously evaluating the tradeoffs between a robot's rate of exploration, localization and map accuracy in comparison with other algorithms;
\item The first instance, to our knowledge, of %fully 
autonomous exploration of a real-world, obstacle-filled outdoor environment, in which an underwater robot (a tethered BlueROV we employ as a proof-of-concept system) directly incorporates its SLAM process and predictions based on that process into its decision-making.
%\item An open source implementation of our proposed framework for real-time segment-aided mapping and exploration, which will be released upon publication\footnote{\url{https://github.com/jinkunw}}.
\end{itemize}

%The rest of the paper is organized as follows. We review relevant prior work in Sec. II, 

\section{Related Work}
\label{sec:related-work}

The various factors considered in our mobile robot exploration problem, localization uncertainty, map uncertainty, and coverage rate, have been considered individually and in combination across a large body of prior work. To manage map uncertainty, planning with a priori maps has been leveraged to actively minimize the uncertainty of known landmarks \cite{Feder1999, Cantin2007, Sim2005}. Similarly, planning in unknown environments but with predefined waypoints is investigated in \cite{Indelman2015}. 
If we do not consider the uncertainty of a robot's state and map, the problem has been approached by following the nearest frontier \cite{Yamauchi1997}, choosing sensing actions to maximize mutual information \cite{Julian2014, Bai2016}, 
using Cauchy-Schwarz quadratic mutual information (CSQMI) to reduce computation time \cite{Charrow2015-1}, %combining global planning with local motion primitives and also refining a trajectory using optimization methods \cite{Charrow2015-2}, 
and by exploring on continuous Gaussian process frontier maps \cite{Jadidi2018}.
Exploring efficiently over varying distances has been achieved by providing a robot with greedy and non-greedy options \cite{Charrow2015-2}, as well as planning over a receding horizon in which robots execute only a small portion of each plan prior to replanning \cite{Bircher2016}.
Most of the existing research on exploration in unknown environments takes advantage of occupancy grid maps, considering utility functions involving map entropy and robot pose uncertainty \cite{Bourgault2002, Makarenko2002}. This paradigm has succeeded in complex applications, including real-time 3D exploration and structure mapping with micro aerial vehicles \cite{Papachristos2017}. However, inaccurate state estimation is likely to result in a complete yet distorted map regardless of exploration speed. The correlation between localization and information gain is often taken into account by integrating over a map's entropy weighted by pose uncertainty \cite{Stachniss2005, Valencia2012, Vallve2015}. However, in prior work we have shown that weighted entropy fails to capture the estimation error of landmarks existing in the map, and map inaccuracies may result \cite{Wang2017}. 

A body of relatively recent work has considered autonomous exploration and active perception with underwater robots in particular. This work has emphasized mapping with sonar, which can perceive obstacles at relatively long ranges in all types of water. Vidal et al. \cite{Vidal2017} first employed sonar-based occupancy mapping, in 2D, to explore cluttered underwater environments with an autonomous underwater vehicle (AUV), which repeatedly selects nearby viewpoints at the map's frontiers. Subsequent work by Palomeras et al. \cite{Palomeras2019} extended the approach to 3D sonar-based mapping and exploration of cluttered underwater environments using a next-best-view framework that weighs travel distance, frontiers, and contour-following when selecting a new viewpoint. In addition to view planning, active SLAM has been applied to underwater robots, used to incorporate planned detours into visual inspections that achieve loop closures \cite{Chaves2014, Chaves2015, Chaves2016}, and to deviate from three-dimensional sonar-based next-best-view planning to achieve a loop closure when localization uncertainty exceeds a designated threshold \cite{Suresh2020}. Of these two works, the former navigates at a fixed standoff from a ship hull's surface, and the latter explores and maps an indoor tank environment.

Finally, it is also worth noting that autonomous mobile robot exploration is related to a more broadly-focused body of works that study belief space planning (BSP) in unknown environments. Such works have succeeded in relaxing assumptions often imposed on exploration, such as that of maximum likelihood observations \cite{Indelman2015}, and in achieving more efficient belief propagation and covariance recovery \cite{Indelman2019}. However, such techniques have thus far been implemented in simple, obstacle-free domains populated with point landmarks.
The belief space planning problem, in turn, falls under the larger umbrella of partially observable Markov decision processes (POMDPs), and recent work has succeeded in finding efficient ways to solve POMDPs that could potentially be applied to autonomous mobile robot exploration \cite{Silver2010,Hsu2013}.

In the sections that follow, we will describe our proposed EM exploration framework and its application to underwater mapping with sonar, demonstrating that it achieves far lower localization and mapping errors than frontier-based \cite{Yamauchi1997} and next-best-view \cite{Bircher2016} exploration algorithms, and that the localization and mapping errors realized are comparable to that of \cite{Suresh2020}, while achieving a superior rate of exploration.

%\section{Background}
%\label{sec:background}

\vspace{-2mm}

\section{EM Exploration}
\label{sect:em}

We address the problem of autonomous exploration for a range-sensing mobile robot in an initially unknown environment. Our robot performs SLAM and constructs an occupancy grid map as it explores. We assume a bounded 2D workspace $V \subset \mathbb R^2$ where all discretized cells $\mathbf m_i$ are initialized as \textit{unknown} $P(m_i=1)=0.5$. A \textit{frontier} is defined as the boundary where free space meets unmapped space. The exploration is considered complete if no frontier can be detected. However, highly uncertain poses are likely to result in complete, yet inaccurate occupancy grid maps, limiting the usefulness of information gained by exploring unknown space. Assuming the environment contains individual landmarks $L = \{\mathbf l_k\ \in \mathbb R^2\}$, apart from discovering more landmarks, minimizing the estimation error is equally crucial.

%\section{EM Exploration}
\vspace{-3mm}

\subsection{Simultaneous Localization and Mapping}
\label{subsec:slam}

We use a \textit{smoothing}-based approach rather than a \textit{filtering}-based approach, adopting incremental smoothing and mapping \cite{Kaess2012} to repeatedly estimate the entire robot trajectory. %While the smoothing 
This affords us the flexibility to adopt either landmark-based SLAM, or pose SLAM, scalably over long-duration missions. Importantly, it also allows corrections to be made throughout the mission to both the robot's estimated pose history, and to the map of the environment resulting from that pose history.
%The benefit of this approach will be discussed in Sections \ref{sec:actions} and \ref{sec:landmarks}.

Let a mobile robot's motion model be defined as
\begin{equation}
	\mathbf x_i = f_i(\mathbf x_{i-1}, \mathbf u_i) + \mathbf w_i,\quad \mathbf w_i \sim \mathcal N(\mathbf 0, Q_i),
\end{equation}
where $\mathbf x_i =[x_i, y_i, \theta_i] \in SE(2)$ denotes the vehicle pose at the $i$-th time-step, $\mathbf u_i \in \mathbb R^n$ denotes the control input to the vehicle, and $\mathbf w_i \sim \mathcal N(\mathbf 0, Q_i)$ is zero-mean additive Gaussian process noise with covariance matrix $Q_i$. Let the measurement model be defined as
\begin{equation}
	\mathbf z_{ij} = h_{ij}(\mathbf x_i, \mathbf l_j) + \mathbf v_{ij},\quad \mathbf v_{ij} \sim \mathcal N(\mathbf 0, R_{ij}),
	\label{eq:meas_model}
\end{equation}
where we obtain measurement between landmark $\mathbf l_k$ and pose $\mathbf x_i$ corrupted by zero-mean additive Gaussian sensor noise $\mathbf v_{ij}$ with covariance matrix $R_{ij}$. We assume the data association between $\mathbf x_i, \mathbf l_j$ is known. It is worth noting that the above definitions can be extended to 3D.

Given measurements $\mathcal Z = \{\mathbf z_k\}$, we seek to obtain the best estimate of the entire trajectory $\mathcal X = \{\mathbf x_i\}$ and observed landmarks $\mathcal L = \{\mathbf l_j\}$,
\begin{equation}
\begin{aligned}
	\mathcal X^*, \mathcal L^* | \mathcal Z &= \argmax_{\mathcal X, \mathcal L} P(\mathcal X, \mathcal L, \mathcal Z)\\
	                                        &= \argmax_{\mathcal X, \mathcal L} P(\mathcal Z| \mathcal X, \mathcal L)P(\mathcal X),
\end{aligned}
\label{eq:slam_prob}
\end{equation}
where $P$ is the likelihood of the measurements or control inputs given vehicle poses and landmark positions. The maximum a posteriori (MAP) estimate can be used by maximizing the joint probability, which afterwards leads to a nonlinear least-squares problem. By constructing a graph representation and linearizing nonlinear functions, the marginal distributions and joint marginal distributions, both of which are Gaussian, can be extracted using graphical model-based inference algorithms \cite{Kaess2012}.\

\subsection{EM-Exploration}

In the formulation of the SLAM problem as a \textit{belief net} \cite{Dellaert2006}, the solution is obtained by maximizing the joint probability distribution as in Equation (\ref{eq:slam_prob}). During exploration, we are confronted with unknown landmarks that have not been observed yet. Therefore, we introduce the concept of \textit{virtual landmarks} $\mathcal V$ as latent variables, which describe potential landmark positions that would be observed when following the planned path. Then the objective is to maximize the following marginal model,
\begin{equation}
\begin{aligned}
\mathcal X^* &= \argmax_{\mathcal X}\ \log P(\mathcal X, \mathcal Z) \\&= \argmax_{\mathcal X}\ \log \sum_{\mathcal V} P(\mathcal X, \mathcal Z, \mathcal V).
\end{aligned}
\end{equation}
\begin{figure}
	\centering
	\begin{tikzpicture}
	\draw[thick,->,rounded corners=0.2cm] (0,0) -- (4,0) -- (4, 2) -- (2.5, 2);
	\draw[thick,-] (1,0) -- (1,-0.2) node[anchor=north] {$\mathbf x_i, \Sigma_{x_i}$};
	\draw[thick,-] (3,0) -- (3,-0.2) node[anchor=north] {$\mathbf x_j,\Sigma_{x_j}$};
	\draw[thick,-] (3,2) -- (3,2.2) node[anchor=south] {$\mathbf x_k,\Sigma_{x_k}$};
	\draw[-] (3,2) -- (3,0);
	\end{tikzpicture}\
	\begin{tikzpicture}
	\draw[thick,->,rounded corners=0.2cm] (0,0) -- (4,0) -- (4, 2) -- (2.5, 2);
	\draw[thick,-] (1,0) -- (1,-0.2) node[anchor=north] {$\mathbf x_i,\Sigma_{\mathbf x_i}$};
	\draw[thick,-] (3,0) -- (3,-0.2) node[anchor=north] {$\mathbf x_j,\Sigma_{\mathbf x_j}$};
	\draw[thick,-] (3,2) -- (3,2.2) node[anchor=south] {$\mathbf x_k,\Sigma_{\mathbf x_k}$};
	\draw[-] (3,2) -- (3,0);
	\draw[thick,->,red] (2.5,2) -- (0,2);
	\draw[thick,->,red,opacity=0.3] (2.5,2) to [out=180,in=-30] (0,2.5);
	\draw[thick,->,red,opacity=0.3] (2.5,2) to [out=180,in=30] (0,1.5);
	\end{tikzpicture}\hspace{1em}\\
	\begin{tikzpicture}
	\draw[thick,->,rounded corners=0.2cm] (0,0) -- (4,0) -- (4, 2) -- (2.5, 2);
	\draw[thick,-] (1,0) -- (1,-0.2) node[anchor=north] {$\mathbf x_i,\color{red}{\Sigma_{\mathbf x_i}}$};
	\draw[thick,-] (3,0) -- (3,-0.2) node[anchor=north] {$\mathbf x_j,\color{red}{\Sigma_{\mathbf x_j}}$};
	\draw[thick,-] (3,2) -- (3,2.2) node[anchor=south] {$\mathbf x_k,\color{red}{\Sigma_{\mathbf x_k}}$};
	\draw[thick,-,red] (1,2) -- (1,2.2) node[anchor=south] {$\mathbf x_{k+n},\Sigma_{\mathbf x_{k+n}}$};
	\draw[-] (3,2) -- (3,0);
	\draw[thick,->,red] (2.5,2) -- (0,2);
	\draw[-,red] (1,2) -- (1,0);
	\end{tikzpicture}\
	\begin{tikzpicture}
	\draw[thick,->,rounded corners=0.2cm,opacity=0.3] (0,0) -- (4,0) -- (4, 2) -- (2.5, 2);
	\draw[thick,-] (1,0) -- (1,-0.2) node[anchor=north] {$\mathbf x_i,\color{red}{\Sigma_{\mathbf x_i}}$};
	\draw[thick,-,opacity=0.3] (3,0) -- (3,-0.2) node[anchor=north] {$\mathbf x_j,\color{red}{\Sigma_{\mathbf x_j}}$};
	\draw[thick,-,opacity=0.3] (3,2) -- (3,2.2) node[anchor=south] {$\mathbf x_k,\color{red}{\Sigma_{\mathbf x_k}}$};
	\draw[thick,-,red] (1,2) -- (1,2.2) node[anchor=south] {$\mathbf x_{k+n},\Sigma_{\mathbf x_{k+n}}$};
	\draw[-,opacity=0.3] (3,2) -- (3,0);
	\draw[thick] (0.5,1) circle (0.05) node[anchor=east] {$\mathbf v$}; 
	\draw[thick,->,red,opacity=0.3] (2.5,2) -- (0,2);
	\draw[-,red,opacity=0.3] (1,2) -- (1,0);
	\draw[-] (1,0) -- (0.5, 1) node[midway, anchor=east] {$\Sigma_{\mathbf v}^{(i)}$};
	\draw[-] (1,2) -- (0.5, 1) node[midway, anchor=east] {$\Sigma_{\mathbf v}^{(k+n)}$};
	\end{tikzpicture}
	\caption[Belief propagation of candidate actions and virtual landmarks.]{An illustration of belief propagation of candidate actions and virtual landmarks in EM-exploration.}
	\label{fig:bruce-belief-propagation}
\end{figure}
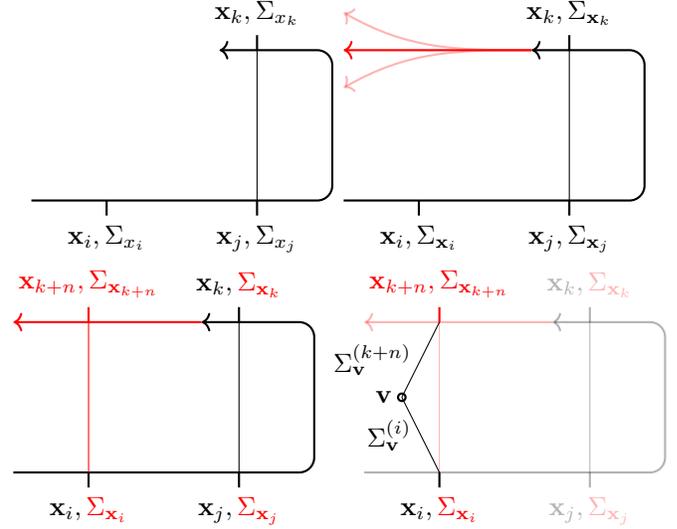

The above equation involves unobserved variables, which can be approached intuitively using an expectation-maximization (EM) algorithm as follows,
\begin{align}
\text{E-step:\quad} & q(\mathcal V) = p(\mathcal V|\mathcal X^{\text{old}}, \mathcal Z)\\
\text{M-step:\quad} & \mathcal X^{\text{new}} = \arg\max_{\mathcal X} \mathbb E_{q(\mathcal V)}[\log P(\mathcal X, \mathcal V, \mathcal Z)].
\end{align}
In the E-step, latent virtual landmarks $\mathcal V$ are computed based on the current estimate of the trajectory $\mathcal X_{\text{old}}$ and the history of measurements $\mathcal Z$. In the M-step, a new trajectory $\mathcal X^{\text{new}}$ is selected such that the expected value of joint probability, given the virtual landmark distributions, is maximized. The iterative algorithm alternates between the E-step and M-step, but each iteration is accomplished by the execution of control actions and the collection of measurements.

The equation above poses a challenge for efficient solution due to the exponential growth of potential virtual landmark configurations with respect to the number of virtual landmarks. Inspired by classification EM algorithms, an alternative solution would
add a classification step (C-step) before the M-step to provide the maximum posterior probability estimate of the virtual landmark distributions,
\begin{align}
\text{C-step:\quad} & \mathcal V^* = \argmax_{\mathcal V}\ p(\mathcal V|\mathcal X^{\text{old}}, \mathcal Z)\\
\text{M-step:\quad} & \mathcal X^{\text{new}} = \arg\max_{\mathcal X}\ \log P(\mathcal X, \mathcal V^*, \mathcal Z).\label{eq:m}
\end{align}
If we further assume measurements are assigned to maximize the likelihood 
\begin{equation*}
\mathcal Z = \argmax_{\mathcal Z} h(\mathcal X, \mathcal V), 
\end{equation*}
where $h$ is the measurement model in Equation (\ref{eq:meas_model}), then the joint distribution can be expressed as a multivariate Gaussian centered at the proposed poses and landmark positions, and the covariance can be approximated by the information matrix inverse,
\begin{equation}
P(\mathcal X,\mathcal  V, \mathcal Z) \sim \mathcal N\bigg(\begin{bmatrix}\mathcal X \\ \mathcal V\end{bmatrix}, \begin{bmatrix} \Sigma_{\mathcal X\mathcal X} & \Sigma_{\mathcal X\mathcal V} \\ \Sigma_{\mathcal V\mathcal X} & \Sigma_{\mathcal V\mathcal V}  \end{bmatrix}\bigg),
\end{equation}
where $\Sigma_{\mathcal X \mathcal X}$ and $\Sigma_{\mathcal V \mathcal V}$ are the marginal covariances of poses and landmarks respectively. The solution of Eq. (\ref{eq:m}) is equivalent to evaluating the log-determinant of the covariance matrix,
\begin{equation}
\argmax_\mathcal X\ \log P(\mathcal X, \mathcal V^*, \mathcal Z) = \argmin_\mathcal X\ \log \text{det}(\Sigma).
\end{equation}
This implies that the performance metric for our proposed exploration is consistent with the D-optimality criterion in active SLAM \cite{Carrillo2012}, except that the subjects considered include unobserved landmarks.

\subsection{Belief Propagation on Candidate Actions}
\label{sec:actions}

Since we are more interested in the uncertainty of the virtual landmarks and the most recent pose $\mathbf x_{T+N}$ at step $T$ with planning horizon $N$, we can marginalize out irrelevant poses in $\Sigma_{\mathcal X\mathcal X}$, ending up with $\Sigma_{\mathbf x_{T+N}}$. Typically, there exist thousands of virtual landmarks, thus approximation of $\Sigma_{\mathcal V\mathcal V}$ is critical for real-time applications. Combined with pose simplification, we can obtain that, for a positive definite covariance matrix,
\begin{equation}
\log \text{det}(\Sigma) < \log \text{det}(\Sigma_{\mathbf x_{T+N}}) + \sum_k \log \text{det}(\Sigma_{\mathbf v_k}),
\label{eq:approx1}
\end{equation}
where $\Sigma_{\mathcal V_k}$ is the diagonal block involving the $k$th virtual landmark in $\Sigma_{\mathcal V\mathcal V}$. This approximation is reasonable considering an overestimate of information using the introduced virtual landmarks.

We will discuss the process for estimating $\Sigma_{\mathbf x_{T+N}}$ in this section and $\Sigma_{\mathbf v_k}$ in the next section. This process is illustrated in Figure~\ref{fig:bruce-belief-propagation}. At the beginning, the existing pose estimate and covariance (top left) can be obtained from SLAM, and candidate actions (top right) are generated from a path library, which will be discussed later. First, given one candidate path and predicted measurements, the covariance of poses along the entire trajectory is updated, i.e., $\Sigma_{\mathbf x_{1:T+N}}$ (bottom left). Second, provided the predicted pose uncertainty, we approximate the covariance of virtual landmarks independently without the full pose covariance matrix (bottom right).

The SLAM problem defined in Section \ref{subsec:slam} can be approximated by performing linearization and solved by iteration through the linear form given by
\begin{equation}
\boldsymbol \delta^* = 	\argmin_{\boldsymbol{\delta}} \frac 12 \Vert A\boldsymbol{\delta} - \mathbf b\Vert^2,
\end{equation} 
where $A$ represents the Jacobian matrix and the vector $b$ represents measurement residuals. The incremental update $\boldsymbol{\delta}$ in the above equation is obtained by solving the linear system,
\begin{equation}
A^\top A \boldsymbol \delta = A^\top \mathbf b.
\end{equation}
It can also be formulated as $\Lambda \boldsymbol{\delta} = \boldsymbol \eta$ where $\Lambda = A^\top A$ is known as the information matrix. In general, the covariance matrix may be obtained by inverting the information matrix
\begin{equation}
\Sigma = \Lambda^{-1} = (A^\top A)^{-1}.
\end{equation}
As shown in \cite{kaess2009covariance}, the recovery of block-diagonal entries corresponding to pose covariances can be implemented efficiently. As a result, we concern ourselves with the problem of updating uncertainty upon the arrival new measurements assuming that $\Sigma_{\mathbf x_i}$ are given. Figure~\ref{fig:bruce-belief-propagation} shows a scenario (from top right to lower left) in which the robot's belief about its current trajectory has been obtained (black) and updates (red) are to be investigated if new measurements are available.

\begin{equation}
\begin{bmatrix}
\color{blue}{\Sigma_{\mathbf x_0}} & & \color{green}{\Sigma_{\mathbf x_{0i}}} & & & & \color{green}{\Sigma_{\mathbf x_{0k}}} & \color{red}{\cdots} & \color{red}{\Sigma_{\mathbf x_{0,k+n}}}\\
& \color{blue}{\ddots} & \color{green}{\vdots} & & & &\color{green}{\vdots} & \color{red}{\dots} & \color{red}{\vdots} \\
& & \color{blue}{\Sigma_{\mathbf x_i}}  & & & &\color{green}{\Sigma_{\mathbf x_{ik}}} & \color{red}{\dots} & \color{red}{\Sigma_{\mathbf x_{i,k+n}}}\\
&  & & \color{blue}{\ddots }& & & \color{green}{\vdots} & \color{red}{\dots} & \color{red}{\vdots} \\
&  & & & \color{blue}{\Sigma_{\mathbf x_j}} & &\color{green}{\Sigma_{\mathbf x_{jk}}} & \color{red}{\dots} & \color{red}{\Sigma_{\mathbf x_{j,k+n}}}\\
&  & & & & \color{blue}{\ddots} & \color{green}{\vdots} & \color{red}{\dots} & \color{red}{\vdots} \\
&  & & & & & \color{blue}{\Sigma_{\mathbf x_k}} & \color{red}{\dots} & \color{red}{\Sigma_{\mathbf x_{k,k+n}}}\\
&  & & & & & & \color{red}{\ddots} & \color{red}{\vdots}\\
&  & & & & & &  & \color{red}{\Sigma_{\mathbf x_{k + n}}} \\
\end{bmatrix}
\label{eq:bruce-cov-matrix}
\end{equation}

Covariance recovery is performed in three steps. We use Figure~\ref{fig:bruce-belief-propagation} and Equation~(\ref{eq:bruce-cov-matrix})  as illustrative examples, where the index $k$ denotes the most recent pose. First, we compute diagonal entries in the covariance matrix (blue). Second, we ignore loop closure measurements and propagate the covariance using only odometry measurements, which could be derived from wheel odometry, inertial odometry or sequential scan matching. This open-loop covariance recovery is given by
\begin{equation}
\Sigma_{\mathbf x_{i,k+t}} = \Sigma_{\mathbf x_{i, k+t-1}} F_{k+t}^\top,
\label{eq:bruce-block-diag}
\end{equation}
where $ F_{k+t} = \frac{\partial \mathbf f_{k+t}}{\partial \mathbf x_{k+t}}$ is the Jacobian matrix of $\mathbf f_{k+t}$ with respect to pose $\mathbf x_{\mathbf x_{i, k-1}}$. The equation is applied recursively and the initial value $\Sigma_{\mathbf x_{i,k}}$ can be computed by 
\begin{equation}
\Lambda \Sigma_{\mathbf x_{\cdot,k}} = I_{k},\quad R \Sigma_{\mathbf x_{\cdot,k}} = R^\top I_{k}
\label{eq:bruce-block-col},
\end{equation}
where $\Sigma_{\mathbf x_{\cdot,k}}$ represents the cross-covariance between past poses and the current pose (green column $k$) and $I_{k}$ is a sparse block column matrix with an identity block only at the position corresponding to pose $k$. The solution of Equation~(\ref{eq:bruce-block-col}) is obtained by Cholesky decomposition on the right (the $R$ is available immediately after incremental update in iSAM2 \cite{Kaess2012}), whose computational complexity, for sparse $R$ with $N_{nz}$ nonzero elements, is $O(N_{nz}$).

Finally, we use the Woodbury formula to update the covariance matrix \cite{Ila2015},
\begin{equation}
\Sigma' = \Sigma + \Delta \Sigma, \Delta \Sigma = -\Sigma A_u^\top(I + A_u\Sigma A_u^\top)^{-1}A_u\Sigma,
\label{eq:bruce-cov-update}
\end{equation}
where $A_u$ is a Jacobian matrix with each block row corresponding to one loop closure measurement. Using the above formula results in a highly efficient update, as it avoids the inversion of a large dense matrix $(A^\top A + A_u^\top A_u)^{-1}$. In total, our belief propagation process requires the recovery of block columns for which the respective pose appears in the measurements (e.g., green column $i$ of Eq. (\ref{eq:bruce-cov-matrix})), and a matrix inversion of a relatively small matrix with the number of block rows equal to the number of loop closure measurements.

% \begin{figure}
% 	\centering
% 	\includegraphics[width=0.9\columnwidth]{main/bp_action}
% 	\caption[Efficient calculation of pose covariance on candidate actions.]{Efficient calculation of pose covariance on candidate actions.}
% 	\label{fig:bruce-woodbury}
% \end{figure}

\subsection{Belief Propagation on Virtual Landmarks}
\label{sec:landmarks}
% Let $\Sigma_{\mathbf v_k}$ be the actual covariance matrix, and let $\Sigma_{\mathbf v_k}^{(i)}, \Sigma_{\mathbf v_k}^{(j)}$ be the covariance estimates from two individual measurements at poses $\mathbf x_i$ and $\mathbf x_j$ respectively.

\begin{figure*}[ht]
	\centering
	\includegraphics[width=0.9\textwidth]{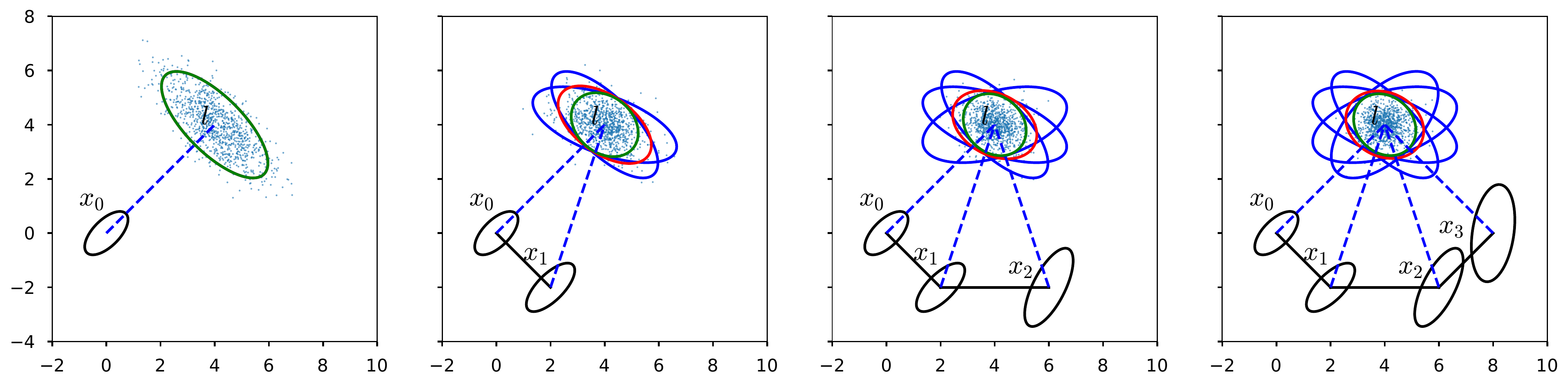}
	\caption[Covariance intersection.]{Covariance intersection across several robot poses is used to compute an upper bound (red ellipses) on the actual landmark covariance (green ellipses). Raw range-bearing returns from the object represented by landmark $l$, matched with their nearest neighbors across consecutive robot poses using ICP, are indicated as blue dots.}
	\label{fig:bruce-sci}
\end{figure*}

Let us assume the robot following a certain path is able to take measurements from landmarks in the surrounding environment. Let $\{\mathbf x_i \in \mathbf{SE}(2)\}$ be the robot poses that observe the same landmark $\mathbf l \in \mathbb R^2$. In the following derivation we do not distinguish between virtual landmark $\mathbf v$ and actual landmark $\mathbf l$ (the process will accommodate both). The measurement $\mathbf z_i$ can be obtained from the following sensor model
\begin{equation}
\mathbf z_i = \tilde {\mathbf z}_i + \mathbf v_i = h_i(\mathbf x_i, \mathbf l) + \mathbf v_i,\ \mathbf v_i \sim \mathcal N(\mathbf 0, \mathbf R_i).
\label{eq:bruce-virtual-meaurement-model}
\end{equation}
We further assume that the sensor model is invertible, i.e., that we are able to predict landmark positions given robot poses and their corresponding measurements. Mathematically, the Jacobian matrix has full rank, $\text{rank}(\frac{\partial h_i}{\partial \mathbf l}) = 2$. One example of such a model is bearing-range measurements, which are commonly produced by laser range-finders. In our experiments, we extend this assumption to multibeam sonars. We neglect their horizontal beamwidth due to their high resolution and narrow beams, and we neglect vertical beamwidth because our vehicle operates at constant depth in an environment whose man-made structures possess a fixed cross-sectional profile. In the following, we will use the inverse sensor model for convenience,
\begin{equation}
\mathbf l = h^{-1}_i(\mathbf x_i, \tilde {\mathbf z}_i) = h^{-1}_i(\mathbf x_i, \mathbf z_i - \mathbf v_i),\ \mathbf v_i \sim \mathcal N(\mathbf 0, \mathbf R_i).
\end{equation}
The Jacobian matrices of the inverse sensor model are represented as $\mathbf H = \frac{\partial h^{-1}}{\partial \mathbf x}, \mathbf G = \frac{\partial h^{-1}}{\partial \tilde{\mathbf z}}$. Given the covariance matrices of poses $\{{\mathbf \Sigma}_{i} | {\mathbf \Sigma}_{i} \succ 0\}$, we intend to provide a consistent estimate of a landmark's covariance without the computation of $\{{\mathbf \Sigma}_{ij}| i \ne j\}$, as demonstrated in the bottom right of Figure~\ref{fig:bruce-belief-propagation}.

%\begin{equation}
%\begin{bmatrix}
%{\mathbf \Sigma}_{i} & {\mathbf \Sigma}_{ij} \\
%{\mathbf \Sigma}_{ij}^\top & {\mathbf \Sigma}_{j} \\
%\end{bmatrix}
%\end{equation}

Suppose we obtain two measurements of the same landmark at two distinct poses $\mathbf x_i, \mathbf x_j$, resulting in a joint distribution
\begin{equation}
\begin{bmatrix}
{\mathbf l}_i \\ {\mathbf l}_j
\end{bmatrix}
=
\begin{bmatrix}
h^{-1}({\mathbf x}_i, \mathbf z_i)\\
h^{-1}({\mathbf x}_j, \mathbf z_j)
\end{bmatrix}.
\end{equation}

\begin{align*}
\text{cov}\big(\begin{bmatrix}
{\mathbf l}_i \\ {\mathbf l}_j
\end{bmatrix}\big) &= 
\begin{bmatrix}
{\mathbf \Sigma}^{\mathbf l}_i & {\mathbf \Sigma}^{\mathbf l}_{ij} \\
{\mathbf \Sigma}^{\mathbf l\top}_{ij} & {\mathbf \Sigma}^{\mathbf l}_{j}
\end{bmatrix}\\
&=
\begin{bmatrix}
\mathbf H_i & \mathbf 0 \\
\mathbf 0 & \mathbf H_j
\end{bmatrix}
\begin{bmatrix}
{\mathbf \Sigma}_i & {\mathbf \Sigma}_{ij} \\
{\mathbf \Sigma}_{ij}^\top & {\mathbf \Sigma}_{j} \\
\end{bmatrix}
\begin{bmatrix}
\mathbf H_i & \mathbf 0 \\
\mathbf 0 & \mathbf H_j
\end{bmatrix}^\top\\& +
\begin{bmatrix}
\mathbf G_i & \mathbf 0 \\
\mathbf 0 & \mathbf G_j
\end{bmatrix}
\begin{bmatrix}
\mathbf R_i & \mathbf 0 \\
\mathbf 0 & \mathbf R_j
\end{bmatrix}
\begin{bmatrix}
\mathbf G_i & \mathbf 0 \\
\mathbf 0 & \mathbf G_j
\end{bmatrix}^\top
\end{align*}
Let $\mathbf P_{i1} = \mathbf H_i {\mathbf \Sigma}_i \mathbf H_i^\top \succ \mathbf 0, \mathbf P_{i2} = \mathbf G_i \mathbf R_i \mathbf G_i^\top  \succ \mathbf 0, \mathbf P_{ij} = \mathbf H_i {\mathbf \Sigma}_{ij} \mathbf H_j^\top$.
\begin{align*}
\text{cov}\big(\begin{bmatrix}
{\mathbf l}_i \\ {\mathbf l}_j
\end{bmatrix}\big) = 
\begin{bmatrix}
{\mathbf \Sigma}^{\mathbf l}_i & {\mathbf \Sigma}^{\mathbf l}_{ij} \\
{\mathbf \Sigma}^{\mathbf l\top}_{ij} & {\mathbf \Sigma}^{\mathbf l}_{j}
\end{bmatrix} =
\begin{bmatrix}
\mathbf P_{i1} + \mathbf P_{i2} & \mathbf P_{ij} \\
\mathbf {P}_{ij}^\top & \mathbf P_{j1} + \mathbf P_{j2}
\end{bmatrix}
\end{align*}
We obtain two covariance estimates independently from two poses and we will use split covariance intersection (SCI) \cite{Li2013} to compute an upper bound on the actual landmark covariance as follows
\begin{equation}
{\hat {\mathbf \Sigma}}^{\mathbf l, -1} = (\frac{1}{\omega} \mathbf P_{i1} + \mathbf P_{i2})^{-1} + (\frac{1}{1-\omega}\mathbf P_{j1} + \mathbf P_{j2})^{-1},\omega \in [0, 1],
\label{eq:bruce-sci}
\end{equation}
where $\omega$ can be optimized via
\begin{equation}
\omega^* = \argmin_{\omega \in (0, 1)} \det({\hat {\mathbf \Sigma}}^{\mathbf l}).
\label{eq:bruce-sci-w}
\end{equation}
Readers are referred to \cite{Li2013} for analysis of SCI and proof of the upper bound. But the core idea is that if we construct an optimal linear, unbiased estimator $\hat{\mathbf l} = \mathbf K_i  {\mathbf l}_i + \mathbf K_j  {\mathbf l}_j (\mathbf K_i + \mathbf K_j = \mathbf I)$, then it can be proved that ${\hat {\mathbf \Sigma}}^{\mathbf l} \succeq \mathbb E[(\hat{\mathbf l} - \mathbf l)(\hat{\mathbf l} - \mathbf l)^\top]$. Therefore we are able to further approximate Equation~(\ref{eq:approx1}) by
\begin{equation}
\log \text{det}(\Sigma) < \log \text{det}(\Sigma_{\mathbf x_{T+N}}) + \sum_k \log \text{det}(\hat{\Sigma}_{\mathbf v_k}).
\label{eq:approx2}
\end{equation}
We demonstrate the process of incremental split covariance intersection in Fig.~\ref{fig:bruce-sci}. At each step, a landmark covariance estimate derived from the measurement collected at a specific pose is indicated as a dark blue ellipse. After the first step, we are able to fuse landmark observations from different poses via covariance intersection, using Equation~(\ref{eq:bruce-sci}), as shown by red ellipses. It is evident that the resulting ellipse from SCI (red) contains the true result obtained from SLAM (green).

\subsection{Extensibility to Pose SLAM} 
\label{sec:em-pose-slam} 
From the above derivation, an upper bound on the uncertainty of a real or virtual landmark may be computed from multiple poses that observe it. However it is worth investigating the meaning of this process in the context of pose SLAM, where no landmarks are incorporated into our optimization. To simplify the problem, we assume environmental features are measured and the transformation between poses with overlapping observations is derived using iterative closest point (ICP) \cite{Besl1992}. 
We visualize such a scenario using hypothetical range-bearing observations of an object (which may be alternately represented as a landmark) in the leftmost plot in Figure~\ref{fig:bruce-sci}.
In this example, the observations are normally distributed and centered at the actual landmark location. We assume this set of range-bearing returns will be matched with those from other poses using ICP. In ICP, a point is associated to its nearest neighbor and the matched feature is given by
\begin{equation}
\tilde{\mathbf l} = \argmin_{ \mathbf l_i } \Vert \mathbf l_i - \mathbf l\Vert_2.
\end{equation}
It is subsequently trivial to show that
\begin{equation}
{\mathbf \Sigma}^{\mathbf l} \succeq \mathbb E[(\hat{\mathbf l} - \mathbf l)(\hat{\mathbf l} - \mathbf l)^\top] \succeq \mathbb E[(\tilde{\mathbf l} - \mathbf l)(\tilde{\mathbf l} - \mathbf l)^\top].
\end{equation}
The distribution of $\tilde{\mathbf l}$ is visualized at every step in Figure~\ref{fig:bruce-sci}. While we aim to select exploration actions that reduce the (virtual) landmark covariances in landmark-based SLAM, we essentially minimize the closeness of measurements in ICP-based pose SLAM. Because association error is one major cause of ICP error, tightly clustered target points greatly contribute to registration performance.

\subsection{Virtual Map}
\label{sec:virtual-map}
How can we predict unobserved landmarks without prior knowledge of the characteristics of an environment? We can approach this question by making a conservative assumption that any location which has not been mapped yet has a virtual landmark. Additionally, the probability that a location is potentially occupied with a landmark can be captured using an occupancy grid map. Let $P(m_i \in \mathcal M)$ denote the occupancy of a discretized cell, then we define a virtual map $\mathcal V$ consisting of virtual landmarks with probability
\begin{eqnarray}
P(v_i = 1) = \begin{cases}
1, & \text{if}\ P(m_i = 1) \ge 0.5\\
0, & \text{otherwise}.
\end{cases}
\label{eq:virtual-landmark}
\end{eqnarray}
%Under circumstances with severe drift, the expected map could be utilized to calculate a weighted average of occupancy grid maps from all trajectory hypotheses, as in our previous work \cite{Wang2018}. In Fig. \ref{fig:vm}, we whiten the grid cells that, once observed, no longer possess virtual landmarks.
In its definition, existing landmarks have been incorporated into the virtual map, which is essential because minimizing the uncertainty of observed landmarks is also desired. 

\begin{figure*}
	\centering
	\subfloat[Current pose]{\includegraphics[width=0.32\textwidth]{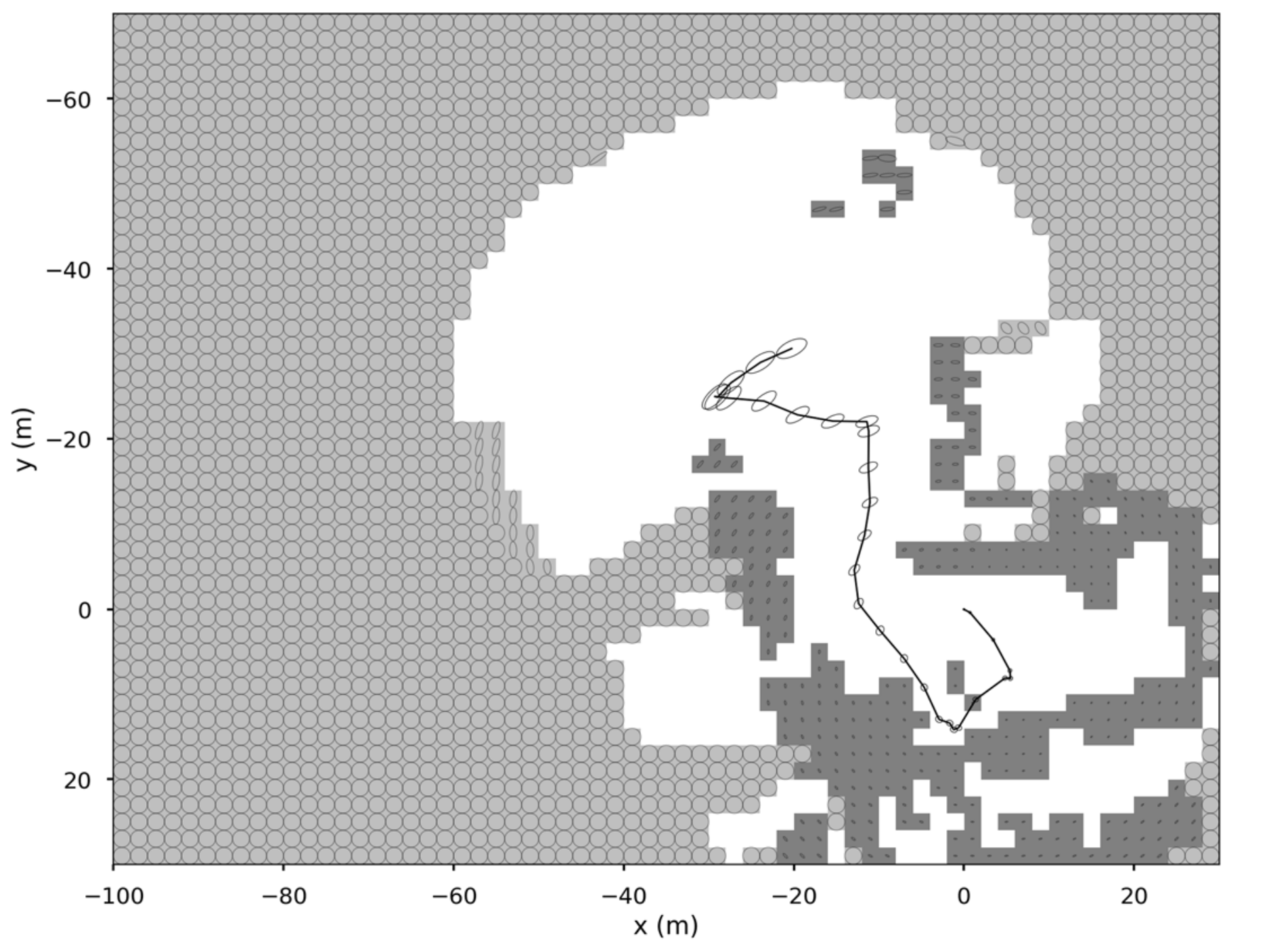}}\
	\subfloat[Candidate trajectory 1]{\includegraphics[width=0.32\textwidth]{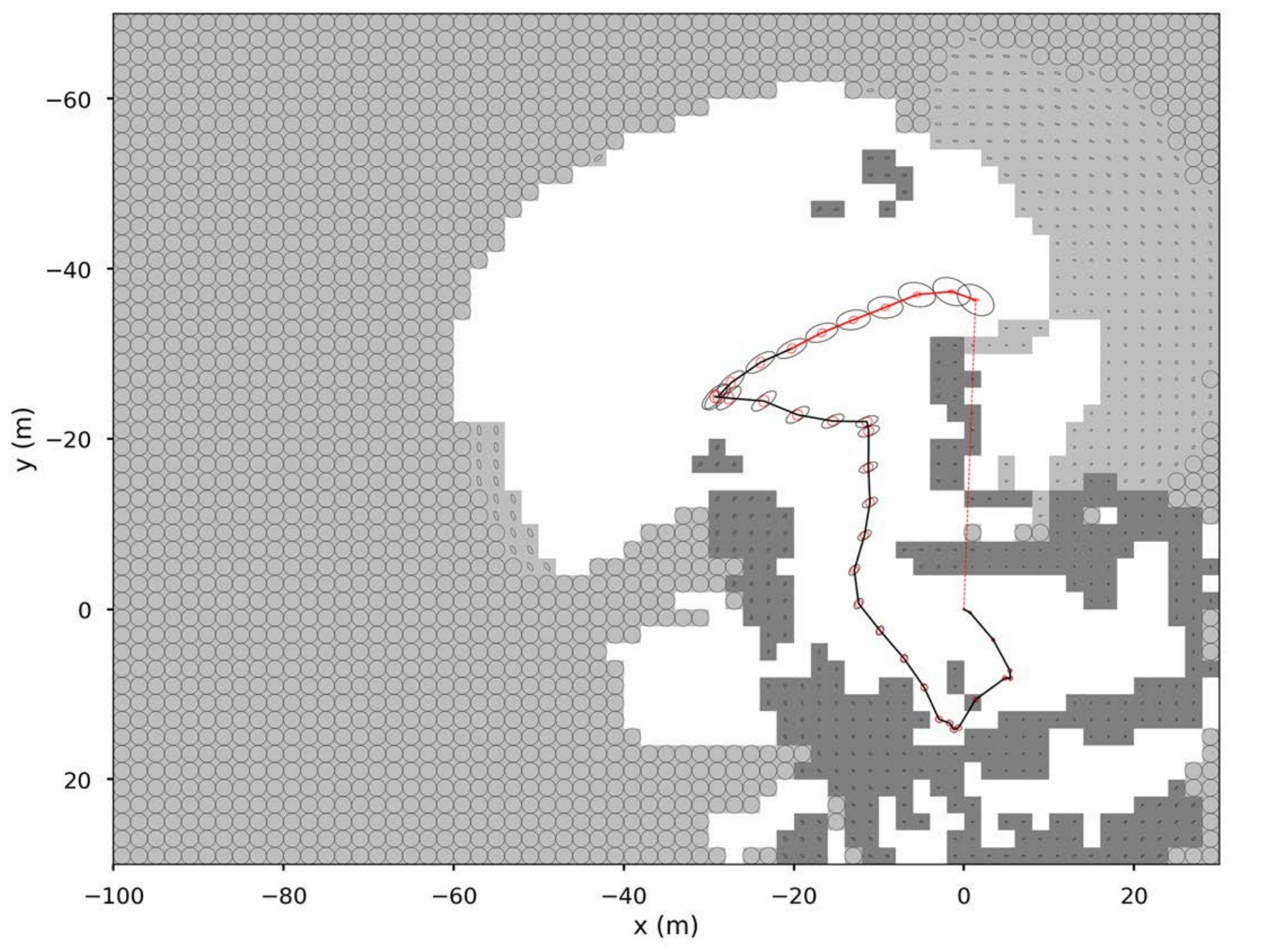}}\
	\subfloat[Candidate trajectory 2]{\includegraphics[width=0.32\textwidth]{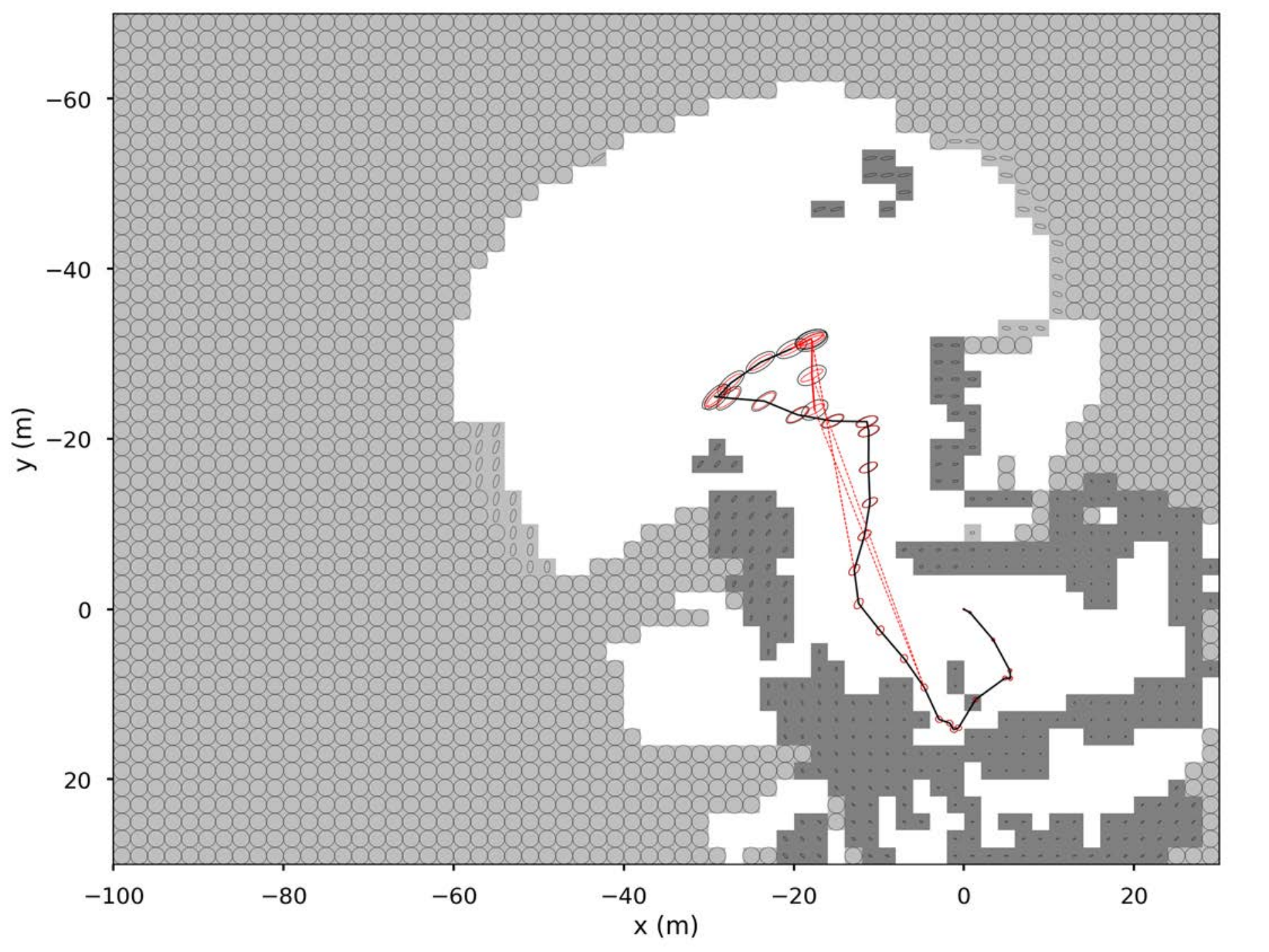}}\\
	\subfloat[Candidate trajectory 3]{\includegraphics[width=0.32\textwidth]{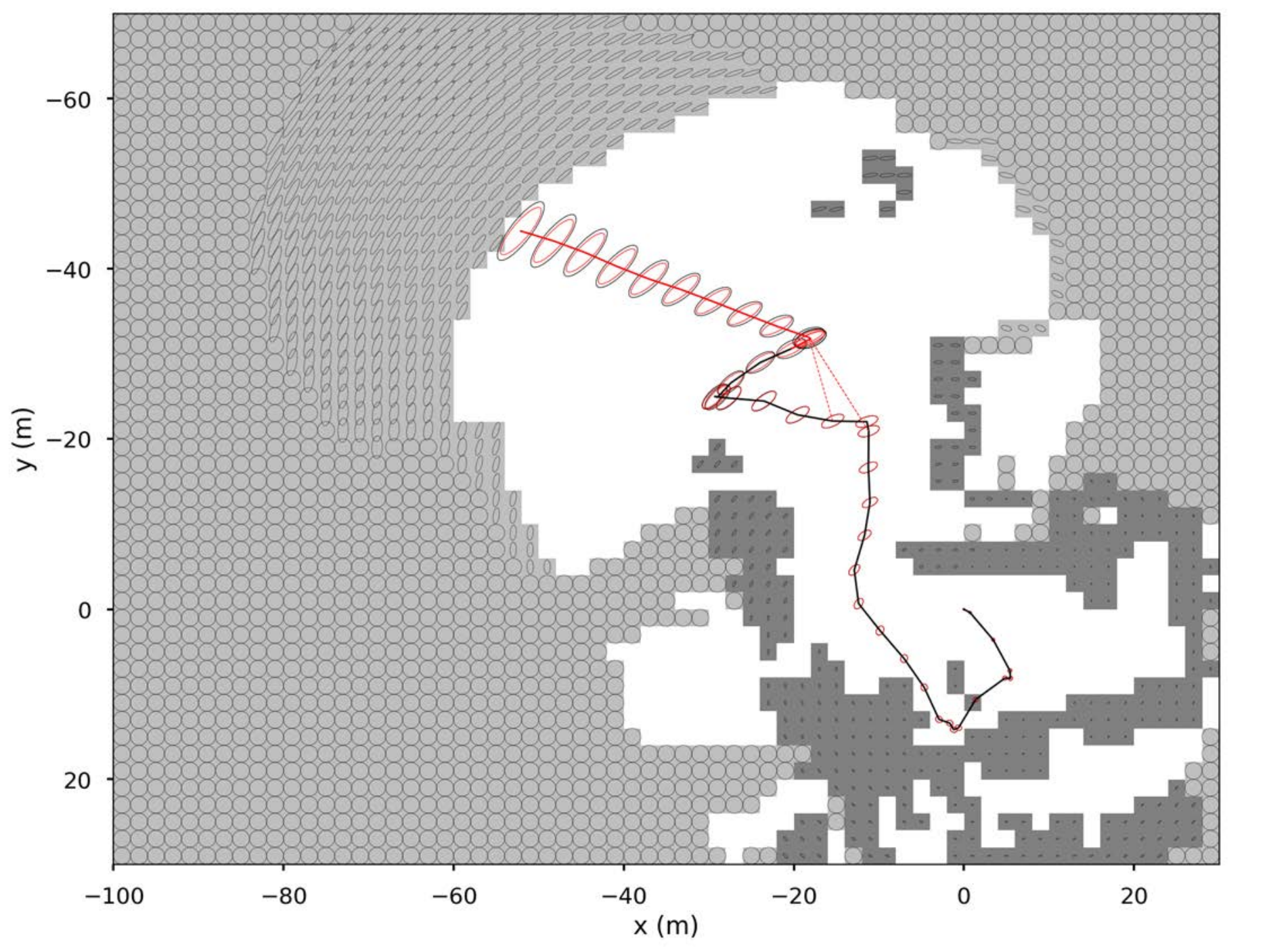}}\
	\subfloat[Candidate trajectory 4]{\includegraphics[width=0.32\textwidth]{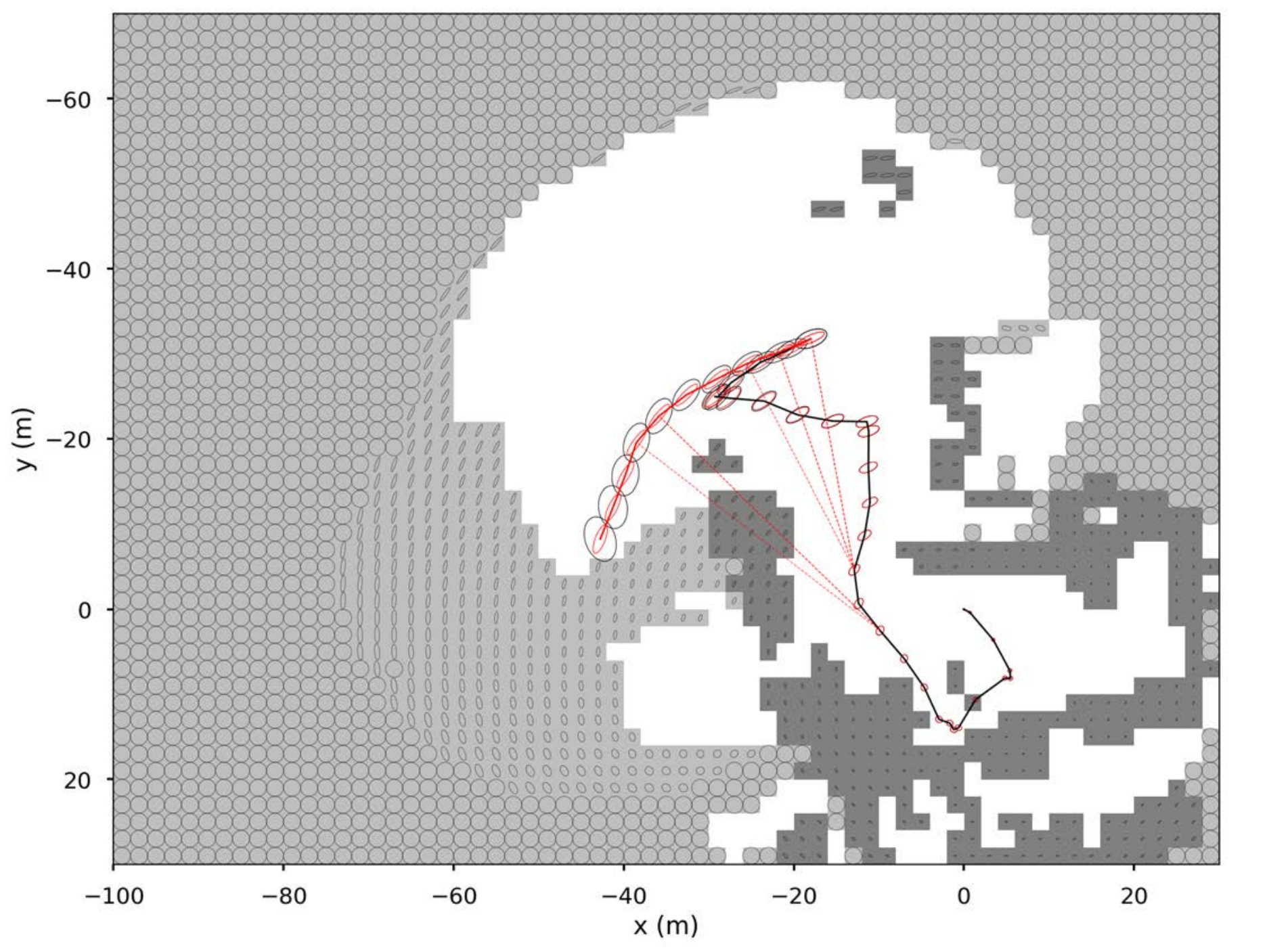}}\
	\subfloat[Candidate trajectory 5]{\includegraphics[width=0.32\textwidth]{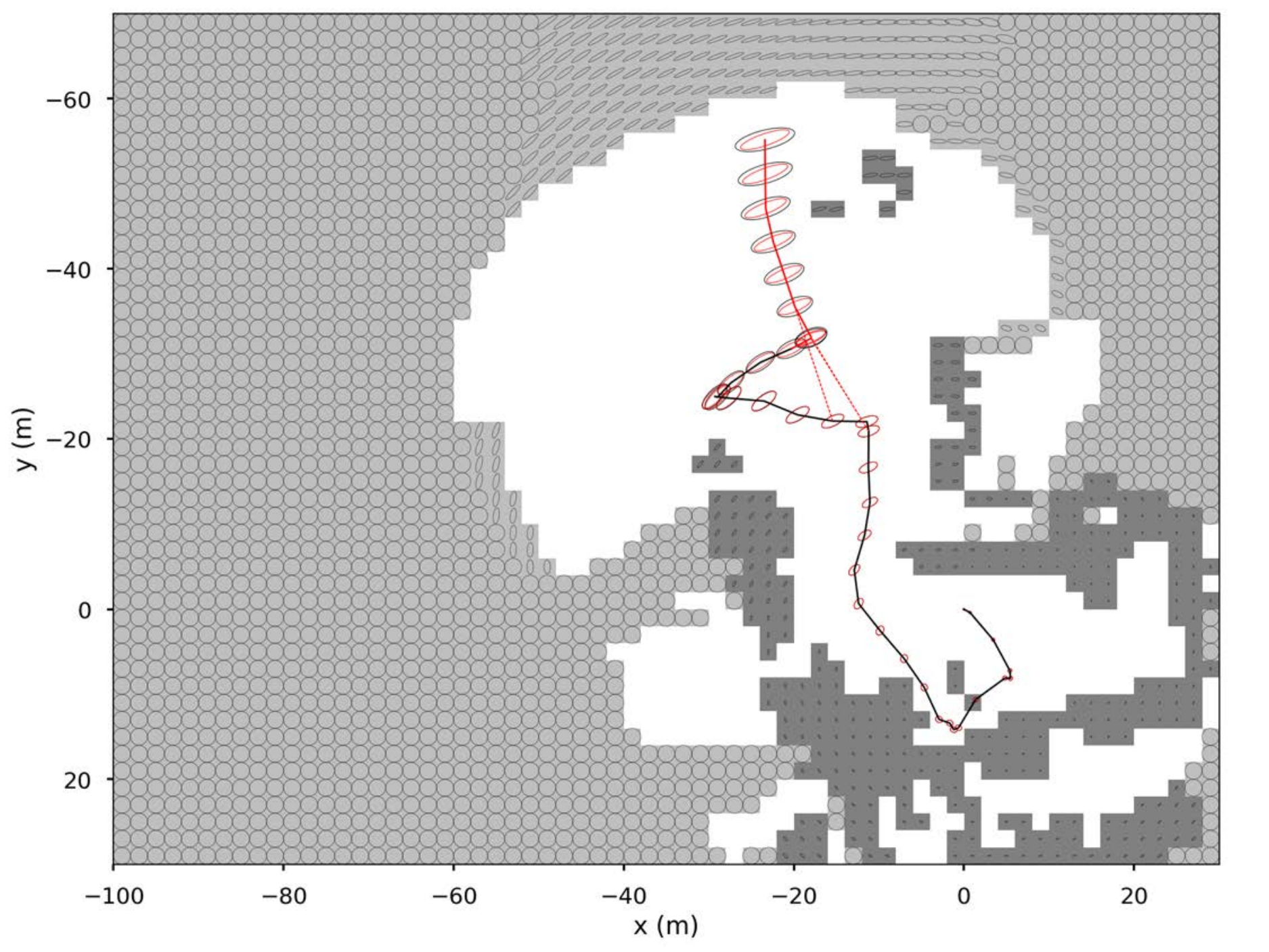}}\\
	\caption{Planning over virtual maps, with several candidate trajectories shown. Virtual landmarks are maintained within the cells of a low-resolution occupancy map (the higher-resolution occupancy maps used for motion planning and collision avoidance are shown throughout Section \ref{sec:results}). The error covariances of all occupied and unknown cells are shown as ellipses drawn inside each respective map cell, along with anticipated loop closure constraints. The maps shown were produced using real sonar data from the environment depicted in Fig. \ref{fig:bruce-real-setup}.
	%\textcolor{red}{BE: Not exactly sure which section this belongs in. I think this should be shown, so the reader understands what virtual maps look like, but does it go in Section 3.D, 4.D, or 5.C?}\Comment{YH: if we don't put fig:bruce-occ-mapping, we can cite this in 4.D}
	}
	\label{fig:bruce-occ-mapping}
	\vspace{-4mm}
\end{figure*}

What distinguishes a typical occupancy grid map from our virtual map is the treatment of unknown space, which consequentially determines what metric we use for exploration. In occupancy grid mapping, in order to enable exploration towards unknown space, Shannon entropy is frequently leveraged to optimize a path that provides maximum information gain \cite{Elfes1995}. However, another metric, such as an optimality criterion defined over the covariance matrix, is required to take into account mapping and localization uncertainty. Accordingly, virtual landmarks are initialized to have high uncertainty, which is reduced by taking measurements of them, resulting in information gain. The same metric can be further optimized by improving localization through loop closures, thus unifying the utility measure used in both exploration and localization. An example of an occupancy map used to maintain virtual landmarks is provided in Figure \ref{fig:bruce-occ-mapping}, where it is used for belief propagation in support of planning with an underwater robot.

%\revision{
Although we can use the same occupancy grid map from path planning, which is typically high-resolution, to construct the virtual map using Eq. (\ref{eq:virtual-landmark}), it has been demonstrated in our earlier work \cite{Wang2017} that a low-resolution virtual map provides similar exploration performance but requires less computation. Therefore, in our experiments, we downsample the occupancy grid map from \SI{0.2}{m} resolution to $d = \SI{2}{m}$, shown in Fig. \ref{fig:bruce-occ-mapping}.

The update of the virtual map and estimation of virtual landmark covariances are described in Algorithm \ref{alg:virtual-map}. The virtual map is first obtained by downsampling the occupancy grid map, which is constructed from the robot's trajectory and measurements collected at each pose. Given a candidate path $X_{T:T+N}$, we are able to predict potential loop closure measurements depending on the specific SLAM algorithm. For instance, for landmark-based SLAM, a loop closure occurs when an existing landmark is observed at a future pose. As for pose SLAM, we predict a loop closure whenever two poses observe obstacles with a certain amount of overlap, such that scan matching is possible. We note that we assume maximum likelihood observations when predicting future measurements. The pose uncertainty is updated provided the candidate actions and measurements using Eq. (\ref{eq:bruce-cov-update}). For each virtual landmark, we use the covariance intersection algorithm to estimate the virtual landmark covariance as described in Sec. \ref{sec:landmarks}.

\begin{algorithm}
\begin{algorithmic}[1]
\caption{Virtual Landmark Covariance Estimate}
\label{alg:virtual-map}
\Require {virtual map resolution $d$, trajectory $\mathcal X$ and measurements $\mathcal Z$, candidate path $X_{T:T+N}$ and future measurements $Z_{T:T+N}$}
\State $\mathcal M \leftarrow \text{UpdateOccupancyMap}(\mathcal X, \mathcal Z)$
\State {$\mathcal V \leftarrow \text{Downsample}(\mathcal M, d)$}\Comment{Eq. (\ref{eq:virtual-landmark})}
\State $\mathbf \Sigma' \leftarrow \text{UpdateCovarianceDiagonal}(\mathcal X, \mathcal Z, X_{T:T+N}, Z_{T:T+N})$\Comment{Eq. (\ref{eq:bruce-block-diag}) -- (\ref{eq:bruce-cov-update})}
\For{$\mathbf v_k \in \{\mathbf v_i \in \mathcal V\ |\ P(\mathbf v_i = 1) = 1\}$}

\State $\mathbf \Sigma_{\mathbf v_k} \leftarrow \text{CovarianceIntersect}(\mathbf \Sigma', X_{T:T+N}, \mathbf v_k)$\Comment{Eq. (\ref{eq:bruce-virtual-meaurement-model}) -- (\ref{eq:bruce-sci-w})}
\EndFor
\end{algorithmic}
\end{algorithm}

%} % end revision

\vspace{-5mm}

\subsection{Motion Planning}
\label{sec:motion-planning}

To carry out the M-step of our algorithm introduced in Equation (\ref{eq:m}), given the distribution of virtual landmarks, path candidates must be generated and evaluated using the selected utility function. If we are to consider global paths that will be followed over a long span of time, we must take into account two types of actions, exploration and place-revisiting \cite{Stachniss2005}. \textit{Exploration actions} normally have destinations near frontier locations where mapped cells meet unknown cells, and to reduce localization uncertainty, \textit{place-revisiting actions} travel back to locations the robot has visited, or where it is able to observe a previously observed obstacle. The prevalence of these locations requires us to examine a large number of free grid cells in order to obtain a near-optimal solution.

Therefore, we define two sets of goal configurations in our action space. A set of frontier goal configurations $\mathcal G_{\text{frontier}}$ is used for exploration, and a set of place-revisiting goal configurations $\mathcal G_{\text{revisitation}}$ is used to correct the robot's localization error. Algorithm \ref{alg:frontier} is designed for generating the frontiers, which are uniformly sampled on the boundary between the unknown and explored regions of the workspace. In Algorithm \ref{alg:revisting}, the place-revisiting set of goal configurations is generated by choosing the candidates which can observe the largest number of occupied cells. The size of $\mathcal G_{\text{revisitation}}$ is correlated with the number of occupied cells in the current map. Specifically, occupied cells are divided into $k$ clusters using the $k$-means algorithm, and revisitation goals are generated along the boundary of a circle placed at each cluster center, whose radius depends on the sensor range. For each set of goal configurations, we sample from available candidates in favor of maximizing the distance to the nearest obstacle, which is achieved using the clearance map $c$. This map contains, in every unoccupied cell, the distance to the nearest obstacle.

\begin{algorithm}
\begin{algorithmic}[1]
\caption{Frontier Goal Configurations}
\label{alg:frontier}
\Require {Frontier cells $\mathcal F$, clearance map $c$, number of candidate goals $N_f$, minimum separation distance $d$}
\State $\mathcal G_{\text{frontier}} \leftarrow \emptyset$
\While {$|\mathcal G_{\text{frontier}}| < N_f$}
\State $\mathbf f^* \leftarrow \argmax_{\mathbf f_i} c(\mathbf f_i)$
\State $\mathcal G_{\text{frontier}} \leftarrow \mathcal G_{\text{frontier}} \cup \{\mathbf f^*\}$
\State $\mathcal F \leftarrow \mathcal F \setminus \{\mathbf f_i\ |\ |\mathbf f_i - \mathbf f^*| \le d\}$
\EndWhile
\State \Return {Candidate exploration goals $\mathcal G_{\text{frontier}}$}
\end{algorithmic}
\end{algorithm}

\begin{algorithm}
\begin{algorithmic}[1]
\caption{Place-revisiting Goal Configurations}
\label{alg:revisting}
\Require {Occupied cells $\mathcal O$, clearance map $c$, number of candidate goals $N_r$, revisitation radius $r$, minimum separation distance $d$}
\State $\mathcal G_{\text{revisitation}} \leftarrow \emptyset$
\State $\mathcal M = \{\mathbf m_{1, ..., k}\} \leftarrow \text{ComputeKMeans}(\mathcal O, k)$
\While {$|\mathcal G_{\text{revisitation}}| < N_r$}
\State $\mathbf m^* = (x^*, y^*) \leftarrow \argmax_{\mathbf m_i} |\mathcal O_i|$
\State $\mathcal R^* \leftarrow \{(x^* + r\cos(\theta), y^* + r\sin(\theta)) | \theta \in [0, 2\pi)\}$
\State $\mathbf r^* \leftarrow \argmax_{\mathbf r_i\in \mathcal R^*} c(\mathbf r_i)$
\State $d^* \leftarrow \min_{\mathbf g_i \in \mathcal G_{\text{revisitation}}} |\mathbf r^* - \mathbf g_i|$
\If {$d^* \ge d$}
\State $\mathcal G_{\text{revisitation}} \leftarrow \mathcal G_{\text{revisitation}} \cup \{\mathbf r^*\}$
\EndIf
\State $\mathcal M \leftarrow \mathcal M \setminus \{\mathbf m^*\}$
\EndWhile
\State \Return {Candidate revisitation goals $\mathcal G_{\text{revisitation}}$}
\end{algorithmic}
\end{algorithm}

To support motion planning to the above sets of goal configurations, a finely discretized roadmap is generated at the outset of an exploration mission (the boundaries of the region to be explored are known, and roadmap edges are pruned out as obstacles are discovered), and goal configurations are assigned to their nearest neighbors in the roadmap. The A* search algorithm \cite{Hart1968} is adopted to solve single-source shortest paths to all goal configurations from the robot's current pose, and we select the best path among them using a utility function that maps a candidate path to a scalar value. In Eq. (\ref{eq:approx1}), the log-determinant of the covariance matrix is derived from the M-step as our uncertainty metric. Since the estimated covariance has to be fused with a large initial covariance, the log-determinant, or D-optimality criterion, is guaranteed to be monotonically non-increasing during the exploration process, which is consistent with the conclusion in \cite{Carrillo2015}. In addition to uncertainty criteria, it is valuable to incorporate a cost-to-go term to establish a trade-off between traveling cost and uncertainty reduction \cite{Stachniss2005}. Thus, our utility is finalized as,
\begin{align}
\label{eq:utility-em}
U_{\text{EM}}(X_{T:T+N}) = &-\log \text{det}(\tilde \Sigma_{\mathbf x_{T+N}}) - \sum_k \log \text{det}(\hat \Sigma_{\mathbf v_k})\nonumber\\
&- \alpha d(X_{T:T+N}),
\end{align}
where $\alpha$ is a weight on path distance $d(X_{T:T+N})$. In our experiments, we adopt a linearly decaying weight function with respect to traveled distance, whose parameters are determined experimentally and applied consistently throughout our algorithm comparisons below. The selected path is executed immediately, but to account for deviation from the nominal trajectory during execution and inaccurate prediction after taking new measurements (which may discover new obstacles), the path is discarded when it is blocked by obstacles, or the robot has traveled a designated distance. The exploration planning process is repeated until no new frontiers are detected.

As we can observe from Eq. (\ref{eq:utility-em}), the utility of uncertainty reduction can be negated by the action cost, and under poorly selected parameters, a robot may become stuck at its current location. For instance, if the virtual landmark prior is small, but the odometry drift rate and range sensor uncertainty are high, it is unnecessary to take actions to explore, as the uncertainty of virtual landmarks will not be reduced. Therefore, in our experiments, the virtual landmark prior is chosen to be larger than the worst-case uncertainty resulting from a range sensor observation at the end of a lengthy dead-reckoning trajectory. This ensures that our robot will fully explore its environment.

%\revision{
\subsection{Computational Complexity Analysis}
\label{sec:complexity-analysis}

We briefly analyze the computational complexity of the proposed EM exploration algorithm. At each step, we generate $N_f + N_r$ goal configurations in total and plan collision-free trajectories to each of them. As in Algorithms \ref{alg:frontier} and \ref{alg:revisting}, goal configurations can be generated in $O(N_fN_{\text{cell}})$ and $O(N_rN_{\text{cell}})$ respectively, where $N_\text{cell}$ is the number of grid cells in the occupancy grid map. All pose covariances are estimated using predicted measurements on candidate trajectories, each of which has a general bound of $O(N_\text{variables}^3)$, where $N_\text{variables}$ is the number of variables in the SLAM problem (i.e., poses and landmarks). However, by means of incremental covariance update in Sec. \ref{sec:actions}, we are able to achieve more efficient computation considering there are a small number of future loop closures. Finally, virtual landmark covariance is estimated, which formulates the utility function. As virtual landmark covariance is computed by fusing multiple sources, it has complexity $O(N_\text{occupied} N_\text{poses})$, where $N_\text{occupied}$ is the number of occupied cells in the virtual map and $N_\text{poses}$ is the maximum number of poses that can observe the same landmark. The computation is greatly alleviated by the fact that we use a low-resolution virtual map and the number of occupied cells decreases as we explore.
%}

\section{Underwater SLAM}
\label{sec:uw_slam}

In this section, we describe a planar, three degree-of-freedom (position $x, y$ and heading $\theta$) SLAM framework that allows us to robustly implement our EM exploration procedure on a sonar-equipped underwater robot in a cluttered harbor environment. In this setting, sensor observations are dense enough that a pose SLAM approach is adopted, used in conjunction with occupancy mapping to support motion planning. As described in Sections \ref{sec:em-pose-slam} and \ref{sec:virtual-map}, virtual landmarks are also maintained throughout the workspace and used to evaluate the EM exploration utility function. Belief propagation over virtual landmarks in this setting is illustrated in Figure \ref{fig:bruce-occ-mapping}.

\begin{figure}
	\centering
	\includegraphics[width=0.95\columnwidth]{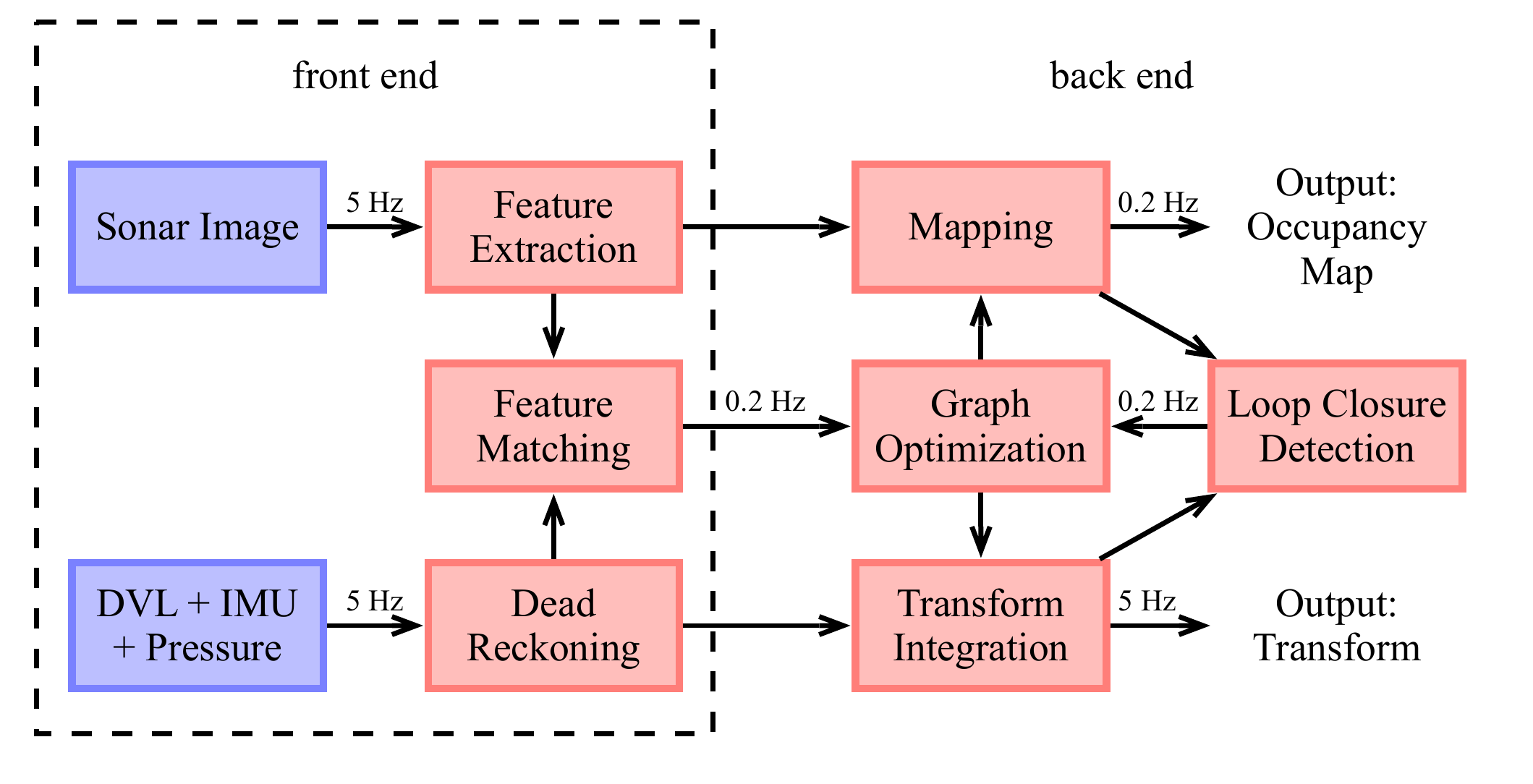}
% 	\resizebox{\textwidth}{!}{%
% 	\begin{tikzpicture}[scale=0.5, every node/.style={scale=0.5}]
	
% 	\node[BlueBlock] (sonar) {Sonar Image};
% 	\node[OrangeBlock] (feature) [right of=sonar] {Feature Extraction};
% 	\node[OrangeBlock] (map) [right of=feature] {Mapping};
% 	\node[EndBlock] (map2) [right of=map]{Occupancy Map Output};
	
% 	\node[EmptyBlock] (empty) [below of=sonar] {};
% 	\node[OrangeBlock] (ssm) [right of=empty] {Feature Matching};
% 	\node[OrangeBlock] (sam) [right of=ssm] {Graph Optimization};
% 	\node[OrangeBlock] (nssm) [right of=sam] {Loop Closure Detection};
	
% 	\node[BlueBlock] (imu) [below of=empty] {DVL/IMU/Pressure};
% 	\node[OrangeBlock] (odom) [right of=imu] {Dead Reckoning};
% 	\node[OrangeBlock] (tf) [right of=odom] {Transform Integration};		
% 	\node[EndBlock] (tf2) [right of=tf]{Transform Output};
	
% 	\draw[Arrow] (sonar) -- node [ArrowText] {\tiny \SI{5}{\hertz}}(feature);
% 	\draw[Arrow] (imu) -- node [ArrowText] {\tiny \SI{5}{\hertz}}(odom);
% 	\draw[Arrow] (feature) -- (ssm);
% 	\draw[Arrow] (odom) -- (ssm);
% 	\draw[Arrow] (ssm) -- node [ArrowText] {\tiny \SI{0.2}{\hertz}} (sam);
% 	\draw[Arrow] (sam) -- (map);
% 	\draw[Arrow] (feature) -- (map);
% 	\draw[Arrow] (map) -- (nssm);
% 	\draw[Arrow] (tf) -- (nssm);
% 	\draw[Arrow] (nssm) -- node [ArrowText] {\tiny \SI{0.2}{\hertz}} (sam);
% 	\draw[Arrow] (sam) -- (tf);
% 	\draw[Arrow] (odom) -- (tf);
% 	\draw[Arrow] (tf) -- node [ArrowText] {\tiny \SI{5}{\hertz}}(tf2);
% 	\draw[Arrow] (map) -- node [ArrowText] {\tiny \SI{0.2}{\hertz}}(map2);
%	\end{tikzpicture}
% }%
	\caption[Diagram of the underwater SLAM framework.]{Diagram of the underwater SLAM framework. Both front end and back end are applied in real underwater robot experiments, while our simulation-based results only utilize the SLAM backend.} 
	\label{fig:bruce-slam-diagram}
	\vspace{-4mm}
\end{figure}

The underwater SLAM framework is built upon a keyframe-based pose graph.
%Each keyframe is chosen based on the following heuristic: a keyframe is inserted when the robot moves a certain distance or rotates a certain angle. 
 A keyframe is added to the graph when the robot translates more than \SI{4}{m} or rotates more than \SI{30}{\degree}, which as shown in Fig.~\ref{fig:bruce-slam-diagram}, typically occurs at about \SI{0.2}{\hertz}. Considering our ROV travels 
%operated manually or autonomously 
at a low speed and its sonar has a large field of view, this keyframe update rate is sufficient to support our experiments. The front end, which is used in our real-time underwater experiments, performs scan matching on features extracted from sonar images in keyframes. The scan matching relies on an initial pose estimate from dead reckoning, which fuses three sources of measurements: Doppler velocity log (DVL), inertial measurement unit (IMU), and a pressure (depth) sensor at \SI{5}{\hertz}. In the back end, the constraints from the front end are incorporated into a pose-only factor graph. Pose graph optimization (using iSAM2 \cite{Kaess2012}) provides a low-frequency state estimate, which is integrated with dead reckoning to produce a high-frequency estimate. Given the estimated trajectory and feature points from keyframes, the mapping module generates an occupancy map which is used to support motion planning. Dead reckoning drift is corrected using the constraints provided by loop closure detection, which serves to match the current keyframe with the past history of keyframes.

\subsection{Feature Extraction}

\begin{figure}[tp]
	\centering
	\includegraphics[width=0.99\columnwidth]{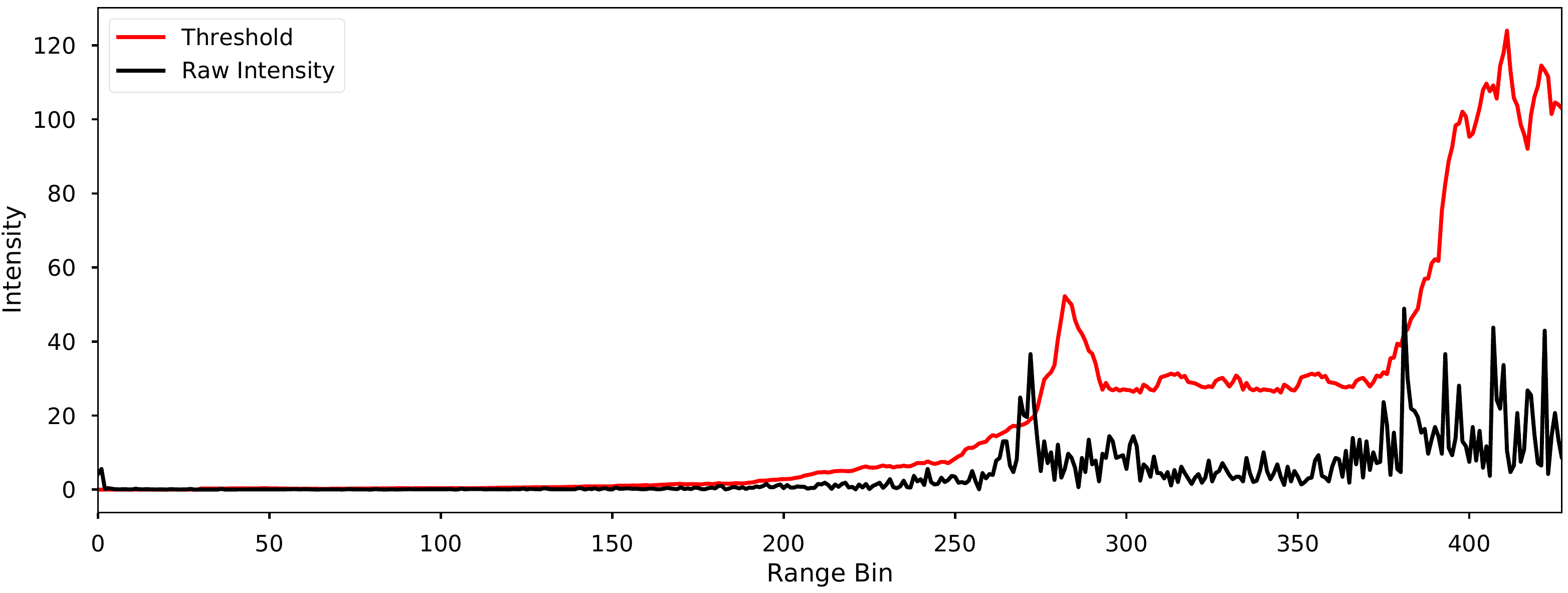}\\
    \includegraphics[width=0.99\columnwidth]{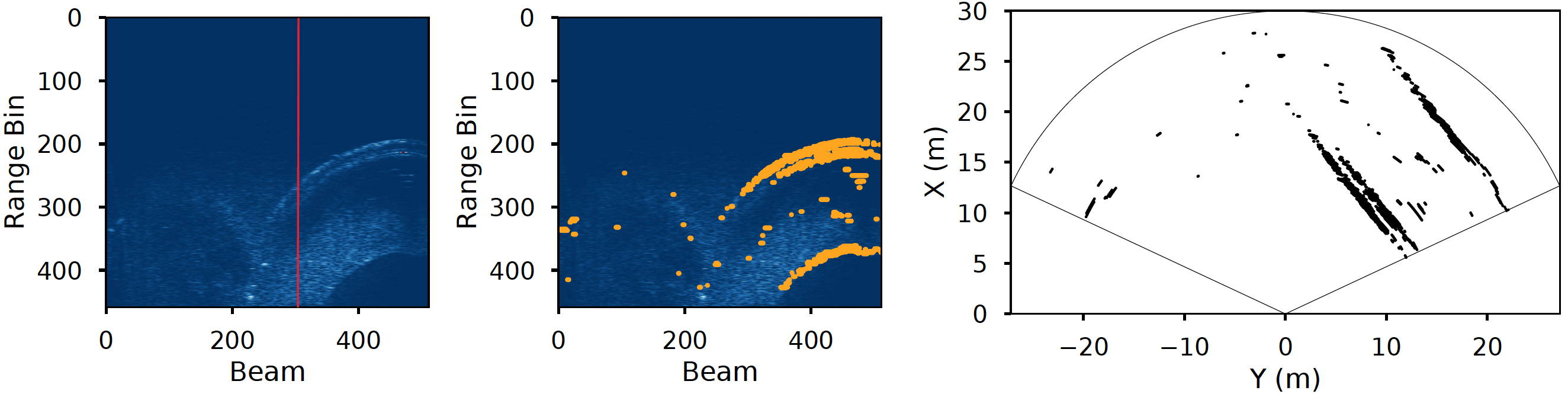}
	\caption{SOCA-CFAR feature detection. A raw sonar image from an Oculus M750d with 512 beams is shown at bottom left. The adaptive threshold corresponding to the sonar beam marked with a red line is represented by the red plot line at top. Cells with intensity larger than the threshold are treated as targets, shown in orange at bottom center and transformed to Cartesian coordinates at bottom right.}
	\label{fig:bruce-soca1}
	\vspace{-4mm}
\end{figure}

The 2D imaging sonar emits sound waves, and the echo received by the sonar carries the characteristics of the imaging area, which is represented by $\{(r_i, b_j), 0 \le i \le N_r, 0 \le j \le N_b\}$ in polar coordinates. We assume the measurements corresponding to every beam are independent and each beam is processed separately. Given a beam of intensity measurement $\mathbf z_j = \{I_{ij} = I(r_i, b_j), 0 \le i \le N_r\}$, we are interested in identifying range bins that represent sound pulses returned by underwater structures. 
%The extracted features provide better situational awareness in cluttered environments, helping improve localization, decision-making and obstacle avoidance. Examples of measurements from 2-d imaging sonar are shown in Figure~\ref{fig:bruce-soca1}.

In order to extract features reliably in all kinds of environments with varying noise power, we employ a Constant False Alarm Rate (CFAR) detector, which initially was designed for radar, for 2D imaging sonar.  
CFAR uses a simple threshold to determine if a pixel in a given image is a contact or not a contact, which is derived by computing a noise estimate for the area around the cell under test (CUT) via cell averaging. 
A CUT
is compared against test statistics using the cells in a sliding window excluding the CUT.
Let $\tilde T_{ij}$ be the statistics derived from the sliding window (e.g., the average intensity in the
window), and let $T_{ij}$ be the test statistics.

A variety of CFAR detectors have been proposed,
%and one of the most widely used ones is cell-averaging CFAR (CA-CFAR), where the test statistic is defined as the average intensity in a sliding window around the tested range bin. To tackle different operational conditions, e.g., single target, multi-target and clutter edge, smallest-of cell-averaging CFAR (SOCA-CFAR), greatest-of cell averaging CFAR (GOCA-CFAR) \cite{Hansen1980} and order-statistic (OS) CFAR \cite{Rohling1983}have been proposed. 
and based on our experiments with imaging sonar, we have found smallest-of cell-averaging CFAR (SOCA-CFAR) \cite{Richards2005} to provide a superior trade-off between accuracy and computational expense. In describing our approach below, we adopt the notation from \cite{Gandhi1988}.

%SOCA-CFAR computes four noise estimates and utilizes the smallest of the four.
Suppose the cells around the CUT are $\{I_{i - N, j},  ..., I_{i - 1, j}, I_{i+1, j}, ..., I_{i +N, j}\}$. The average intensities computed from the leading and trailing edge of the sliding window are given, respectively, by
\begin{equation}
	\tilde T_{ij, <} = \frac{1}{N}\sum_{n=1}^{N}I_{i-n, j}, \;\; \tilde T_{ij, >} = \frac{1}{N}\sum_{n=1}^{N}I_{i+n, j}.
\end{equation} 
The test statistics for SOCA-CFAR are then given by 
\begin{equation}
  T_{ij} = \tau \min(\tilde T_{ij, <}, \;\; \tilde T_{ij, >}),
\end{equation}
where $\tau$, given a specified false alarm rate, can be computed as follows \cite{Richards2005},
\begin{equation}
P_{\text{fa}} = \big( 2 + \frac{\tau}{N}\big)^{-N} \sum_{n=0}^{N-1}  {N - 1 + n \choose n}\big(2 + \frac{\tau}{N}\big)^{-n}. 
\end{equation}
Features detected by SOCA-CFAR $\{(r_{n_i}, b_{n_j}) | 0 \le n \le N\}$ can be transformed to Cartesian coordinates, forming a 2D point cloud as $\{\mathbf p_n = (x_n, y_n) \in \mathbb R^2| 0 \le n \le N\}$, where
\begin{equation}
x_n = r_{n_i}\cos(b_{n_j}),\quad y_n = r_{n_i} \sin (b_{n_j}).
\end{equation}
An example of SOCA-CFAR applied to an imaging sonar scan of structures in a marina is shown in Figure~\ref{fig:bruce-soca1}. Figure~\ref{fig:bruce-soca1} (a) visualizes the adaptive decision threshold $T_{ij}$ for the highlighted sonar beam, and Fig.~\ref{fig:bruce-soca1} (b) visualizes extracted feature points in polar and Cartesian coordinates.

% TODO: Change line color
% TODO: Change points color to oriange

\subsection{Feature Matching}
In this work, we assume our robot is operated at a fixed depth and the sonar takes measurements solely from objects in the same plane.
Let  $\mathbf x_t$ and $\mathbf x_s$ be a \textit{target} pose and \textit{source} pose, and let $\mathcal P_t$ and $\mathcal P_s$ be \textit{target} points and \textit{source} points measured at those poses. 
Specifically, the relative transformation between two poses $ \mathbf x_{ts}  \in \text{SE}(2)$ combining relative rotation $R_{ts} \in \text{SO}(2)$ and relative translation $\mathbf t_{ts} \in \mathbb R^2$, can be recovered using scan matching techniques. 
 %Therefore, both robot poses and relative measurements are elements in \textit{special Euclidean group} $\text{SE}(2) \doteq (\mathbf R, \mathbf t): \mathbf R\in \text{SO}(2), \mathbf t\in \mathbb R^2\}$. 
 %Let $\mathbf x_s \in \text{SE}(2), \mathbf x_t \in \text{SE}(2)$ be two 2D poses at which two sets of features, denoted as $ \mathcal P_s ,\mathcal P_t$, are observed respectively. The 2D points are represented in a local frame. Presumably,
%Assuming the vehicle is operated in a planar environment and the sonar takes measurements solely from objects in the same plane, the relative transformation between two poses at which two sonar images are taken can be recovered using scan matching techniques. 
%\begin{align}
% \mathbf x_s =  \mathbf x_t \oplus \mathbf x_{ts}, \mathbf x_{ts} = \mathbf x_t \ominus \mathbf x_s = (\mathbf R_{ts}, \mathbf t_{ts}),\\
 % \mathbf R_{ts} = \mathbf R_t^\top \mathbf R_s, \mathbf t_{ts} = \mathbf %R_t^\top(\mathbf t_s - \mathbf t_t)),
%\end{align}
%where we use $\oplus$ and $\ominus$ to denote the composition and  inverse composition of two elements in $\text{SE}(2)$.

%\subsubsection{Iterative Closest Point}
The most widely used algorithm is iterative closest point (ICP) \cite{Besl1992}, which minimizes the following cost function
\begin{align}
J({\mathbf x}_{ts}) &= \sum_{\mathbf p_t \in \mathcal P_t, \mathbf p_s \in \mathcal P_s} ||\mathcal T_{ts} \mathbf p_{s} - \mathbf p_{t}||^2\\
 &=\sum_{\mathbf p_t \in \mathcal P_t, \mathbf p_s \in \mathcal P_s} ||\mathbf R_{ts} \mathbf p_{s} + \mathbf t_{ts} - \mathbf p_{t}||^2,
\end{align}
where $\mathcal T \mathbf p = \mathbf R \mathbf p + \mathbf t$ represents the transformation of $\mathbf p \in \mathbb R^2$ due to a robot's motion  $\mathbf x \in \text{SE}(2)$. The correspondence between $\mathbf p_t$ and $\mathbf p_s$ in the above cost function is established by nearest neighbor search. However, it is well known that ICP is plagued by the problem of local minima due to non-convexity, and the final result heavily depends on the quality of the initial transformation. Typically the initial guess is estimated using dead reckoning - for instance, from IMU or wheel odometry. Scan matching using ICP is impaired more significantly in underwater environments, due to the relatively sparse quantity of points in a typical sonar scan and higher levels of noise.

\begin{figure}[t]
	\centering
	\includegraphics[width=0.99\columnwidth]{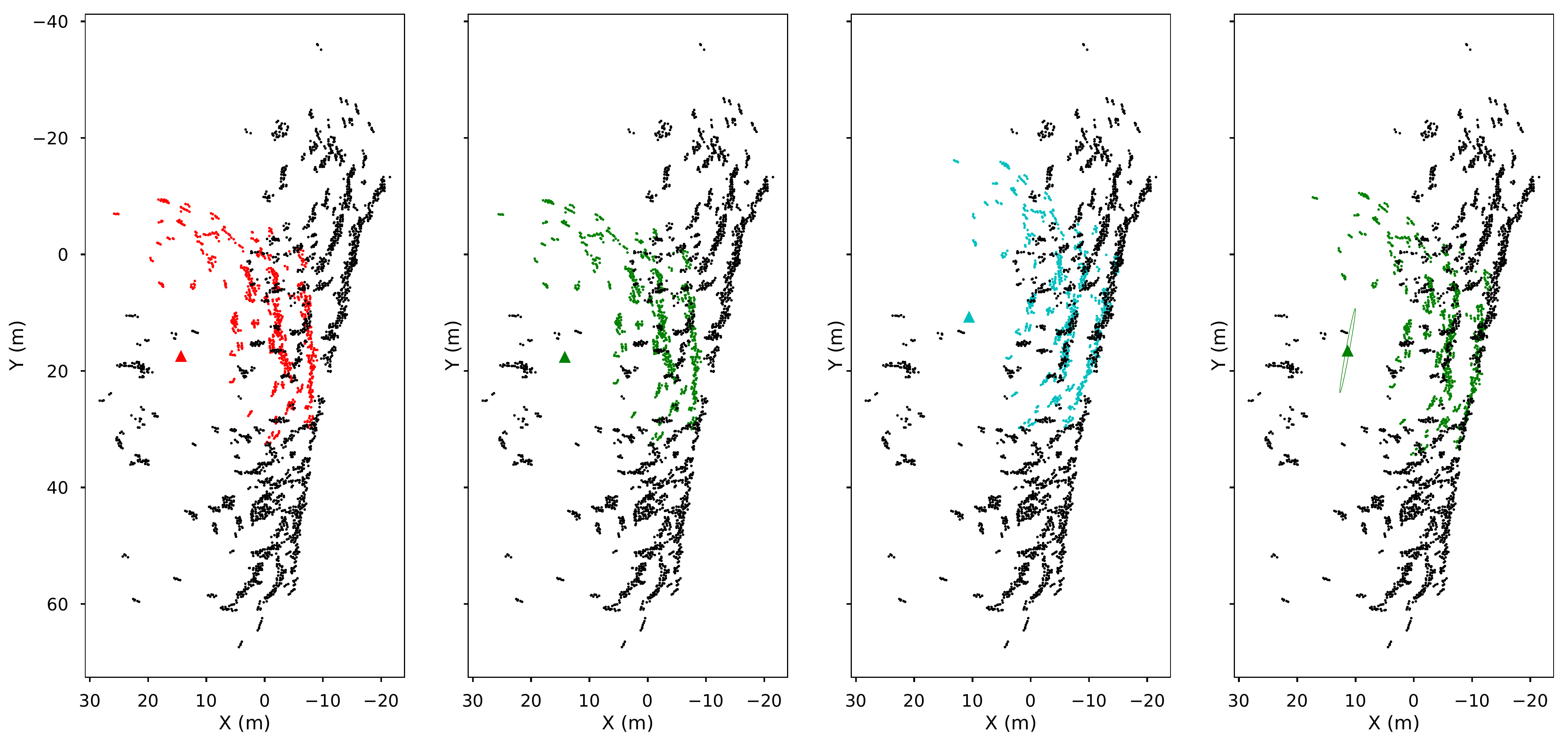}
	\caption[Illustration of our proposed sonar scan matching method applied to non-sequential keyframes.]{Illustration of our proposed sonar scan matching method applied to non-sequential keyframes to detect a loop closure. Colored points (source) represent features extracted from the current sonar frame and black points (target) represent accumulated features. From left to right: initial transformation from odometry; ICP result based on initial guess; global initialization result from our proposed sensing model; refined ICP result based on the global initialization in the third column. }
	\label{fig:bruce-icp-nssm}
\end{figure}

%\subsubsection{Globally initialized ICP}
%TODO: GO-ICP, etc.

%\revision{
We propose to alleviate the problem of local minima in ICP by performing a global initialization beforehand. Then ICP is used to locally refine the estimate. %}
Let $\tilde {\mathbf x}_s$ be the globally initialized source pose. The proposed global initialization follows consensus set maximization \cite{Fischler1981}, which solves the following problem,
\begin{equation}
\tilde {\mathbf x}_s = \argmax_{\mathbf x_s} \sum_{\mathbf p_s \in P_s} \mathbb I(d(\mathbf p_s) \le \epsilon),
\label{eq:bruce-icp-init-1}
\end{equation}
where $\mathbb I = 1$ when the condition is true, and $\mathbb I = 0$ otherwise. $d$ is defined as the distance to the nearest target point,
\begin{equation}
d(\mathbf p_s) = \min_{\mathbf p_t \in \mathcal P_t} \Vert \mathbf R_{ts} \mathbf p_s+ \mathbf t_{ts} - \mathbf p_t \Vert.
\end{equation}

Figure \ref{fig:bruce-icp-nssm} shows representative experimental results of feature matching applied to a loop closure where severe drift from the initial odometry measurement is observed. As we can see, the proposed multi-step solution yields good results.

\subsection{Building the Factor Graph}

\begin{figure}
	\centering
	\includegraphics[width=0.49\textwidth]{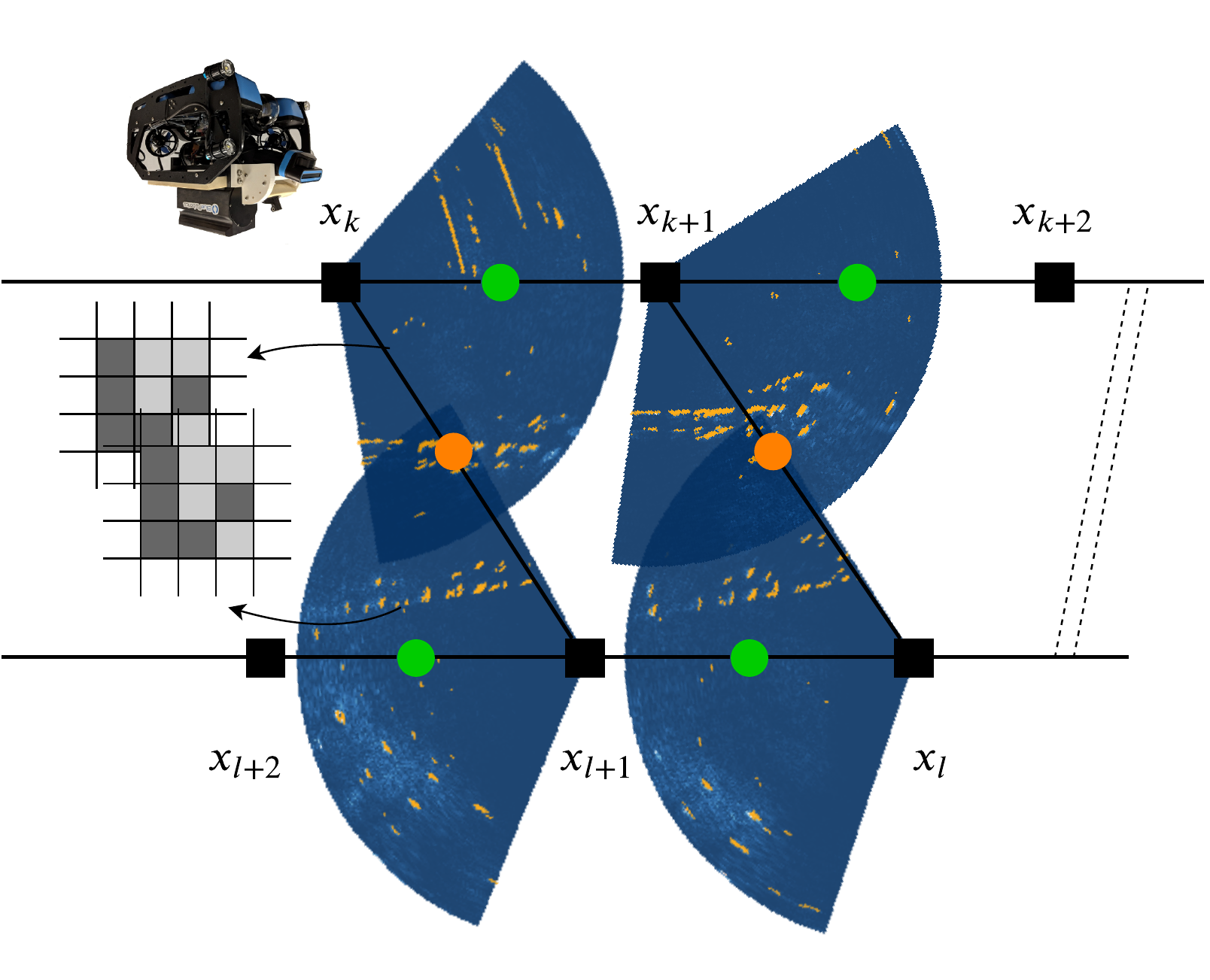}
	\caption[Key components of the SLAM system.]{Key components of the SLAM system. The pose variable at each keyframe is denoted as a black square. The core factor graph is comprised of pose variables at keyframes (black squares) and two types of factors: sequential scan matching factors (green circles) and loop closure factors (orange circles). Every keyframe is associated with feature points detected in the sonar image. The detected points at every keyframe independently contribute to the occupancy map.}
	\label{fig:bruce-factor-graph}
\end{figure}

%TODO: SLAM equation as in Jackal

In this section, we discuss the construction of factor nodes in the SLAM factor graph. As shown in Figure~\ref{fig:bruce-factor-graph}, there are two types of factor nodes, computed from sequential scan matching and non-sequential scan matching (denoted as green and orange nodes respectively). The sequential factors represent the transformation with respect to the previous pose; error is unavoidably accumulated along the chain of poses. In contrast, non-sequential factors, also known as loop closure constraints, connect two poses that are separated in terms of time and thus are able to correct the drift accumulated across sequential poses. The factor graph is given by
\begin{flalign*}
\mathbf f(\boldsymbol \Theta) = \mathbf  f^{\text{0}}(\boldsymbol \Theta_0) & \prod_i \mathbf f^{\text{O}}_{i}(\boldsymbol \Theta_i) \prod_j \mathbf f^{\text{SSM}}_j(\boldsymbol \Theta_j) \prod_q \mathbf f^{\text{NSSM}}_q(\boldsymbol \Theta_q).
\end{flalign*}

%\subsubsection{Sequential Scan Matching}

Sequential factors between two consecutive poses $\mathbf x_{k-1}$ and $\mathbf x_{k}$ can be computed by scan matching, using the point clouds collected at these two poses. To increase the robustness of scan matching, we incorporate extracted features from the previous $N_{ssm} > 1$ frames, assuming the drift is negligible within a short period of time.
%Let $\mathcal T_{k-1, k-i} \mathcal P_{k-i}$ be the transformed points at keyframe $k-i$ in the coordinate frame of pose $k-1$, then the target points are given by
%\begin{equation}
%\mathcal P_t := \bigcup_{i=1}^{N_{\text{ssm}}} \mathcal T_{k-1, k-i} %\mathcal P_{k-i}.
%\end{equation}
%An illustrative diagram of building sequential factors is shown in Figure~\ref{fig:bruce-ssm-factor}. As discussed in the previous section, in the case that scan matching failed, we instead use the initial transformation from dead reckoning.
%\begin{figure}[h]
% 	\centering
% 	\begin{tikzpicture}
% 	\draw[thick,->] (2,0) -- (9,0);
% 	\draw[thick,-] (3,0) -- (3,0.2) node[anchor=south] {$\mathbf x_{k-N_{ssm}}$};
% 	\draw[thick,-] (4,0) -- (4,0.2);
% 	\draw[thick,-] (5,0) -- (5,0.2) node[anchor=south] {$\cdots$};
% 	\draw[thick,-] (6,0) -- (6,0.2);
% 	\draw[thick,-] (7,0) -- (7,0.2) node[anchor=south] {$\mathbf x_{k-1}$};
% 	\draw[thick,-] (8,0) -- (8,0.2) node[anchor=south] {$\mathbf x_{k}$};
% 	\draw[thick,decorate,decoration={brace,amplitude=10pt}] (3,0.8)  -- (7,0.8) node[midway,yshift=0.6cm] {$\mathcal P_t$}; 
% 	\draw[thick,decorate,decoration={brace,amplitude=10pt}] (7.5,0.8)  -- (8.5,0.8) node[midway,yshift=0.6cm] {$\mathcal P_s$};
% 	\draw[thick,->] (7,0) -- (7,-0.4) node[anchor=north] {$\mathbf x_t$};
% 	\draw[thick,->] (8,0) -- (8,-0.4) node[anchor=north] {$\mathbf x_s$};
% 	\end{tikzpicture}
% 	\caption{Building sequential factors from scan matching.}
% 	\label{fig:bruce-ssm-factor}
%\end{figure}
%\subsubsection{Non-sequential Scan Matching}
Non-sequential scan matching
%, or loop closure detection in Figure~\ref{fig:bruce-slam-diagram}, 
follows the same procedure as that in sequential matching, except that target points correspond to scans measured at earlier moments in time. 
%The connection results from recognizing features that have been observed before. Non-sequential factors serve to eliminate the drift from dead reckoning or sequential scan matching.
%Instead of using source points only from the current keyframe, source points are constructed in a similar way to that in sequential scan matching.
% as follows,
% \begin{equation}
% \mathcal P_s := \bigcup_{i=1}^{N_{\text{nssm}}} \mathcal T_{k, k-i} \mathcal P_{k-i}.
% \end{equation}
We limit the search window for target poses to those outside a designated minimum time interval from the current pose, imposing the condition $k - t_n > N_{\text{sep}}$, 
where $t_n$ is the most recent keyframe among the target pose candidates.
All eligible keyframes undergo global ICP initialization (Eq. (\ref{eq:bruce-icp-init-1})), and upon completing this initialization step, keyframes that do not overlap with the set of source points are discarded. The keyframe offering the maximum overlap with the set of source points is adopted as the target pose. 

Although a few verification methods are discussed, avoiding erroneous loop closure detection merits additional countermeasures. 
%The rejection of such outliers could only be achieved with the assistance of other constraints. 
We observe that correct loop closure constraints should be consistent (defined below) with the current pose estimate and potentially with future loop closure constraints. Therefore, assuming outliers occur with low probability, a set of consistent measurements is likely to be correct.

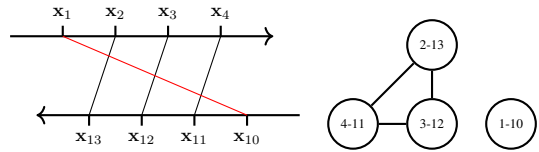
\begin{figure}[t]
	\centering
	\begin{tikzpicture}[scale=0.7, every node/.style={scale=0.7}]
	\draw[thick,->] (-0.5,1.5) -- (4.5,1.5);
	\draw[thick,-] (0.5,1.5) -- (0.5,1.7) node[anchor=south] {$\mathbf x_{1}$};
	\draw[thick,-] (1.5,1.5) -- (1.5,1.7) node[anchor=south] {$\mathbf x_{2}$};
	\draw[thick,-] (2.5,1.5) -- (2.5,1.7) node[anchor=south] {$\mathbf x_{3}$};
	\draw[thick,-] (3.5,1.5) -- (3.5,1.7) node[anchor=south] {$\mathbf x_{4}$};
	\draw[thick,->] (5,0) -- (0,0);
	\draw[thick,-] (1,0) -- (1,-0.2) node[anchor=north] {$\mathbf x_{13}$};
	\draw[thick,-] (2,0) -- (2,-0.2) node[anchor=north] {$\mathbf x_{12}$};
	\draw[thick,-] (3,0) -- (3,-0.2) node[anchor=north] {$\mathbf x_{11}$};
	\draw[thick,-] (4,0) -- (4,-0.2) node[anchor=north] {$\mathbf x_{10}$};
	\draw[-] (1,0) -- (1.5,1.5);
	\draw[-] (2,0) -- (2.5,1.5);
	\draw[-] (3,0) -- (3.5,1.5);
	\draw[red,-] (4,0) -- (0.5,1.5);
	\end{tikzpicture}\quad
	\begin{tikzpicture}[scale=0.7, every node/.style={scale=0.1}]
	\begin{scope}[every node/.style={circle,thick,draw}]
	\node (A) at (0,0) {\tiny{4-11}};
	\node (B) at (1.5,0) {\tiny{3-12}};
	\node (C) at (1.5,1.5) {\tiny{2-13}};
	\node (D) at (3,0) {\tiny{1-10}};
	\end{scope}
	\path [thick,-] (A) edge (B);
	\path [thick,-] (A) edge (C);
	\path [thick,-] (B) edge (C);
	\end{tikzpicture}
	\caption[PCM outlier rejection.]{Four pairs of loop closures are detected as constraints between 2 keyframes. The erroneous one (red) is excluded in the max. clique.}
	\label{fig:bruce-pcm-clique}
\end{figure}

We adopt the pairwise consistent measurement set maximization (PCM) \cite{mangelson2018pairwise} technique to reject outliers. 
Let $\hat {\mathbf x}_{jl}$ and $\hat {\mathbf x}_{ik}$ be a pair of loop closure measurements and we assume $l < k$. An estimate of $\hat {\mathbf x}_{ik}$ can be obtained using estimated poses and measurement $\hat {\mathbf x}_{jl}$ as follows
\begin{equation}
\hat {\mathbf x}_{ik}' = \mathbf x_{ij} \oplus \hat {\mathbf x}_{jl} \oplus \mathbf x_{lk}.
\end{equation}
%The flow of relative transformations is illustrated as the long arrow in Figure~\ref{fig:bruce-pcm}. 
The two constraints are determined to be pairwise consistent if the following condition is satisfied,
\begin{equation}
\Vert \hat{\mathbf x}_{ik} \ominus \hat{ \mathbf x}_{ik}'\Vert_\Sigma \le \eta_{\text{pcm}},
\end{equation}
where $\Vert \cdot\Vert_\Sigma$ is the Mahalanobis distance, $\eta_{\text{pcm}}$ is a predefined threshold and $\Sigma$ can be obtained from ICP covariance $\hat{\Sigma}_{ik}$.
Given a set of measurements, a mutual consistency check is performed. %The outcome forms an undirected graph with nodes representing pairs of measurements and edges representing pairwise consistency. 
An example with four measurements is visualized in Figure~\ref{fig:bruce-pcm-clique}, where only three out of six possible pairs of loop closure measurements are pairwise consistent.
The largest subset of pairwise consistent measurements, where every pair in the subset is pairwise consistent, can be identified as a maximum clique in a graph. 
%Although finding the maximum clique is expensive and hard to approximate, fortunately the graph is typically small enough that exhaustively searching for the maximal clique is achievable in real-time experiments.
%In implementation, we maintain a queue of loop closure measurements by arrival time. At each keyframe the PCM is executed to find the maximum clique. If the clique number is larger than $N_{\text{pcm}}$, all measurements in the clique are regarded as correct and those which haven't been added to factor graph are added. In Figure~\ref{fig:bruce-pcm-clique}, if $N_{\text{pcm}} = 3$, three correct measurements will be accepted until $\mathbf x_{13}$.

\subsection{Occupancy Mapping}
\label{sec:occupancy_mapping}
Our occupancy mapping framework is based on the occupancy grid mapping algorithm \cite{Elfes1989}. The entire map is uniformly discretized into independent grid cells and the occupancy probability of each cell is updated recursively using a Bayes filter. However our state estimation from SLAM exhibits severe drift, and at the moment the drift is corrected by loop closures, it is desirable to correct the occupancy mapping error using the updated trajectory. Therefore, we employ a submap-based occupancy mapping algorithm \cite{ho2018virtual}, which builds a local map anchored at each keyframe. Using this approach, we are able to achieve efficient map recomputation when a segment of the trajectory is changed.

\subsubsection{Merging Submaps}

\begin{figure}[t]
	\centering
	\subfloat[Raw sonar image (color indicates pixel intensity)]{\includegraphics[width=0.4\columnwidth]{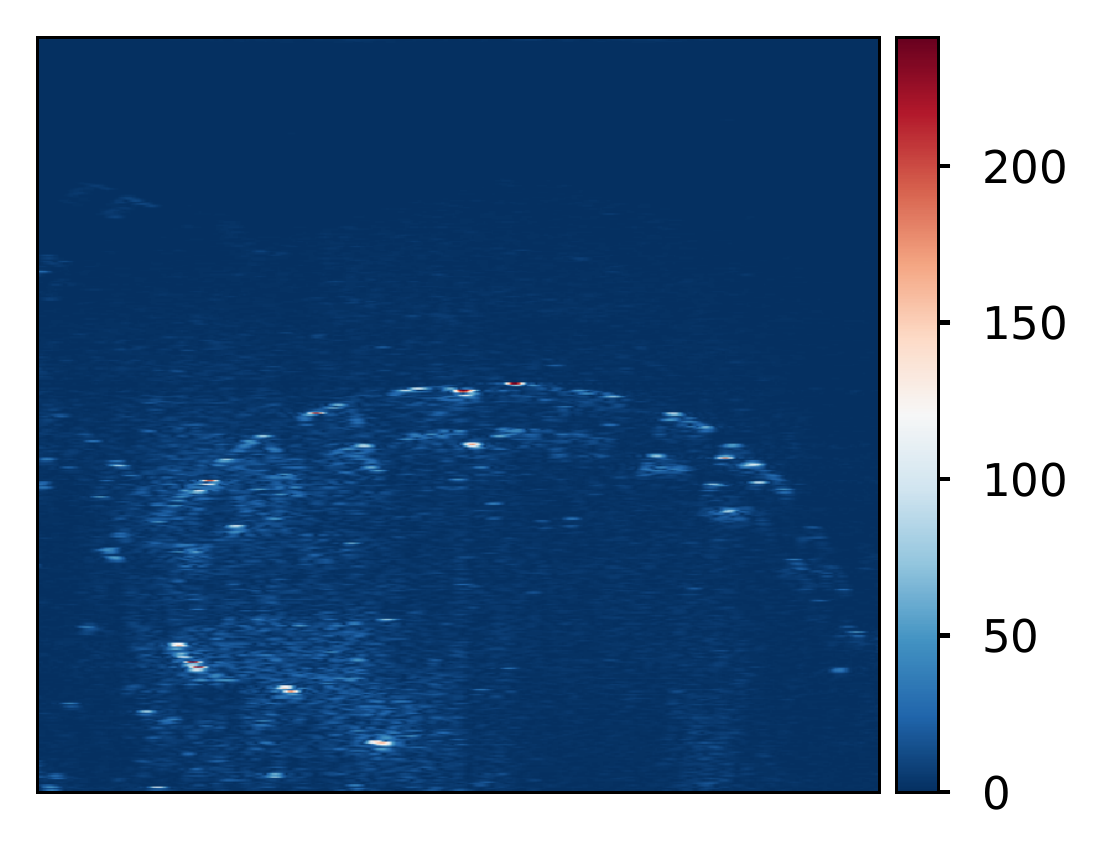}}\
	\subfloat[Submap in polar coordinates (occupancy prob. shown)]{\includegraphics[width=0.4\columnwidth]{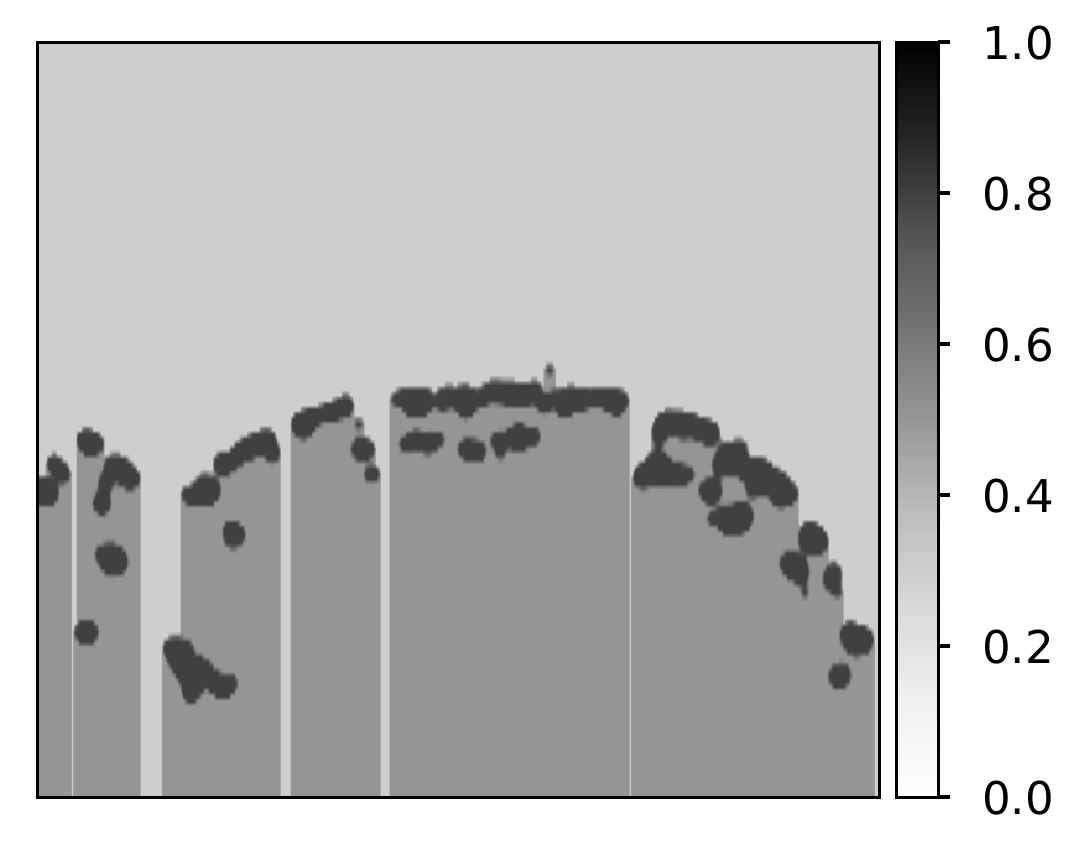}}
	\caption{Inverse sensor model and the submap from one sonar image. Range from the sensor increases from the top of the image down.}
	\label{fig:bruce-sensor-model}
\end{figure}

\begin{figure*}
    \centering
    \subfloat[Nearest frontier]{\includegraphics[width=0.40\textwidth]{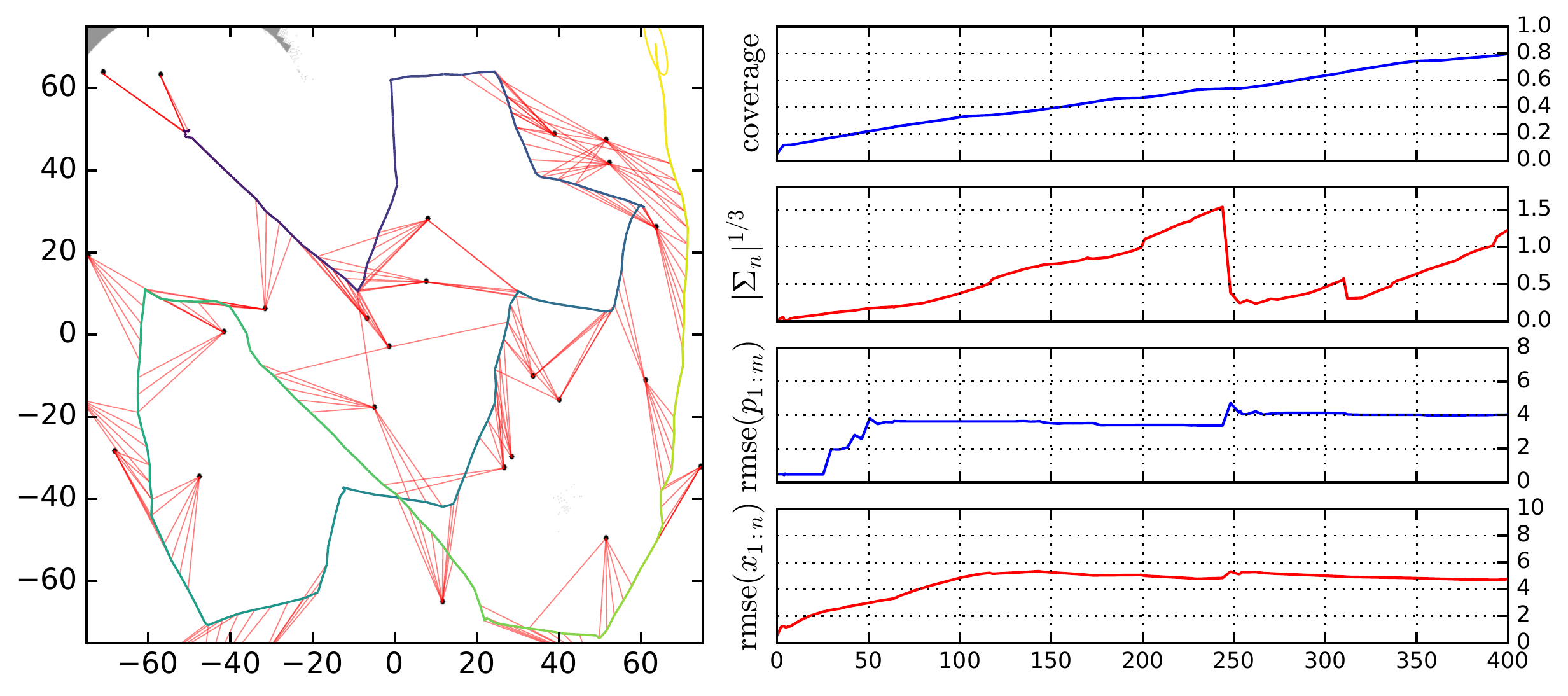}}\quad
    \subfloat[Next-best-view]{\includegraphics[width=0.40\textwidth]{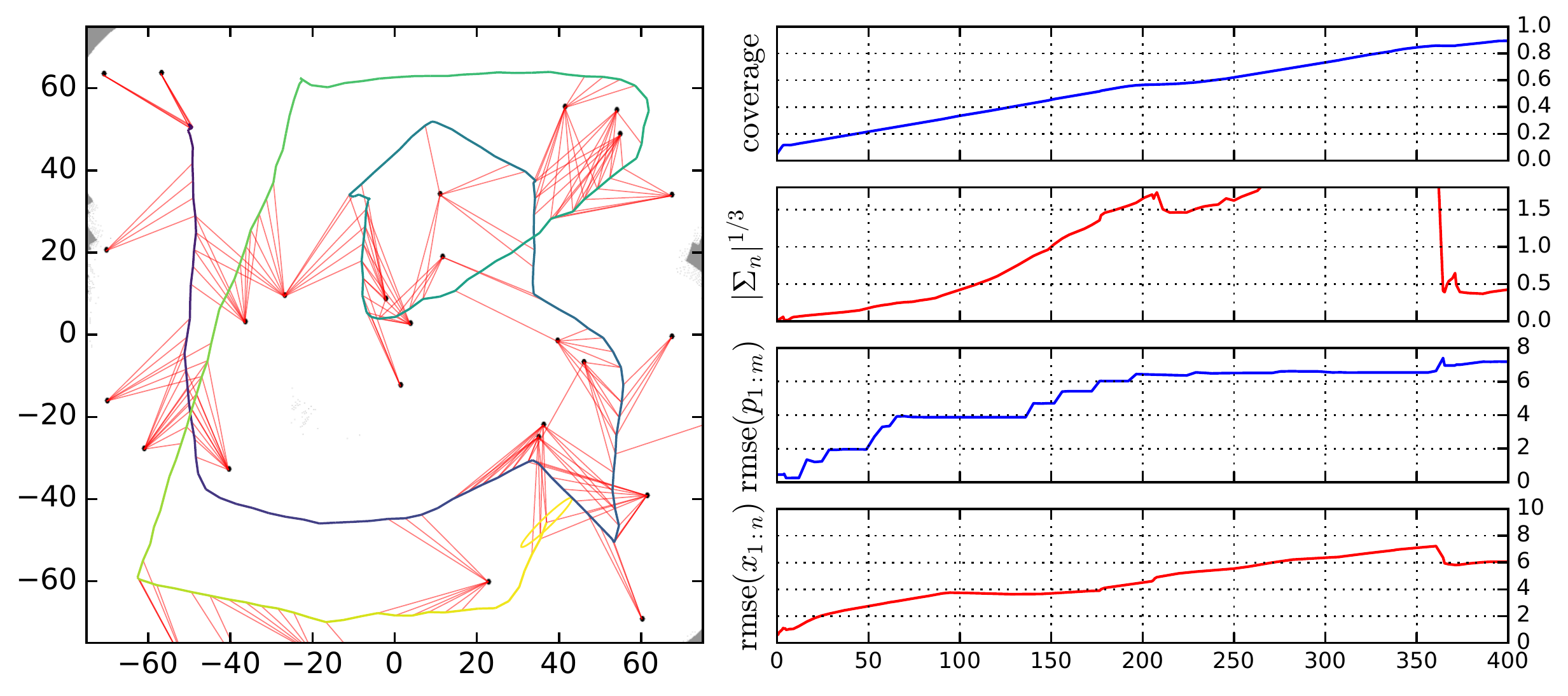}}\\
    \subfloat[Heuristic]{\includegraphics[width=0.40\textwidth]{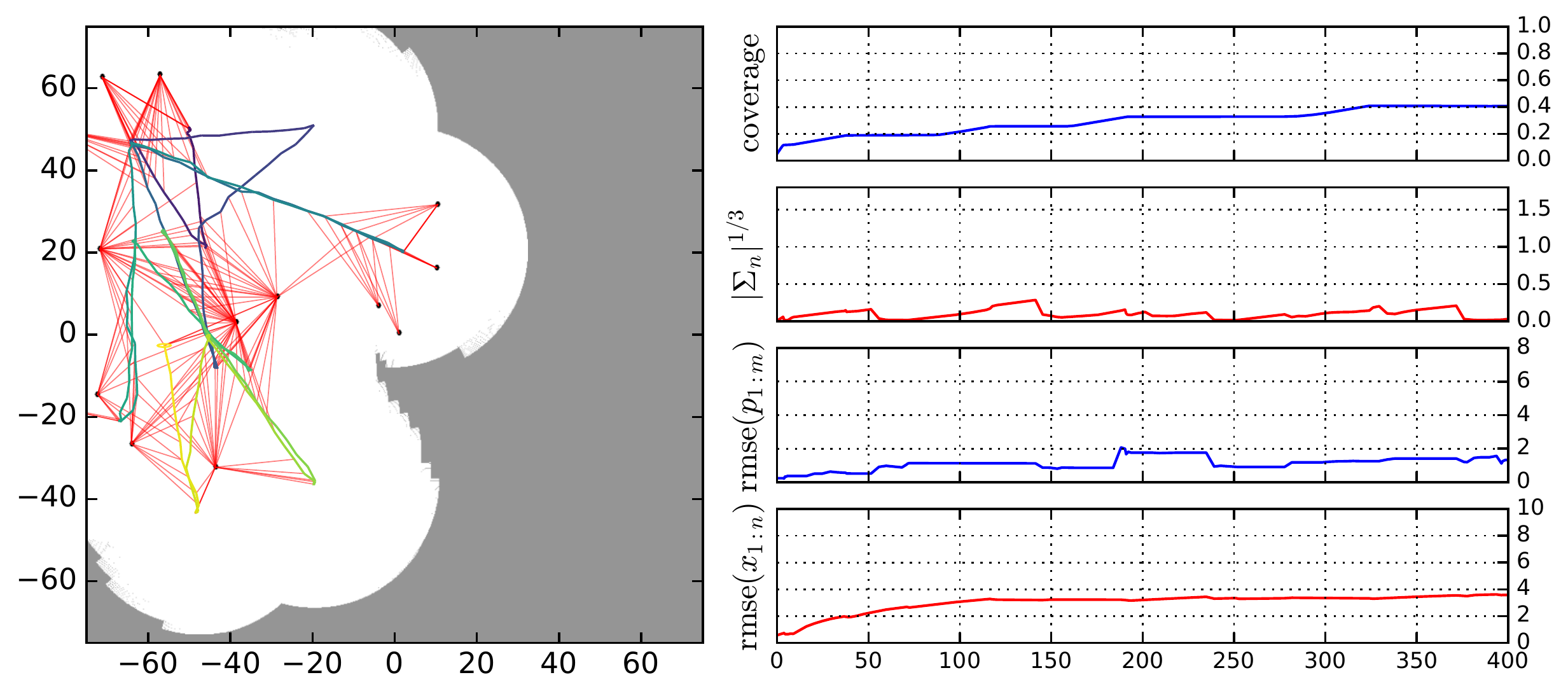}}\quad
    \subfloat[EM]{\includegraphics[width=0.40\textwidth]{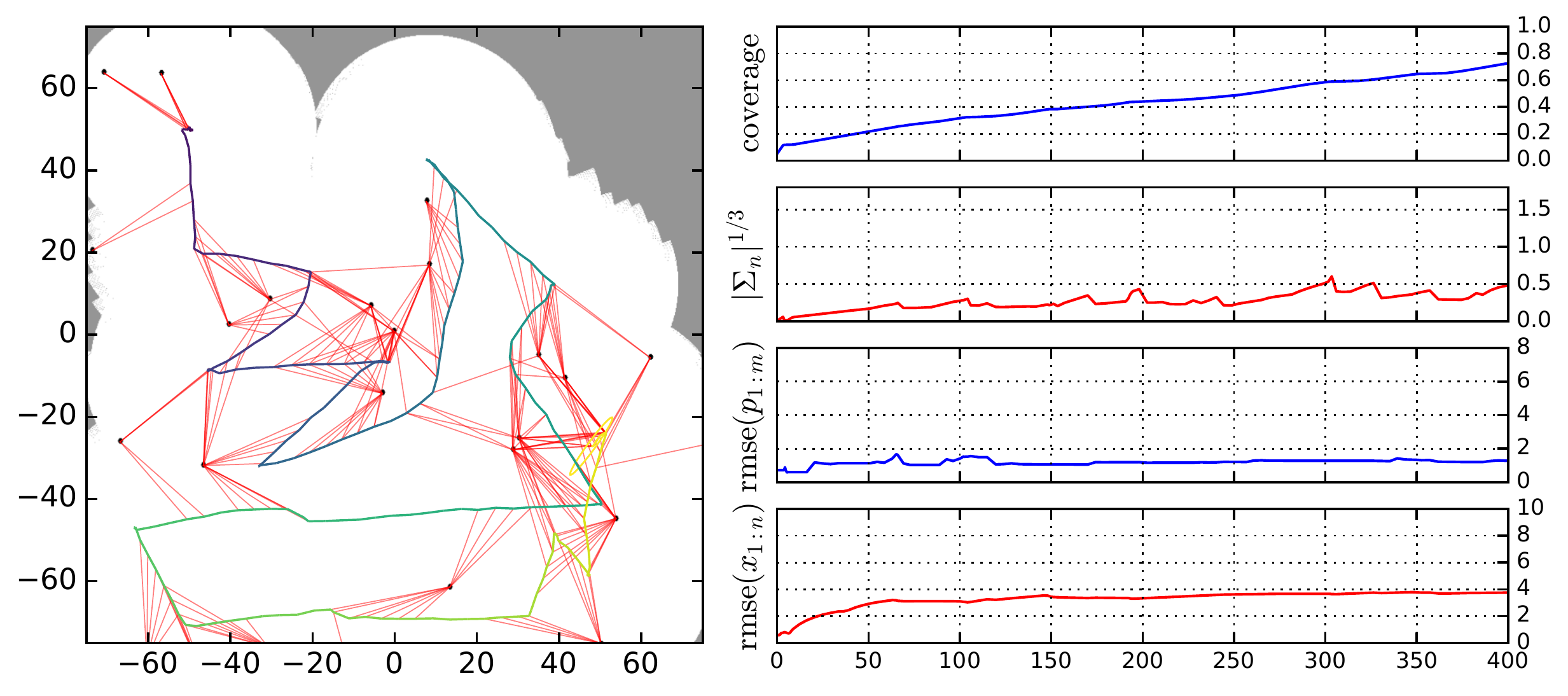}}
    \caption{Landmark SLAM simulation examples using NF, NBV, Heuristic and EM algorithms in the environment depicted in Fig. \ref{fig:bruce-sim-env} (a). Each quadrant shows results from a representative execution trace. In each quadrant at left, the occupancy map with pose history and landmark measurement constraints is shown. At right, a plot of map coverage, pose uncertainty, map error, and pose error are shown vs. travel distance.}
    \label{fig:landmarks_example}
\end{figure*}

\begin{figure*}
    \centering
    \subfloat[Pose uncertainty]{\includegraphics[width=0.23\textwidth]{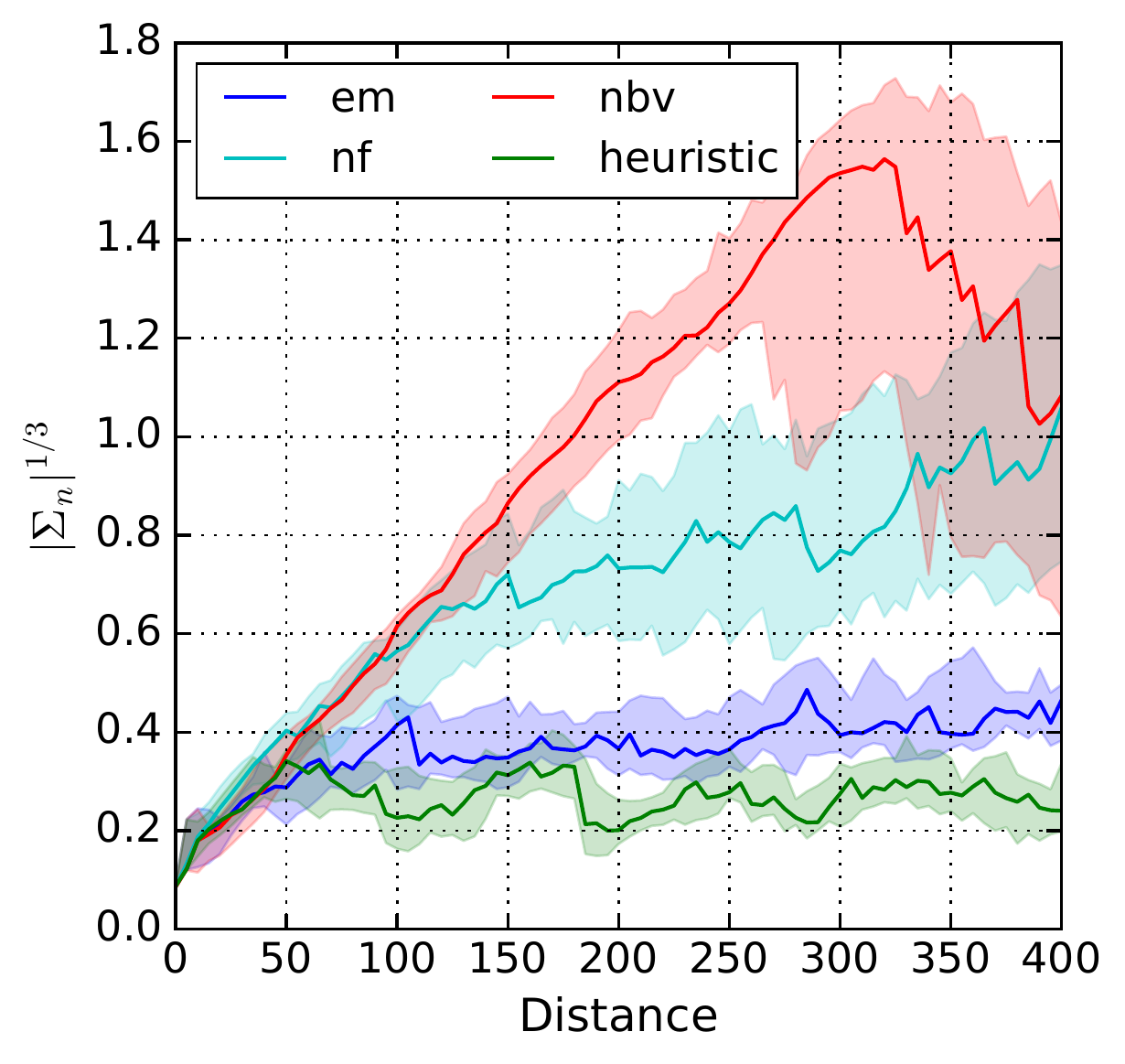}}\quad
    \subfloat[Trajectory error]{\includegraphics[width=0.23\textwidth]{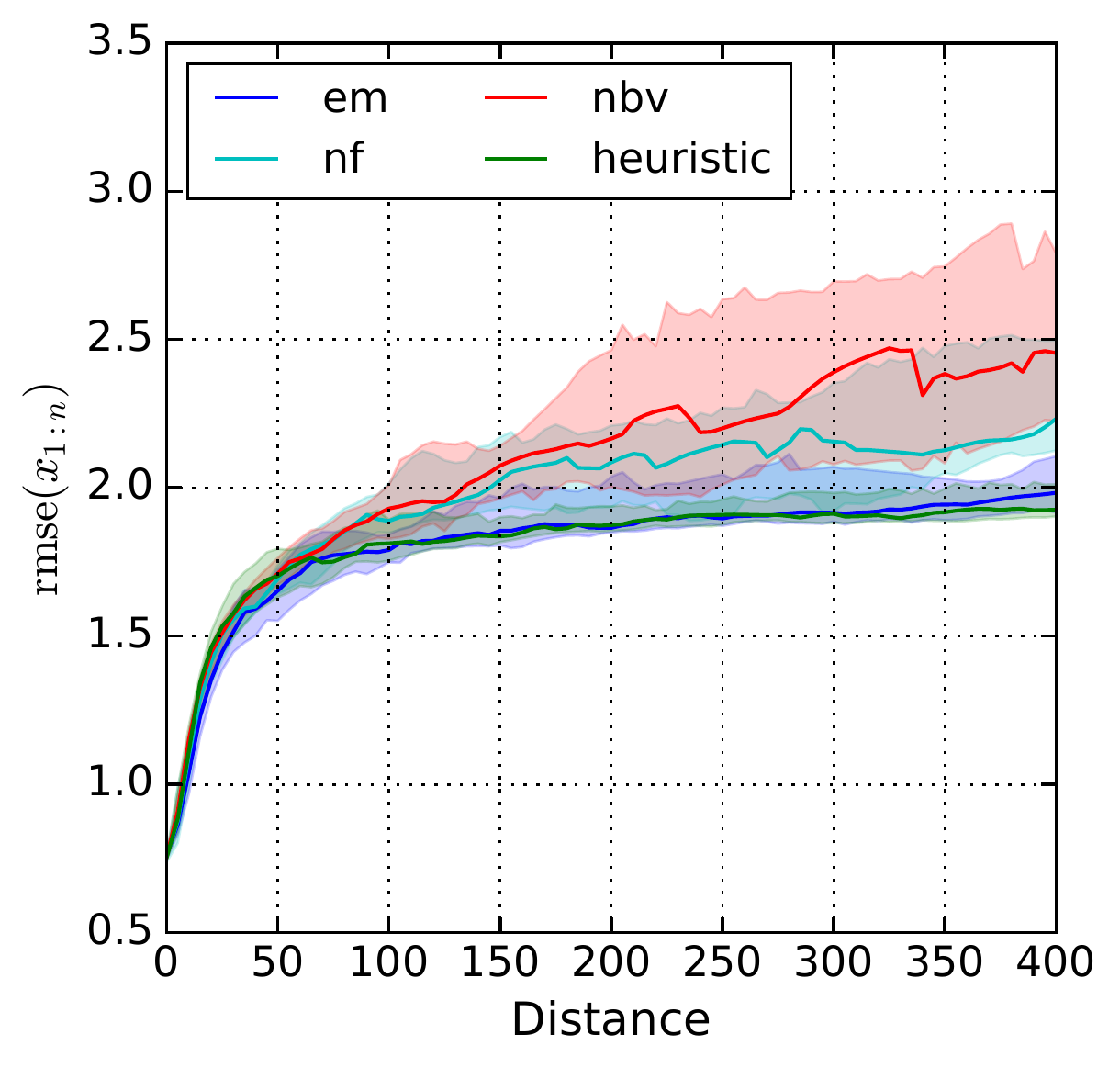}}\quad
    \subfloat[Map error]{\includegraphics[width=0.23\textwidth]{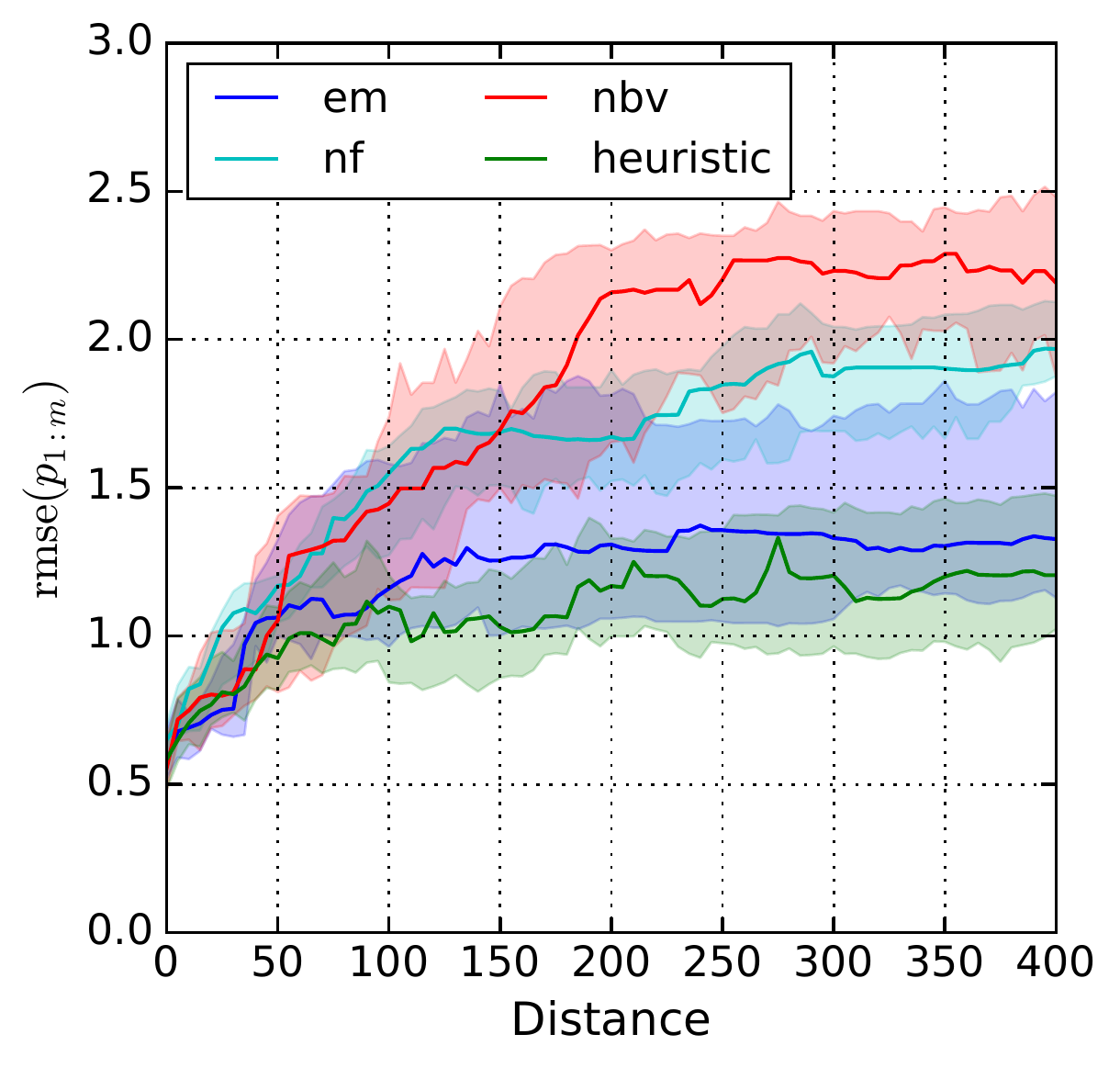}}\quad
    \subfloat[Map coverage]{\includegraphics[width=0.23\textwidth]{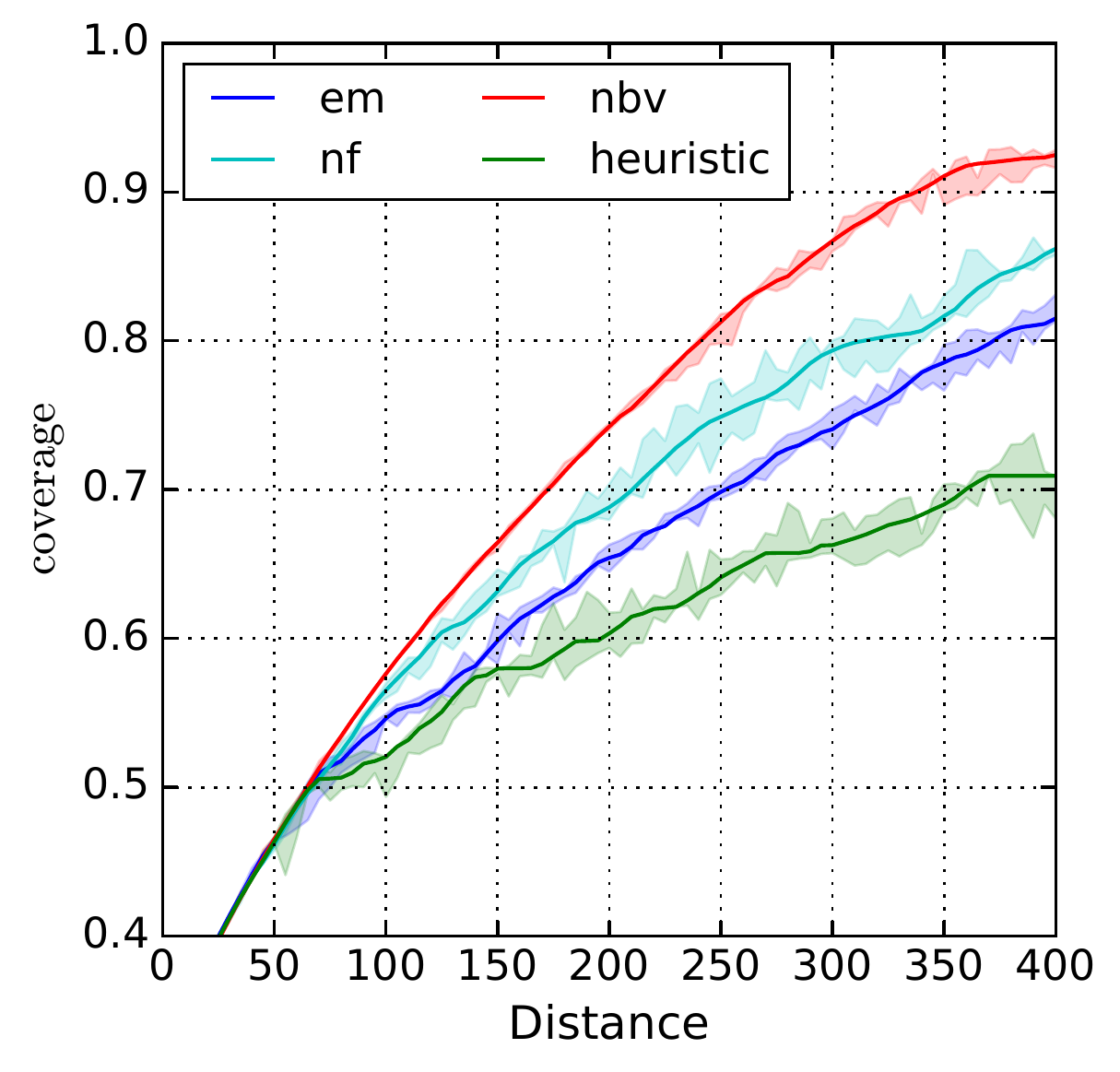}}
    \caption{Exploration performance mean values in the landmark environment of Fig. \ref{fig:bruce-sim-env} (a) for 45 trials (five for each robot start location depicted in Fig. \ref{fig:bruce-sim-env}), with 95 percent confidence intervals shown. Values along the x-axis denote the robot's distance traveled.
    %These are the results I have right now. They are not the best because the heuristic could have better map coverage if we increase its uncertainty threshold such that it can achieve similar pose uncertainty curve to that from em.
    }
    \label{fig:landmarks_performance}
\end{figure*}

% \begin{figure*}
% 	\centering
% 	\subfloat[Step 10]{\includegraphics[width=0.3\textwidth]{bruce/submaps-10}}\
% 	\subfloat[Step 26]{\includegraphics[width=0.3\textwidth]{bruce/submaps-26}}\
% 	\subfloat[Step 42]{\includegraphics[width=0.3\textwidth]{bruce/submaps-42}}\\
% 	\subfloat[Step 58]{\includegraphics[width=0.3\textwidth]{bruce/submaps-58}}\
% 	\subfloat[Step 74]{\includegraphics[width=0.3\textwidth]{bruce/submaps-74}}\
% 	\subfloat[Step 90]{\includegraphics[width=0.3\textwidth]{bruce/submaps-90}}\\
% 	\caption{Underwater occupancy grid mapping. \textcolor{red}{BE: Should this be included in the Occupancy Mapping subsection, or is it redundant?}\Comment{YH: We could put result beefore & after loop closure in Occupancy Mapping Subsection}}
% 	\label{fig:bruce-occ-mapping}
% \end{figure*}

Suppose we are able to obtain keyframe poses in a reference frame from SLAM denoted as $\mathcal X = \{ \mathbf x_k \}_{k=1}^K$. Given the sonar image and extracted features, we can build a submap located at a local frame denoted $\mathcal S_k = \{ \mathbf m^k_i \}$, where $\mathbf m^k_i \in \mathbb R^2$ represents a 2D grid cell in the $k$-th keyframe. Let $p(\mathbf m_i^k = 1)$ represent the probability of the cell being occupied and $l^k(\mathbf m_i^k) = \log \frac{p(\mathbf m_i^k = 1)}{1 - p(\mathbf m_i^k = 1)}$ represents its log-odds notation. We use superscript $k$ to denote cells or log-odds values in the $k$-th keyframe. The entire map is represented as the composition of submaps $\mathcal M = \{\mathcal S_k\}_{k=1}^K$.

In this paper, a submap only consists of one sonar frame, thus the occupancy values can be calculated from the inverse sensor model directly. Unlike laser beams, sonar is capable of detecting targets that are hidden behind the first object along a sensor beam due to the wide vertical aperture. This phenomenon is evident in our experimental results, as there exist floating docks that do not block sonar beams entirely. Similar to the inverse sensor model for laser beams, where cells before, around and behind a \textit{hit point} are modeled as \textit{free}, \textit{occupied} and \textit{unknown}, respectively, we also treat hit points behind the first one as \textit{occupied}. In essence all detected targets are assumed to be occupied, but only the area in front of the first target is assumed to be obstacle free. The inverse sensor model is shown in Figure~\ref{fig:bruce-sensor-model}; Gaussian convolution is used to smooth the occupied probability around hit points.

Given multiple keyframe poses and their corresponding submaps $\mathcal X, \mathcal M$, we wish to produce an occupancy map $\{m_i\}$ in the global frame. Its cells' occupancy probabilities incorporate the measurements from different poses, which can be incrementally updated through the Bayes filter as
\begin{equation}
l(\mathbf m_i) = \sum_{\mathcal S^k \in \mathcal M}  l^k(\mathcal T_{kg}\mathbf m_i),
\end{equation}
where $\mathcal T_{kg}\mathbf m_i$ represents the action of transforming global grid cells to a local frame. We ignore the prior in the summation as by default we assume an uninformative prior for the occupancy map $p(\mathbf m_i = 1) = 0.5$. Therefore, the resulting occupancy log-odds is simply the summation of log-odds across the various individual submaps.

\begin{table*}
    \centering
\begin{tabular}{lrrrrrrrrr}
	\toprule
	\hline
	alg/distance &    0 &   50 &  100 &  150 &  200 &  250 & 300 &  350 &  400 \\
	\hline
	\midrule
	EM & 0.10 & 0.34 & 0.30 & 0.32 & 0.33 & 0.31 & 0.31 & 0.29 & 0.28 \\
	NF & 0.10 & 0.33 & 0.44 & 0.43 & 0.46 & 0.45 & 0.40 & 0.39 & 0.40 \\
	NBV & 0.10 & 0.35 & 0.53 & 0.57 & 0.53 & 0.45 & 0.36 & 0.36 & 0.35 \\
	Heuristic & 0.10 & 0.34 & 0.32 & 0.30 & 0.31 & 0.30 & 0.31 & 0.28 & 0.31 \\
	\hline
	\bottomrule
\end{tabular}
\caption{Pose uncertainty in the simulated marina environment.}
\label{table:sim pose uncertainty}
\vspace{-2mm}
\end{table*}

\begin{table*}
	\centering
\begin{tabular}{lrrrrrrrrr}
	\toprule
	\hline
	alg/distance &    0 &   50 &  100 &  150 &  200 &  250 &  300 &  350 &  400 \\
	\hline
	\midrule
	EM & 0.75 & 1.72 & 1.83 & 1.89 & 1.90 & 1.93 & 1.93 & 1.96 & 1.97 \\
	 NF & 0.75 & 1.79 & 1.94 & 1.97 & 2.06 & 2.07 & 2.08 & 2.07 & 2.08 \\
	NBV & 0.75 & 1.73 & 1.90 & 1.99 & 2.07 & 2.05 & 2.05 & 2.05 & 2.05 \\
	Heuristic & 0.75 & 1.72 & 1.84 & 1.88 & 1.91 & 1.94 & 1.95 & 1.96 & 1.98 \\
	\hline
	\bottomrule
\end{tabular}
	\caption{Pose error in the simulated marina environment.}
	\label{table:sim pose error}
	\vspace{-2mm}
\end{table*}

\begin{table*}
	\centering
	\begin{tabular}{lrrrrr}
		\toprule
		\hline
		alg/coverage &   50\% &   60\% &   70\% &    80\% &    90\% \\
		\midrule
		\hline
		EM & 13.79 & 39.53 & 80.92 & 157.15 & 303.92 \\
		NF & 14.90 & 43.63 & 88.38 & 172.30 & 268.67 \\
		NBV & 13.90 & 40.55 & 70.36 & 117.93 & 234.01 \\
		Heuristic & 14.23 & 41.41 & 80.12 & 186.63 & 396.52 \\
		\bottomrule
		\hline
	\end{tabular}
	\caption{Travel distance to achieve designated \%'s of coverage in the simulated marina environment.}
	\label{table:sim map coverage}
	\vspace{-2mm}
\end{table*}

\begin{table*}
	\centering
	\begin{tabular}{lrrrrrrrrr}
		\toprule
		\hline
		alg/distance &    0 &   50 &  100 &  150 &  200 &  250 &  300 &  350 &  400 \\
		\midrule
		\hline
		EM & 0.56 & 0.90 & 0.95 & 0.97 & 1.00 & 1.03 & 1.05 & 1.06 & 1.05 \\
	NF & 0.60 & 0.97 & 1.03 & 1.08 & 1.13 & 1.15 & 1.12 & 1.13 & 1.13 \\
		NBV & 0.58 & 0.87 & 1.01 & 1.08 & 1.12 & 1.12 & 1.12 & 1.12 & 1.12 \\
		Heuristic & 0.58 & 0.90 & 0.94 & 0.98 & 0.99 & 1.01 & 1.02 & 1.03 & 1.03 \\
		\bottomrule
		\hline
	\end{tabular}
	\caption{Map error in simulated marina environment.}
	\label{table:sim map error}
	\vspace{-2mm}
\end{table*}

\subsubsection{Updating A Submap}

Adding non-sequential factors results in frequent changes to the pose history. We update the contribution of the $k$-th keyframe when a change, including translation and rotation, exceeds a specified threshold. 
%If computational resources allow, this threshold will permit changes that move poses by the dimension of one cell or greater. 
%maximum shift due to pose change shouldn't be larger than 1 grid cell.
Because the clamping rule \cite{Hornung2013} is not used in merging submaps, the update due to pose change can be implemented efficiently by removing previous log-odds values and adding new ones,
\begin{equation}
l'(\mathbf m_i) = l(\mathbf m_i) - \underbrace{l^k(\mathcal T_{kg}\mathbf m_i)}_\text{remove old update} + \underbrace{l^k(\mathcal T'_{kg}\mathbf m_i)}_\text{add new update}.
\end{equation}
In the equation above, the local submap represented by log-odds values does not need to be recomputed.

In practice, the submap for every keyframe stores the indices of its associated grid cells in the global frame. While updating, we simply subtract log-odds values from old global indices and add updated log-odds values to the appropriate new indices. Examples of the occupancy mapping framework described here are shown throughout the experimental results below. % of Section \ref{sec:results}. %in Figure~\ref{fig:bruce-occ-mapping}.

\begin{figure}
	\centering
	\subfloat[Landmark SLAM]{\includegraphics[width=0.45\columnwidth]{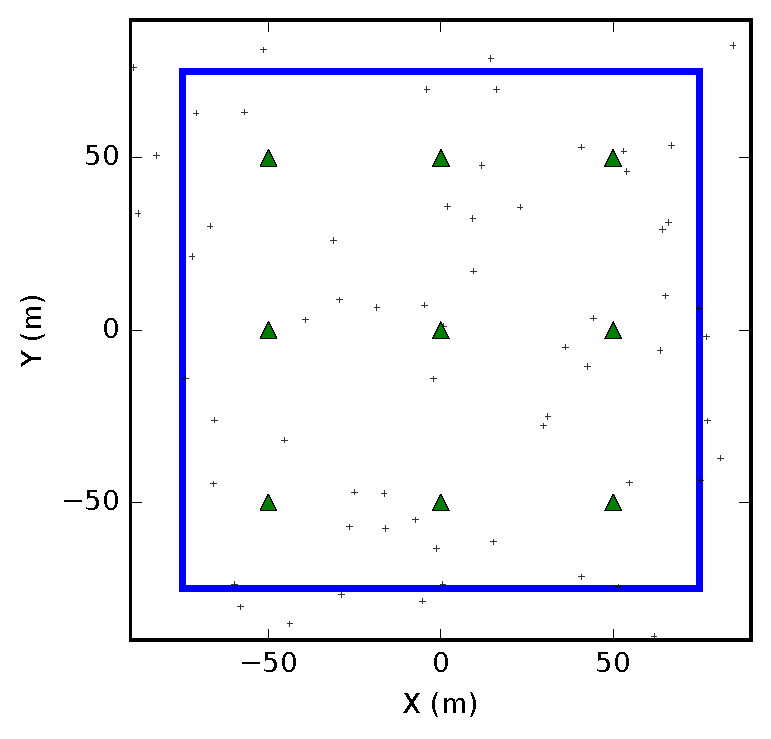}\label{sim:landmark}}
	\subfloat[Pose SLAM]{\includegraphics[width=0.45\columnwidth]{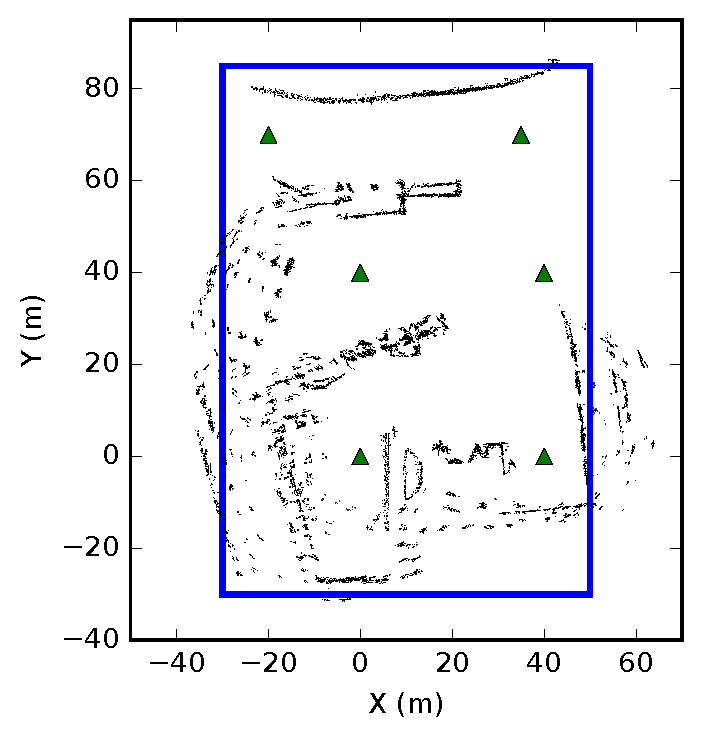}\label{sim:pose}}
	\caption{Simulated exploration environments for landmark and pose SLAM. The exploration bounding box is denoted by a blue rectangle, and exploration is initialized from the green triangles. The robot may only travel within the bounding box, and only observations of the area within the bounding box are used to evaluate map coverage.}
	\label{fig:bruce-sim-env}
	\vspace{-2mm}
\end{figure}

\begin{figure*}
	\centering
	\subfloat[NF]{\includegraphics[width=0.45\textwidth]{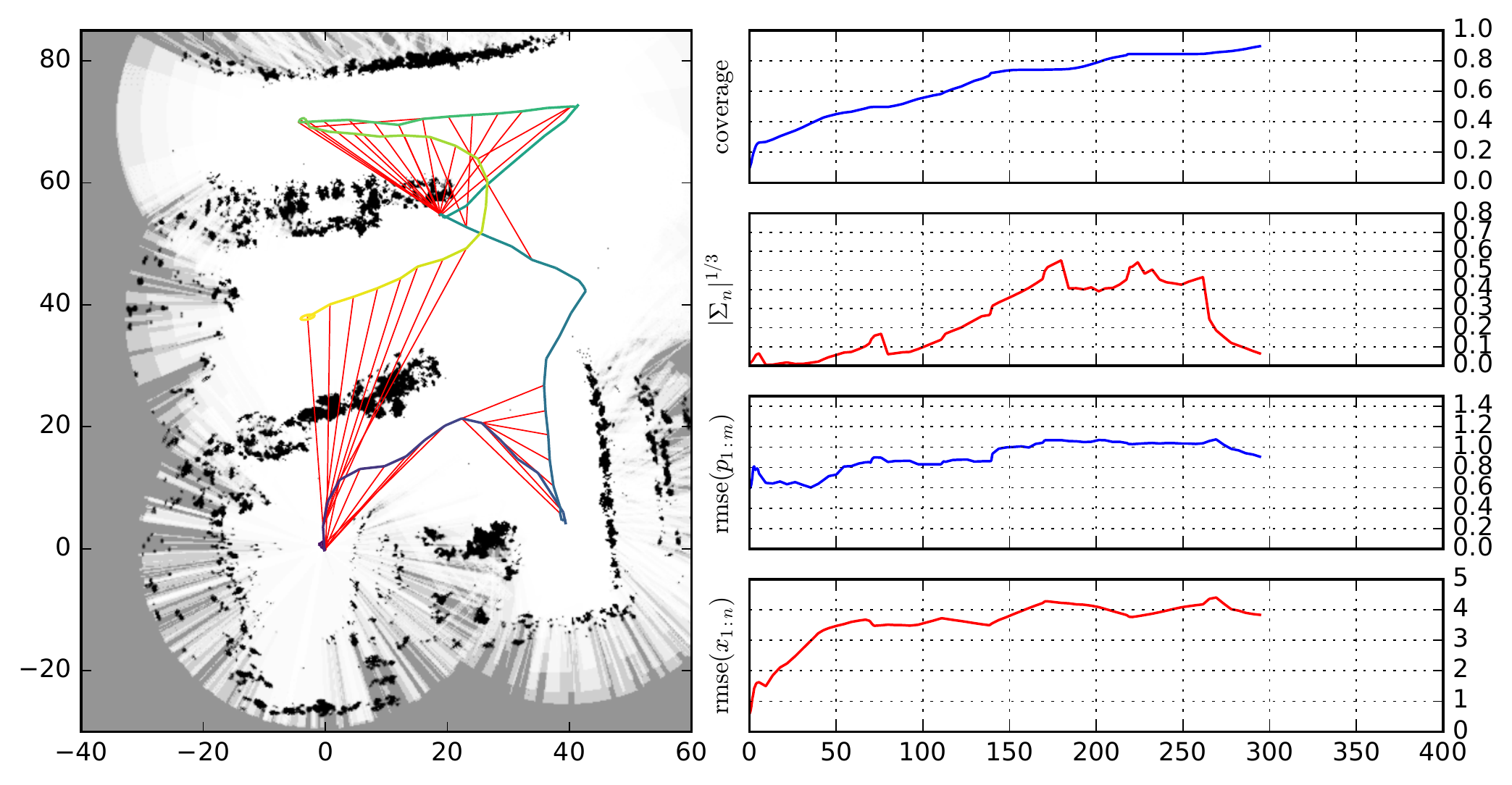}}
	\subfloat[NBV]{\includegraphics[width=0.45\textwidth]{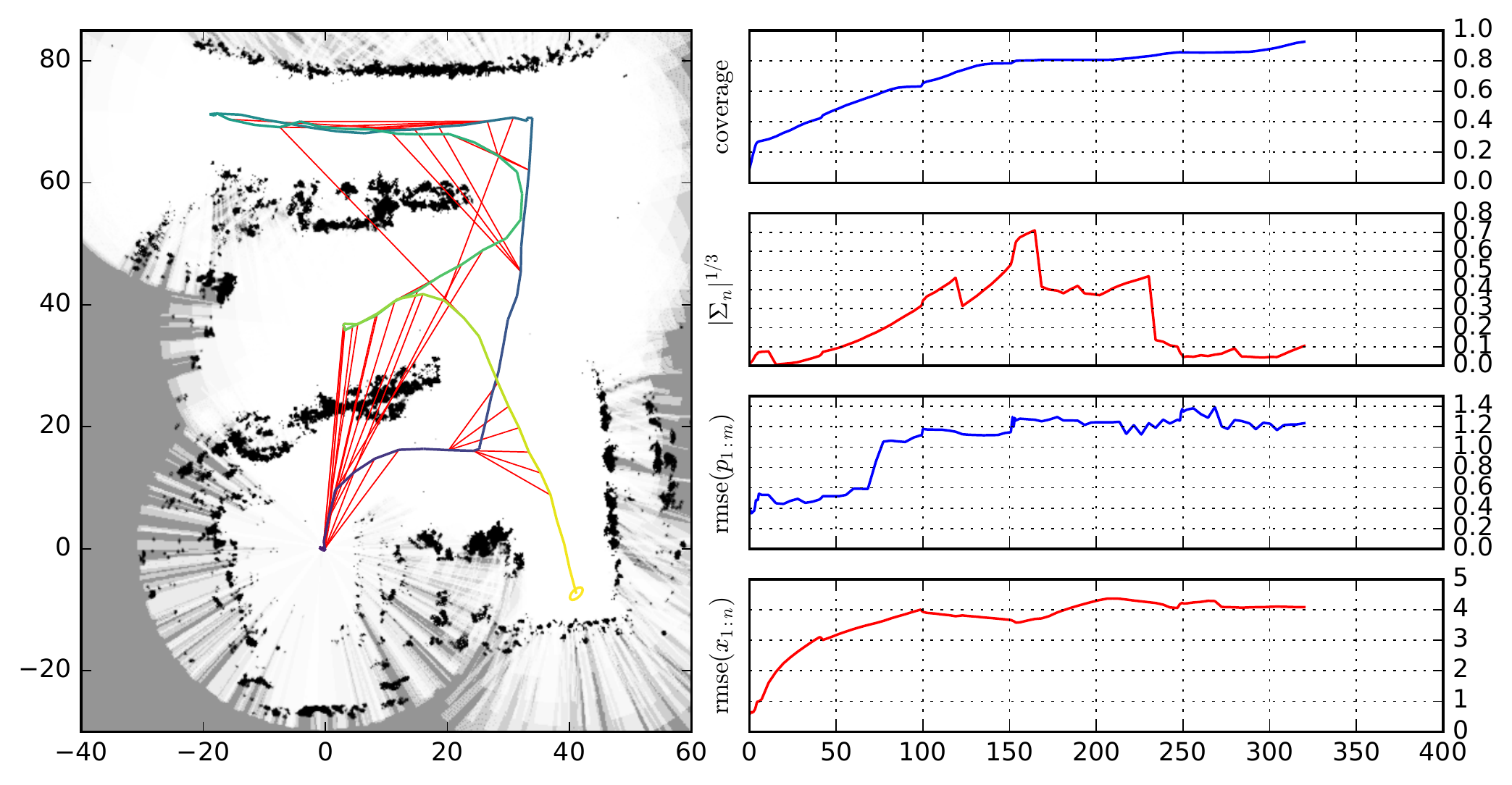}}\\
	\subfloat[Heuristic]{\includegraphics[width=0.45\textwidth]{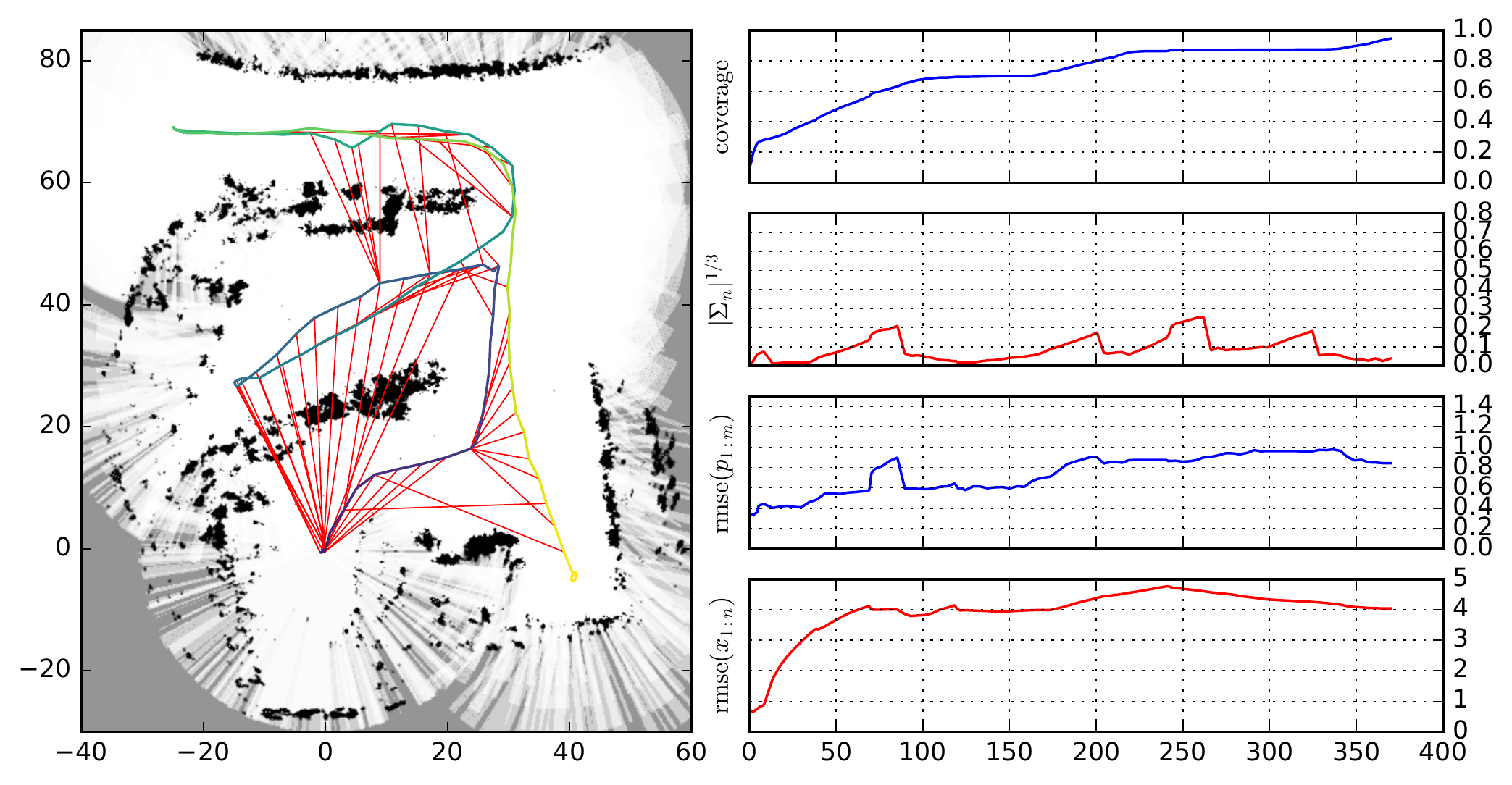}}
	\subfloat[EM]{\includegraphics[width=0.45\textwidth]{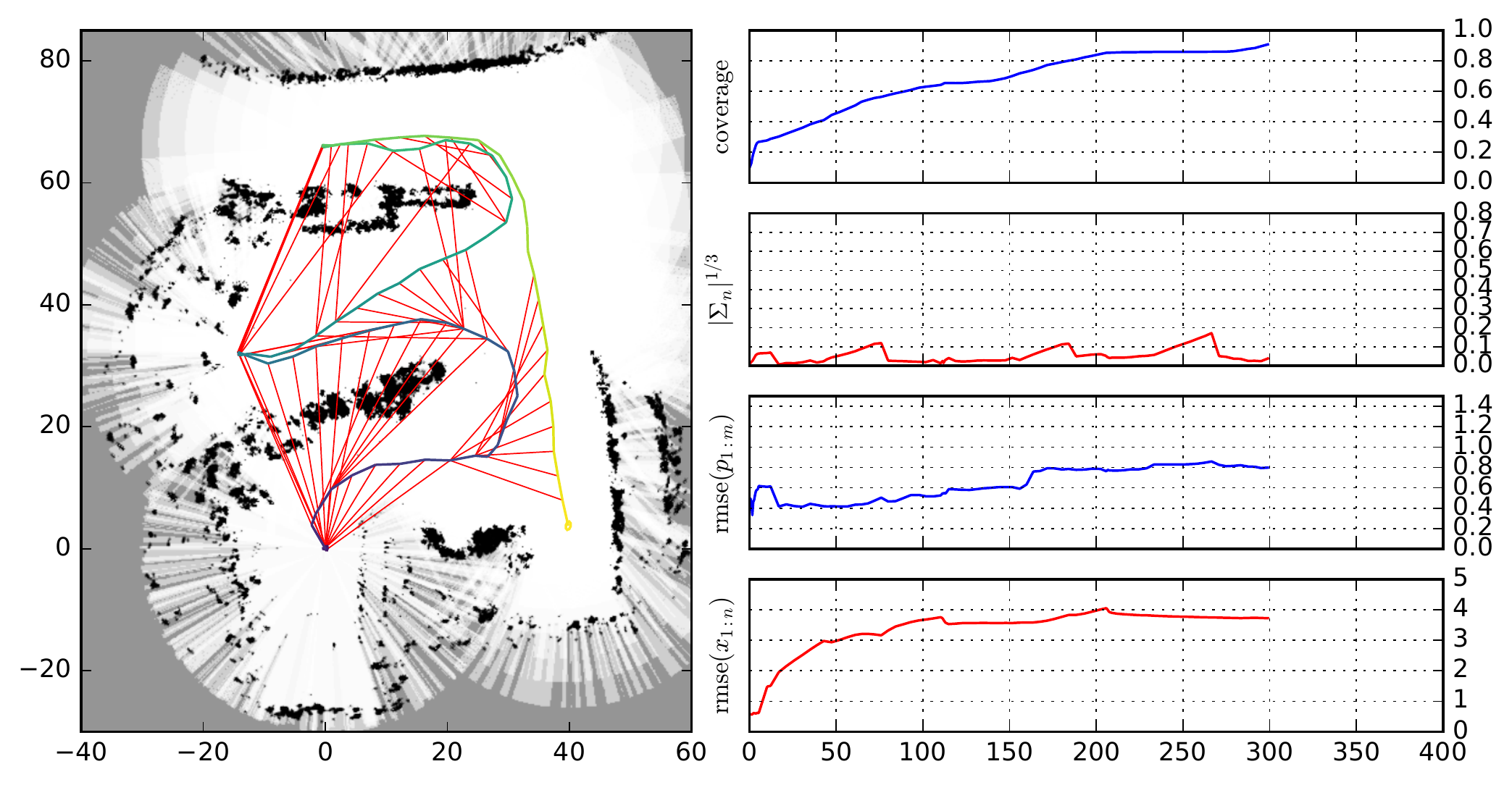}}
	\caption{Pose SLAM simulation examples using NF, NBV, Heuristic and EM algorithms in the marina environment depicted in Fig. \ref{fig:bruce-sim-env} (b). Each quadrant shows results from a representative execution trace. In each quadrant at left, the occupancy map with pose history and loop closure constraints is shown. At right, a plot of map coverage, pose uncertainty, map error, and pose error are shown vs. travel distance.}
	\label{fig:bruce-sim-example}
\end{figure*}

\begin{figure*}
    \centering
    \subfloat[Pose uncertainty]{\includegraphics[width=0.23\textwidth]{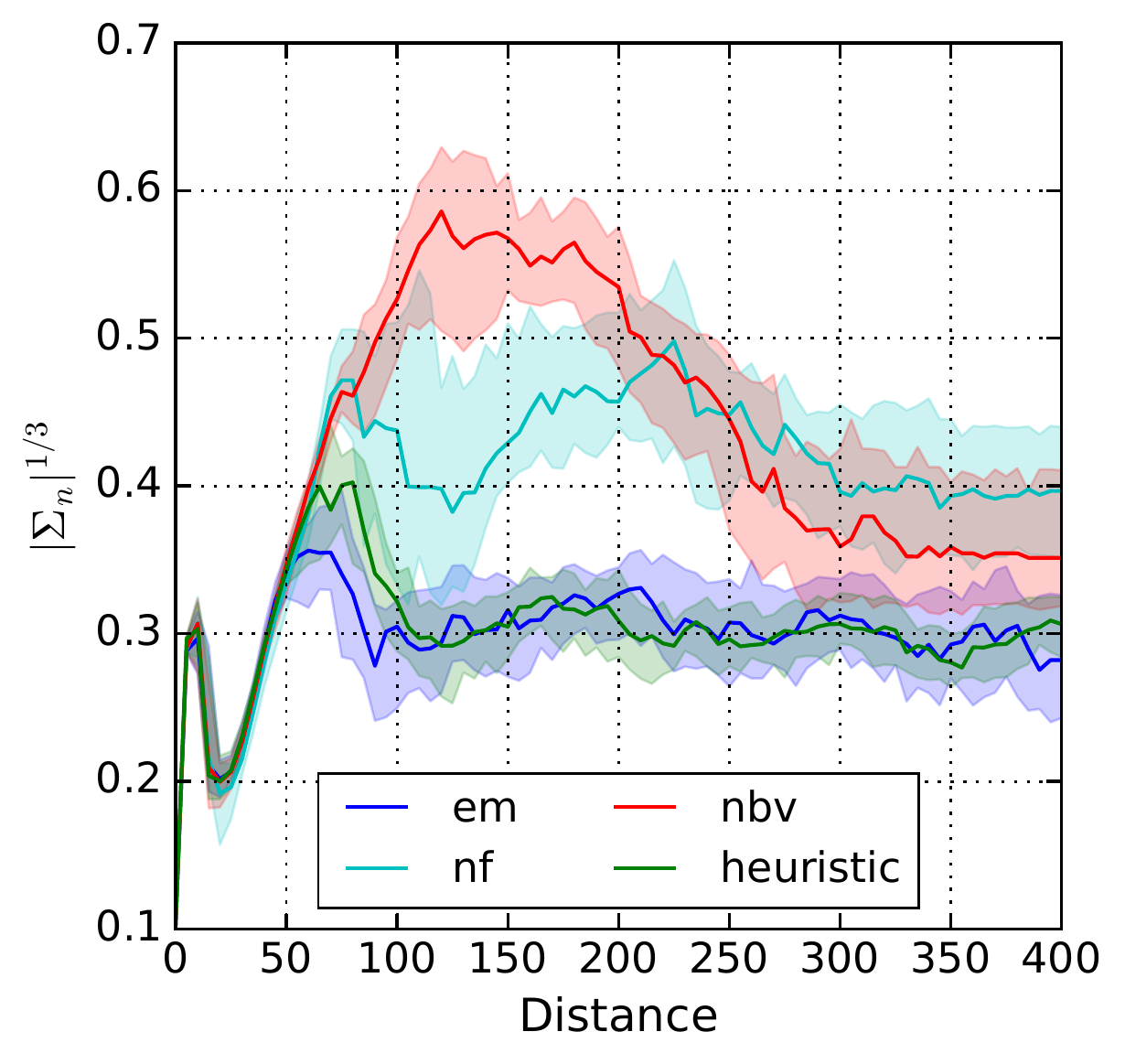}}\quad
    \subfloat[Trajectory error]{\includegraphics[width=0.23\textwidth]{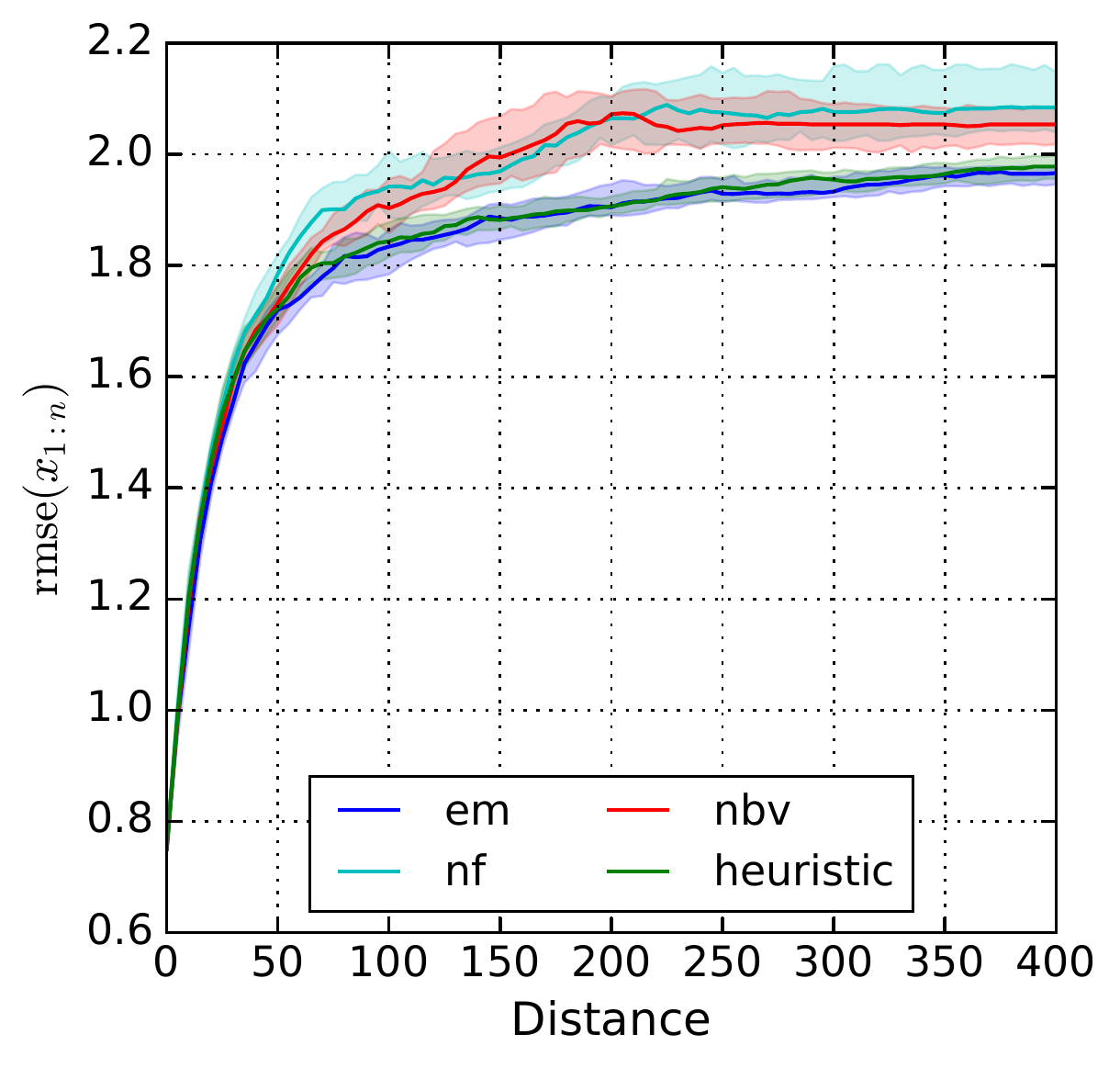}}\quad
    \subfloat[Map error]{\includegraphics[width=0.23\textwidth]{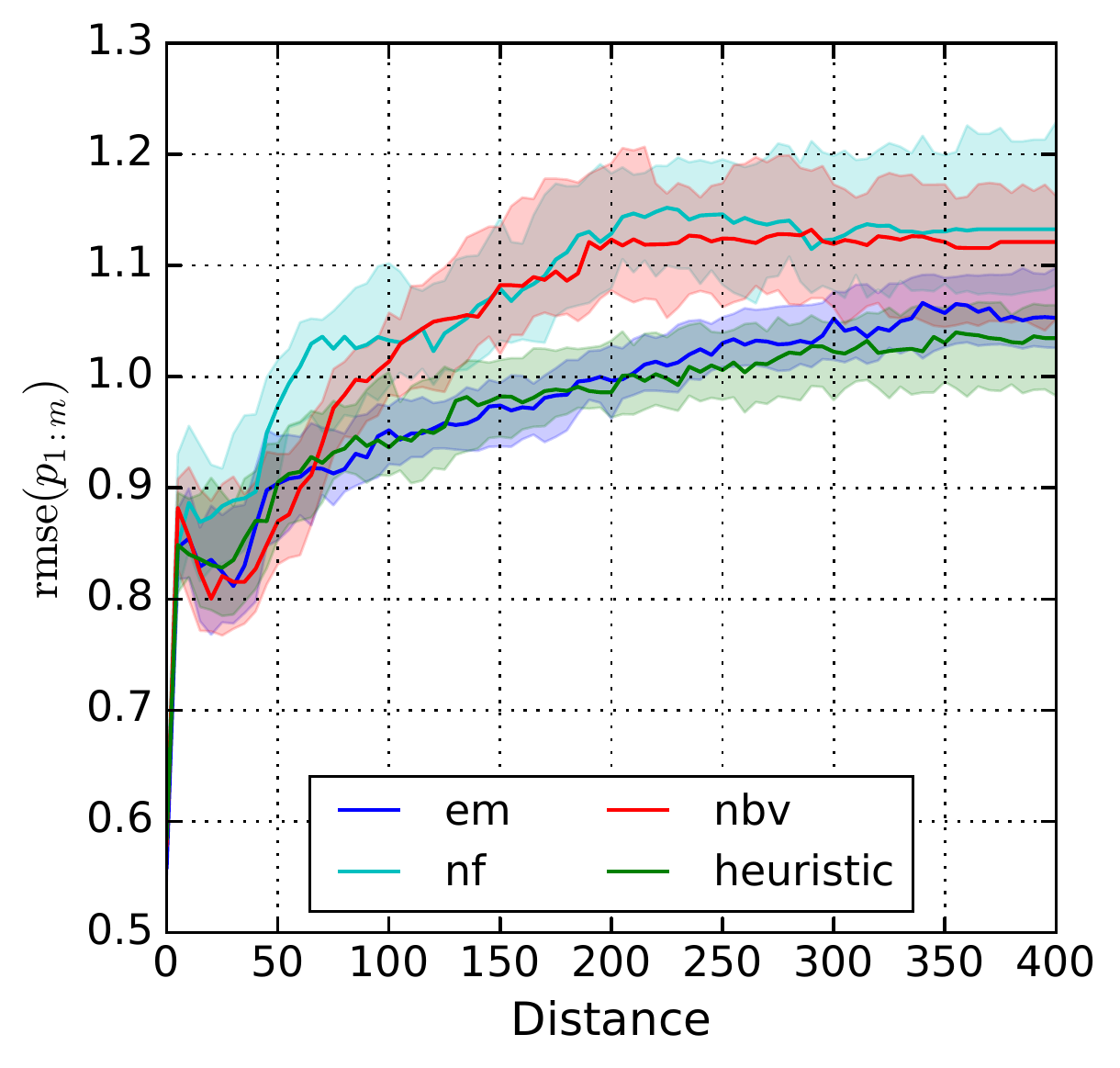}}\quad
    \subfloat[Map coverage]{\includegraphics[width=0.23\textwidth]{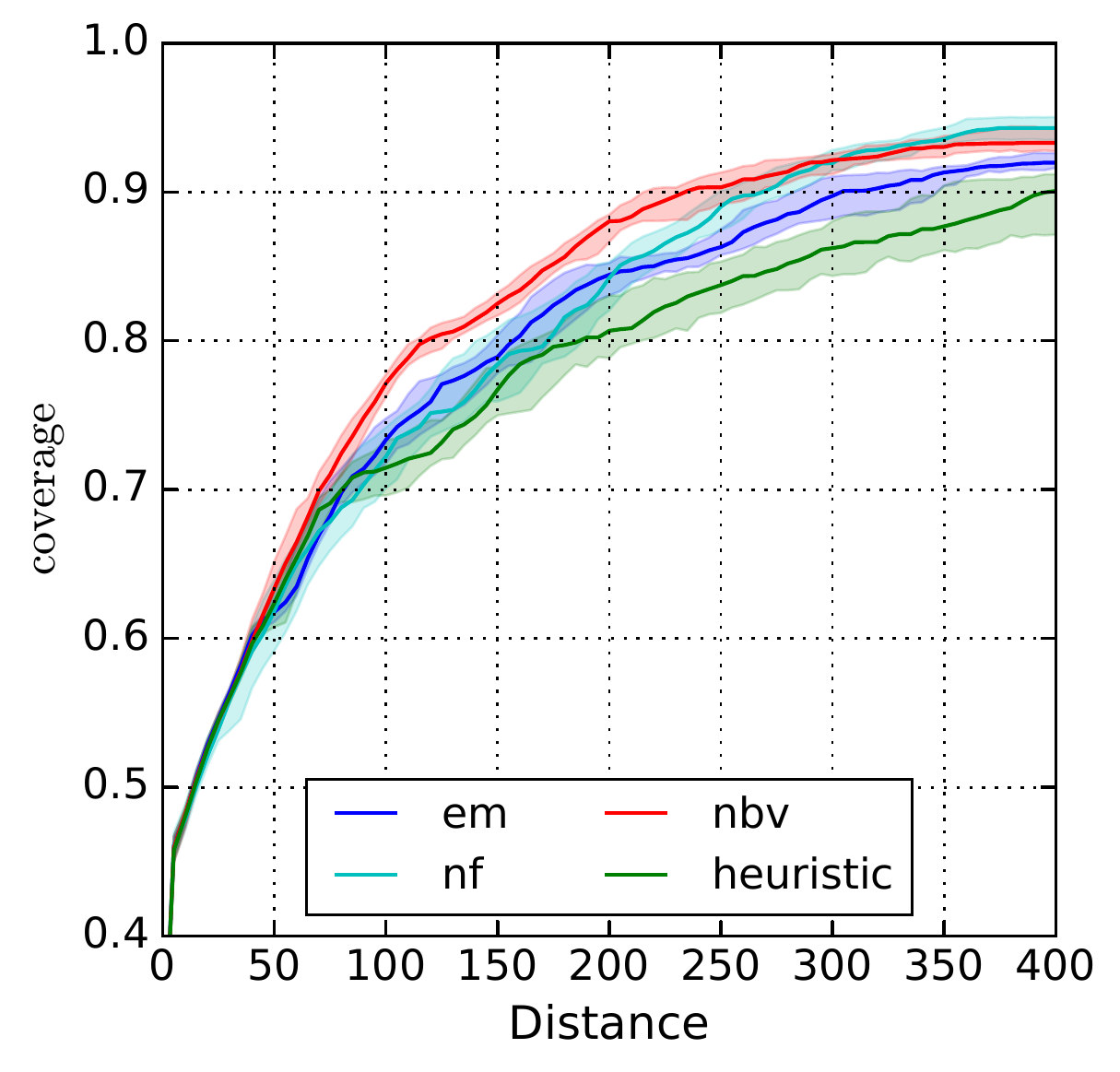}}
    \caption{Exploration performance mean values in the simulated marina environment of Fig. \ref{fig:bruce-sim-env} (b) for 60 trials (ten for each robot start location depicted in Fig. \ref{fig:bruce-sim-env}), with 95 percent confidence intervals shown. Values along the x-axis denote the robot's distance traveled. }
    \label{fig:bruce-sim-example2}
\end{figure*}

\section{Experimental Results}
\label{sec:results}

Here we present experimental results from (1) a simulation of a sonar-equipped underwater robot exploring and mapping two planar, previously unmapped environments, and (2) the real autonomous exploration of a harbor environment with a BlueROV2 underwater robot. One of the simulation environments is populated with point landmarks, permitting a solution that employs landmark nodes in the SLAM factor graph. The other environment is populated with heterogeneous structures, to which pose SLAM is applied instead.

The simulation environments were designed to emulate our subsequent real experiment, and are aimed at evaluating our algorithms quantitatively over a larger number of trials. The simulated sonar operates at \SI{5}{Hz} and has a field of view with range $r=[\SI{0}{m}, \SI{30}{m}]$ and horizontal aperture $\theta=[\SI{-65}{\degree}, \SI{65}{\degree}]$. Since we assume a 2D environment, the vertical aperture is ignored in the simulation. Zero-mean Gaussian noise is added to range and bearing measurements: $\sigma_r = \SI{0.2}{m}, \sigma_\theta=\SI{0.02}{rad}$. We provide simulated odometry measurements at \SI{5}{Hz}, which can be obtained from DVL and IMU sensors in our real experiments. Additionally, we add zero-mean Gaussian noise to the 2D transform: $\sigma_{x} = \sigma_y =  \SI{0.08}{m}, \sigma_{\theta} = \SI{0.003}{rad}.$

The two maps used for simulated experiments are shown in Figure~\ref{fig:bruce-sim-env}. In the landmark map (containing a randomly generated collection of landmarks that is kept constant throughout all trials), the robot is deployed from nine different starting positions. In the pose SLAM map (whose structures are derived from our real-world experimental results), the robot is deployed from one of six starting positions. The exploration process is terminated when there are no frontiers remaining in the map that can be feasibly reached by our planner. 

We consider the following metrics to evaluate an exploration algorithm. Pose uncertainty measures the uncertainty of the current vehicle pose, which is computed as $|\Sigma_{\mathbf x_T}|^{\frac 13}$. Trajectory error represents the root mean square error (RMSE) of the pose estimate over the entire trajectory $(\frac 1T \sum_{i=1}^{T}(||\mathbf x_i - \mathbf x_i^{gt}||^2)^{\frac 12}$, where $\mathbf x^{gt}$ denotes ground truth pose. Exploration progress is represented as map coverage, i.e., the ratio of mapped area to the whole workspace. We use the RMSE of the reconstructed point cloud of landmarks as a proxy for map error $(\frac 1M \sum_{i=j}^{M}(||\mathbf p_j - \mathbf p_j^{gt}||^2)^{\frac 12}$, where estimated landmark position $\mathbf p_j$ is obtained using the inverse sensor model provided estimated pose, and $\mathbf p_j^{gt}$ denotes ground truth landmark position. It is worth noting that the map error can be applied to both landmark-based SLAM and pose SLAM as shown in the experiments below.

\subsection{Algorithm Comparison}
\label{subsec:algorithm_comparison}
In the comparisons to follow, four algorithms are considered, all of which employ different techniques to repeatedly select one of the goal configurations generated by our motion planner described in Sec. \ref{sec:motion-planning}. First, we examine frontier-based exploration \cite{Yamauchi1997}, which explores by repeatedly driving to the \textit{nearest frontier} goal configuration. Secondly, we examine a next-best-view exploration framework that selects the goal configuration anticipated to achieve the largest \textit{information gain} with respect to the robot's occupancy map, in accordance with the technique proposed in \cite{Bircher2016}. Thirdly, we examine the active SLAM framework of \cite{Suresh2020}. When a robot's \textit{pose uncertainty} (per the D-optimality criterion \cite{Carrillo2012}) exceeds a designated threshold, this framework selects the place-revisiting goal configuration that maximizes a utility function expressing a weighted tradeoff between uncertainty reduction and information gain. At all other times, the approach reverts to that of \cite{Bircher2016}, purely seeking information gain. In the results to follow, we will refer to this approach as the \textit{heuristic} approach, due to its use of a tuned threshold on pose uncertainty to determine when place revisiting occurs. These three algorithms are compared against our proposed EM exploration algorithm. The heuristic and EM algorithms have been parameterized to achieve the closest possible equivalence in trajectory estimation error (the root mean square error of the full pose history, recorded at each instant in time during exploration), providing a baseline from which to examine other performance characteristics that differ.

\subsection{Simulated Exploration over Landmarks}
\label{subsec:landmarks_simulation}
%In the landmark environment shown in Figure \ref{sim:landmark}, we compared the proposed EM framework with (1) a nearest frontier (NF) approach, (2) a next-best-view (NBV) approach, and (3) a heuristic approach (Heuristic) over \TODO{JW: how many test trials?} exploration trials. The NF algorithm is choosing the nearest frontier as the goal location for exploration at every decision-making time. The NBV planner computes accumulated gain discounted exponentially by distance from the start. The gain is defined with regard to the occupancy status of the visible volume at the forward simulation steps. Besides, a heuristic approach switches between two modes: if current pose uncertainty is smaller than a specified threshold, NBV is used to achieve rapid exploration; otherwise it will seek a revisitation location to reduce pose uncertainty.

In the landmark environment shown in Figure \ref{sim:landmark}, we compared the nearest frontier (NF), next-best-view (NBV), heuristic, and EM algorithms across 45 simulated trials (five for each robot start location). In this environment, the landmarks are assumed to be infinitesimally small, and not to pose a collision hazard.  
We provide representative examples of the four exploration algorithms in Figure \ref{fig:landmarks_example}. %\TODO{JW: Are these algorithms only have one trail? Do they have the same landmarks distribution?} 
The average performance of the four exploration algorithms is shown in Figure \ref{fig:landmarks_performance}. The NBV method achieves superior coverage of the environment, but it also yields the highest pose uncertainty, trajectory error, and map error. The NF approach achieves poorer coverage than NBV, but achieves lower pose uncertainty and map/trajectory error during exploration. The heuristic approach covers the environment the least efficiently, while it offers the lowest pose uncertainty and map error during the exploration process. Our proposed EM algorithm achieves map coverage superior to the heuristic approach, with comparable but slightly higher pose uncertainty and map error.  
%At the same time, the EM algorithm has a similar performance with the Heuristic approach about the uncertainty and the error of the final results.

\subsection{Simulated Exploration with Pose SLAM in a Harbor Environment}

As there is no ground truth information available in our real-world experiments, we have instead created the simulated map of Fig. \ref{sim:pose} using manually collected data from the harbor environment used in those experiments. 
Specifically, the simulated map of Fig. \ref{sim:pose} is represented by a point cloud that is obtained from the SLAM framework described above in Sec. \ref{sec:uw_slam}.
During simulated exploration, feature points that fall within the sensor field of view are sampled and Gaussian noise is added - these sensor observations are used to populate an occupancy map, as described in Sec. \ref{sec:occupancy_mapping}.
Ten trials of each algorithm were run at each of six different start locations (denoted as green triangles in Figure~\ref{fig:bruce-sim-env}).
Mapping and localization errors are reported with respect to our experimentally-derived ground truth point cloud map. 
The mapping error is computed as follows: for every estimated feature point, we compute its distance to the nearest ground truth point.

The average performance of each algorithm is shown in Figure~\ref{fig:bruce-sim-example2}, with specific numerical values from this performance comparison given in Tables \ref{table:sim pose uncertainty},  \ref{table:sim pose error}, \ref{table:sim map coverage} and \ref{table:sim map error}.
%For each performance metric, average values and 95\% confidence interval are visualized in Figure~\ref{fig:bruce-sim-example2}. 
As before, the NBV algorithm achieves the most efficient exploration, but at the expense of high pose uncertainty, map error, and trajectory error. 
The pose uncertainty, trajectory error, and map error of the heuristic and EM algorithms are similar in value, and significantly lower than that of the NF and NBV algorithms. 
%It is worth mentioning that parameters in Heuristic are tuned to match the same uncertainty-awareness performance as in EM and parameters in EM are manually adjusted for this specific environment. 
While the heuristic and EM algorithms have similar pose uncertainty and map/trajectory error, the heuristic approach falls behind EM in terms of map coverage, which is especially evident in Table \ref{table:sim map coverage}, which denotes the total distance traveled to reach various percentages of coverage.

In Figure~\ref{fig:bruce-sim-example}, one representative execution trace of each algorithm is presented. %with four metrics with respect to traveled distance. 
A similar phenomenon, a dense interconnection of loop closure constraints among poses, can be observed in the performance of the Heuristic and EM algorithms. 
%\Comment{YH: what is a dense interconnection here?} \Comment{JM: Agreed, perhaps a question for Jinkun.}
In contrast, the greedy algorithms (NF and NBV) take similar paths toward the top right corner in the beginning, which leads to growing pose uncertainty. Although the pose uncertainty is ultimately reduced, map accuracy is impacted as demonstrated in Figure \ref{fig:bruce-sim-example2} and Table \ref{table:sim map error}.

\begin{figure}[t]
	\centering
	\includegraphics[width=0.3\textwidth]{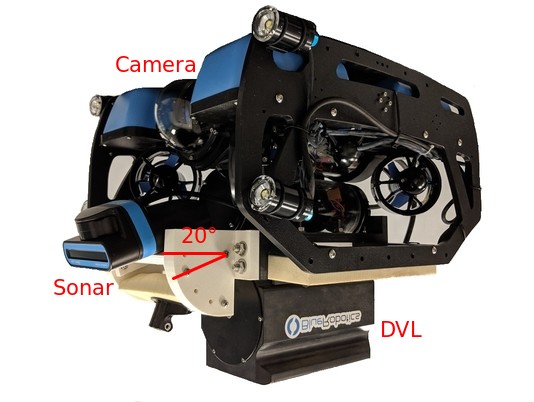} %\\
	\includegraphics[width=0.5\textwidth]{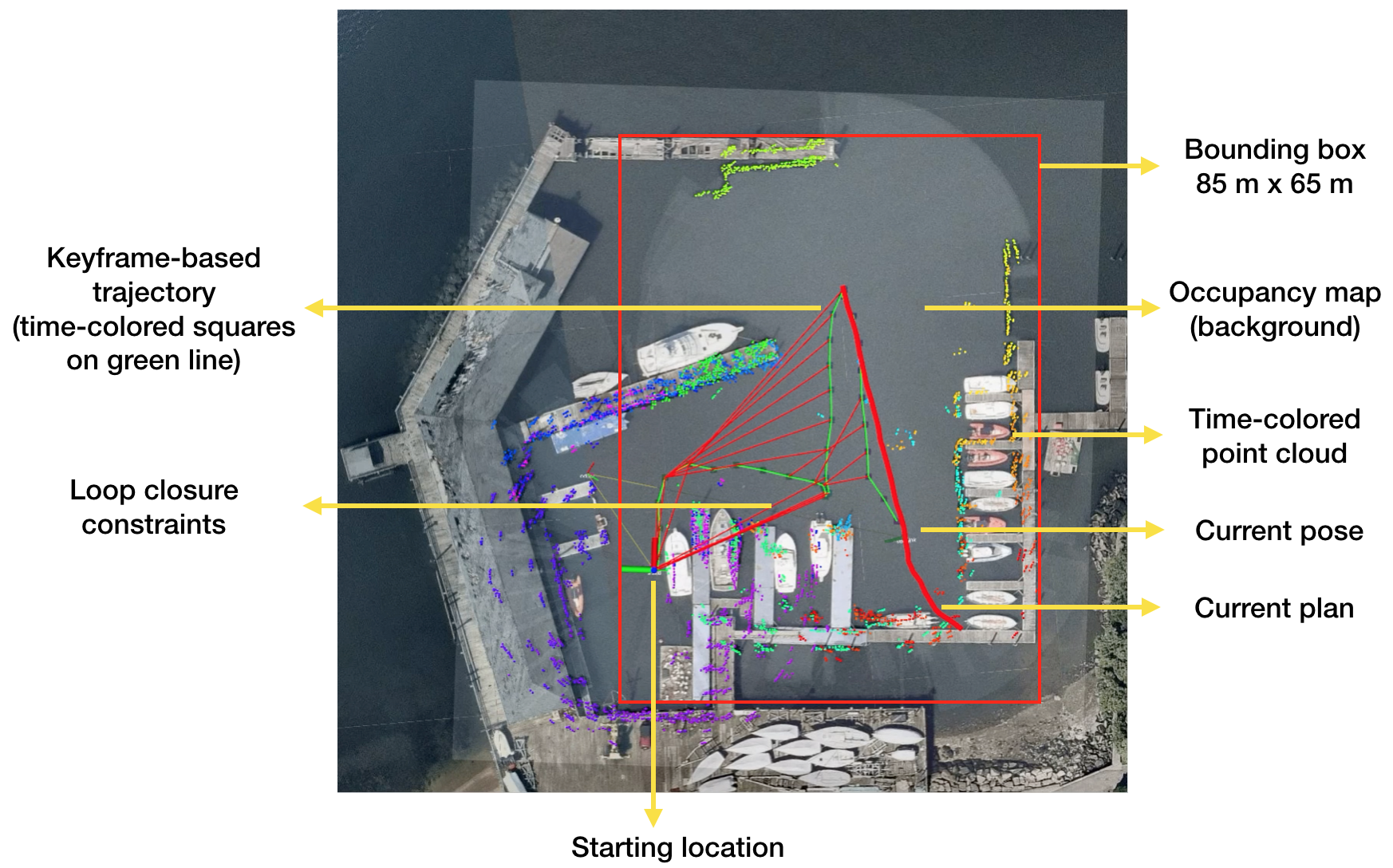}
	\caption{The custom-instrumented BlueROV2 robot used in our experiments, and an illustration of the setup of our experiments at the U.S. Merchant Marine Academy in King's Point, NY.}
	\label{fig:bruce-real-setup}
\end{figure}

\begin{figure*}
	\centering
	\subfloat[Frame 1]{\includegraphics[width=0.33\textwidth]{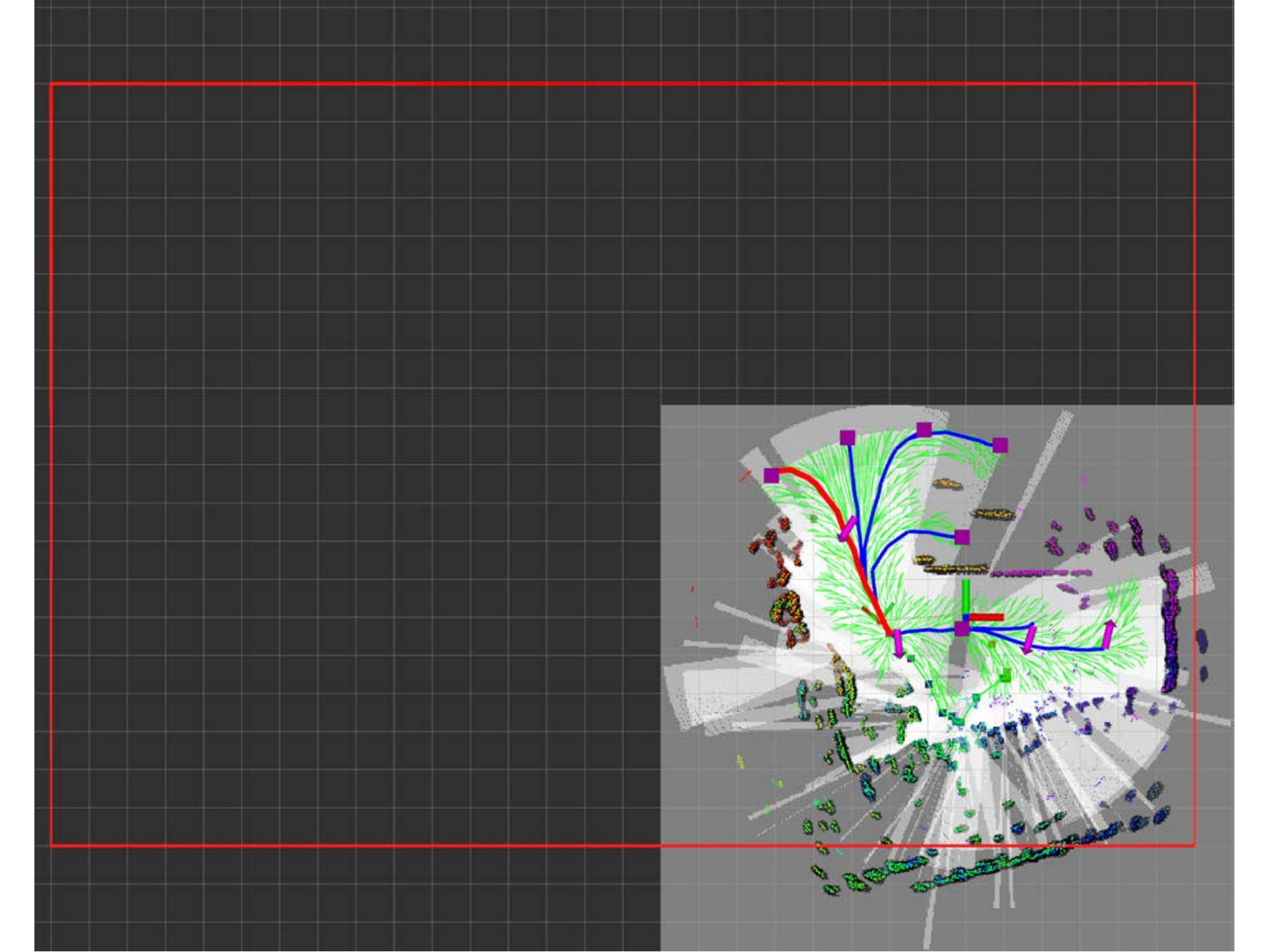}}
	\subfloat[Frame 2]{\includegraphics[width=0.33\textwidth]{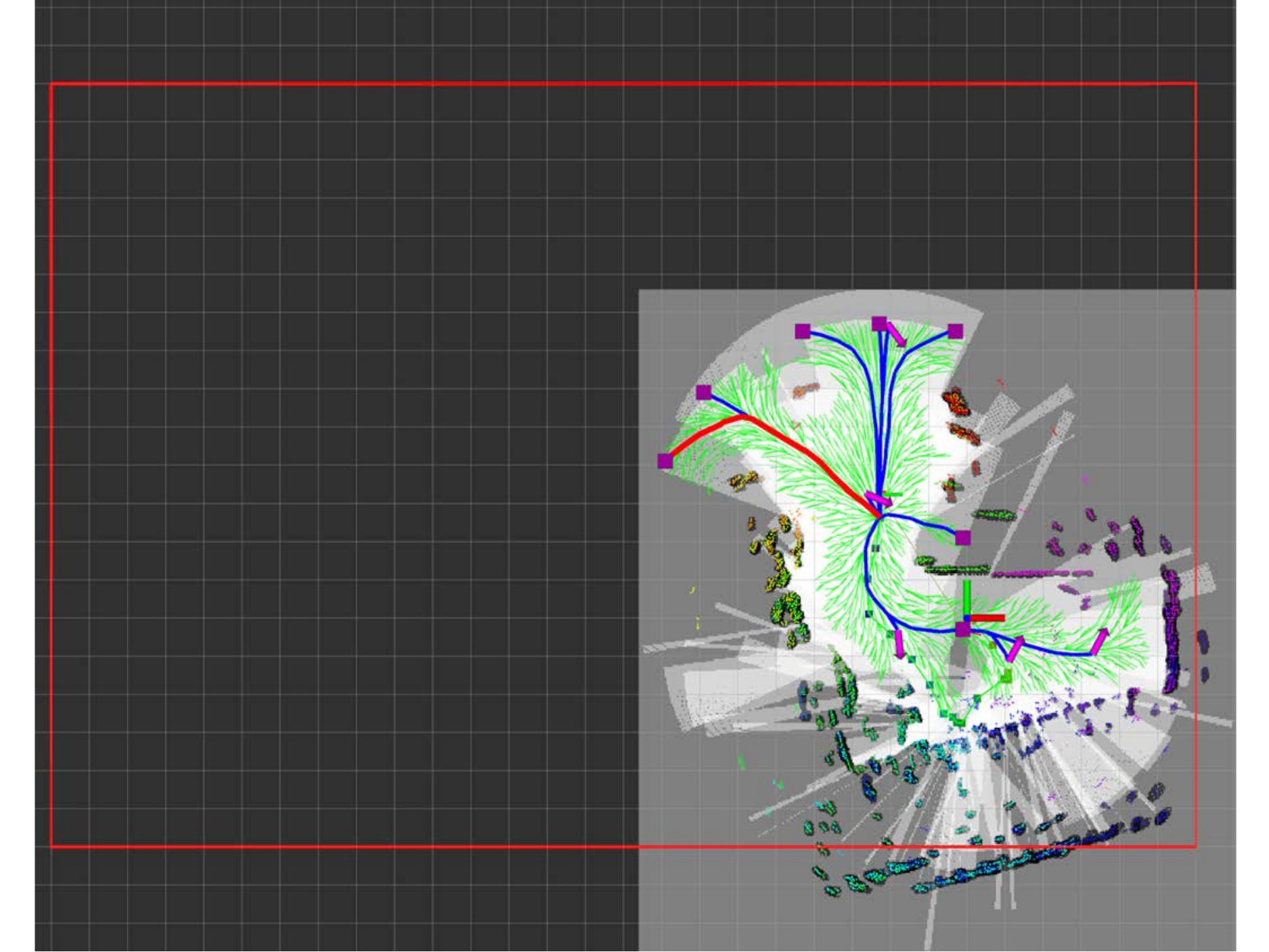}}
	\subfloat[Frame 3]{\includegraphics[width=0.33\textwidth]{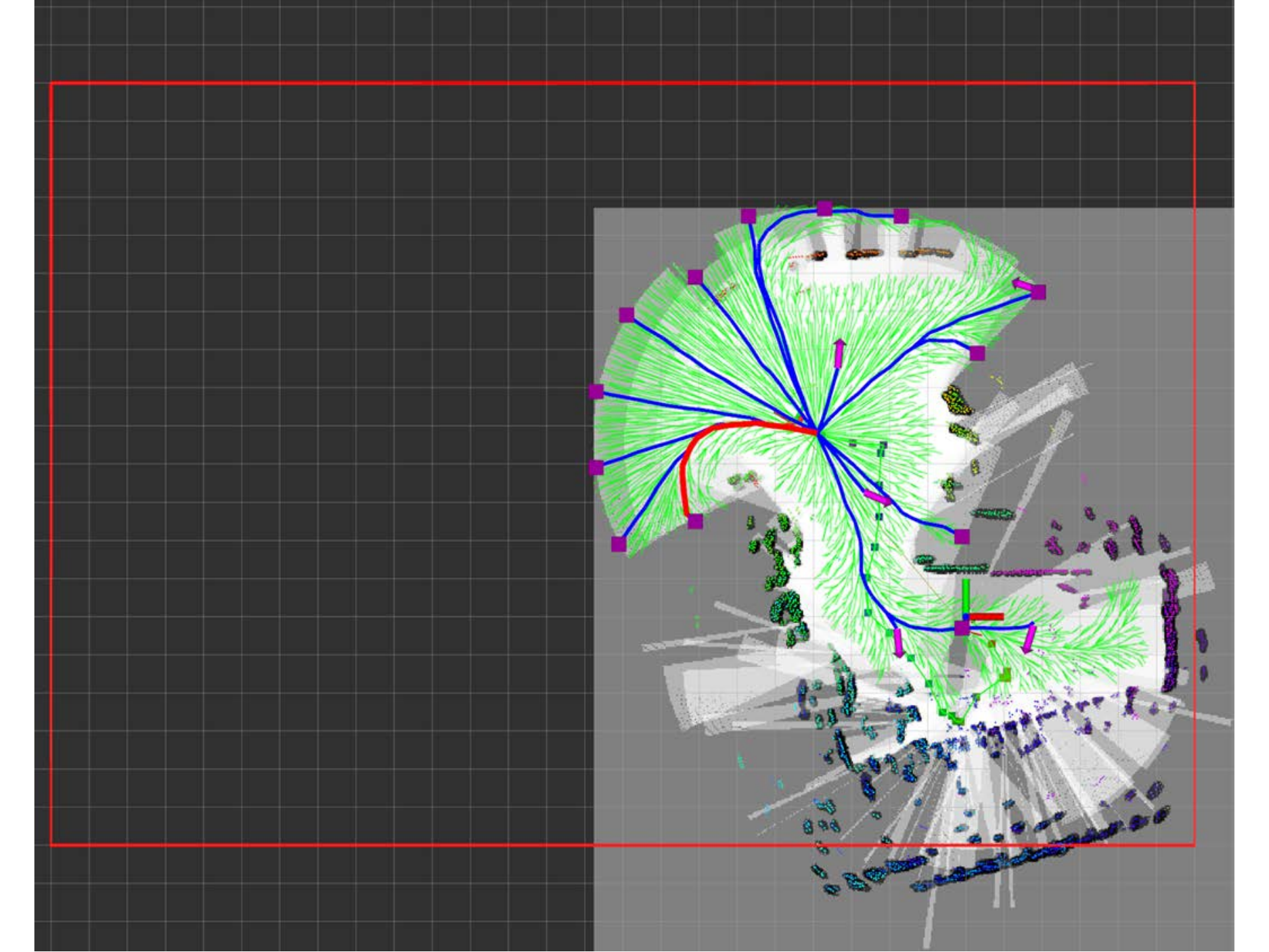}}\\
	\subfloat[Frame 4]{\includegraphics[width=0.33\textwidth]{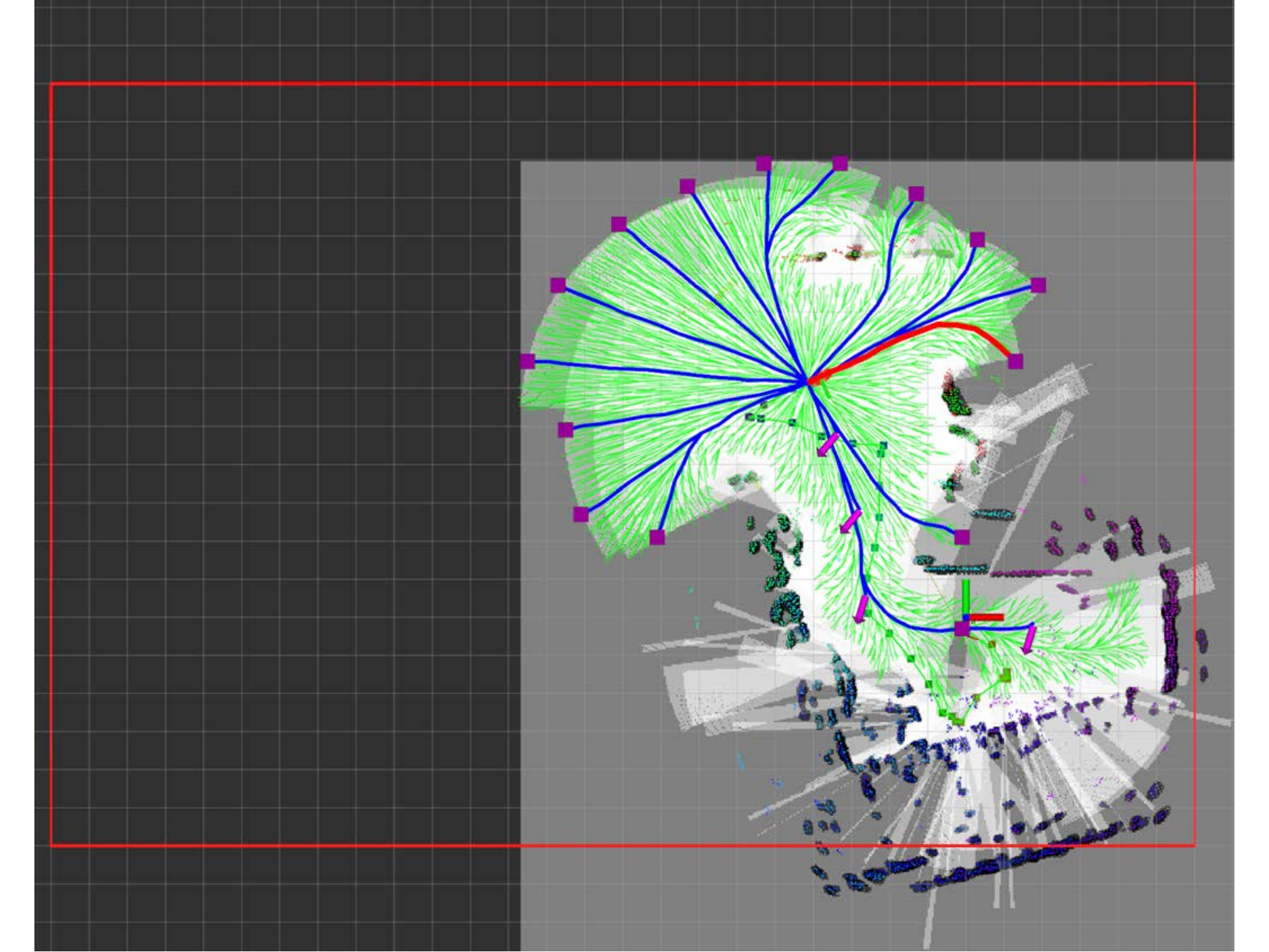}}
	\subfloat[Frame 5]{\includegraphics[width=0.33\textwidth]{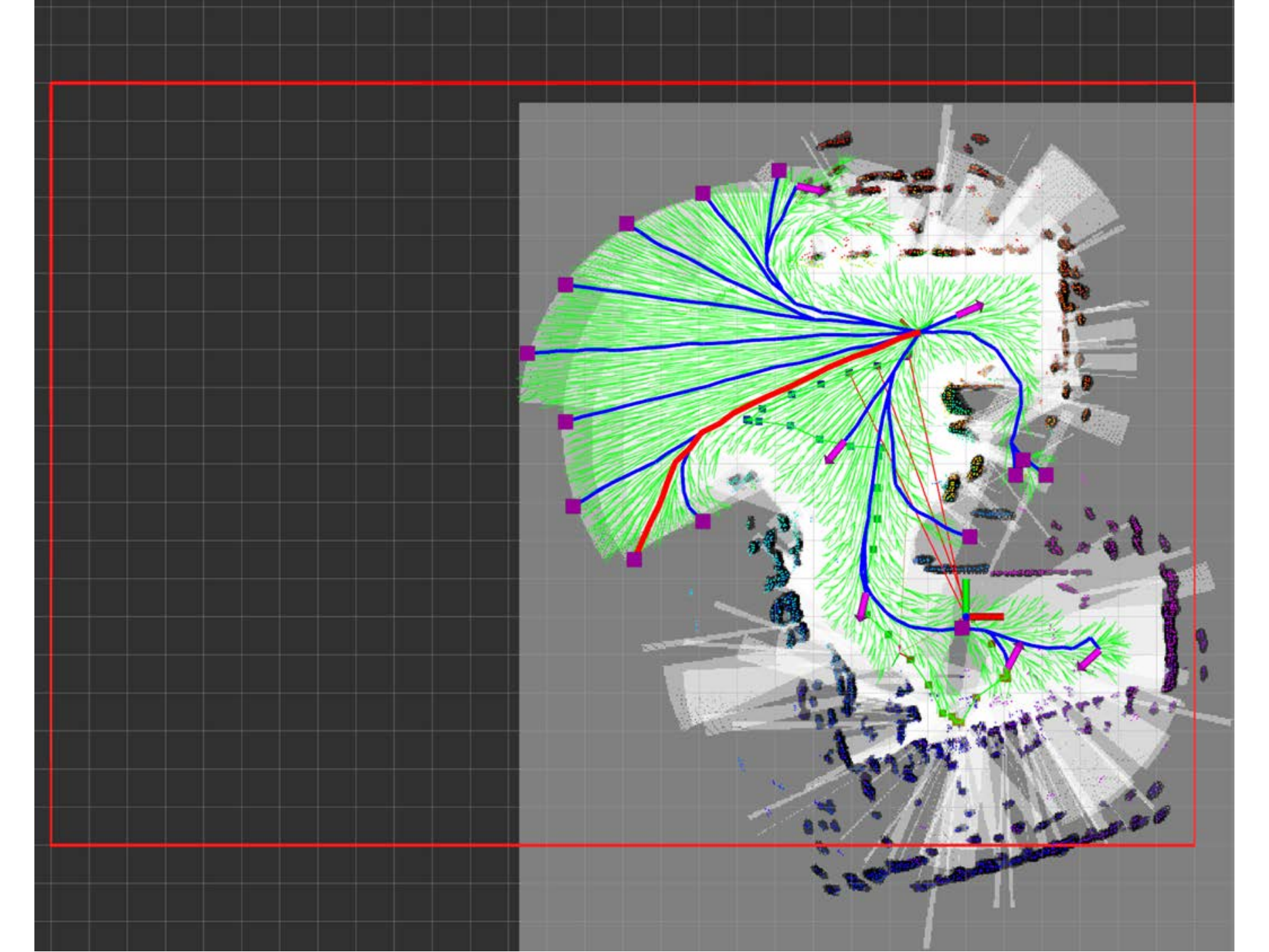}}
	\subfloat[Frame 6]{\includegraphics[width=0.33\textwidth]{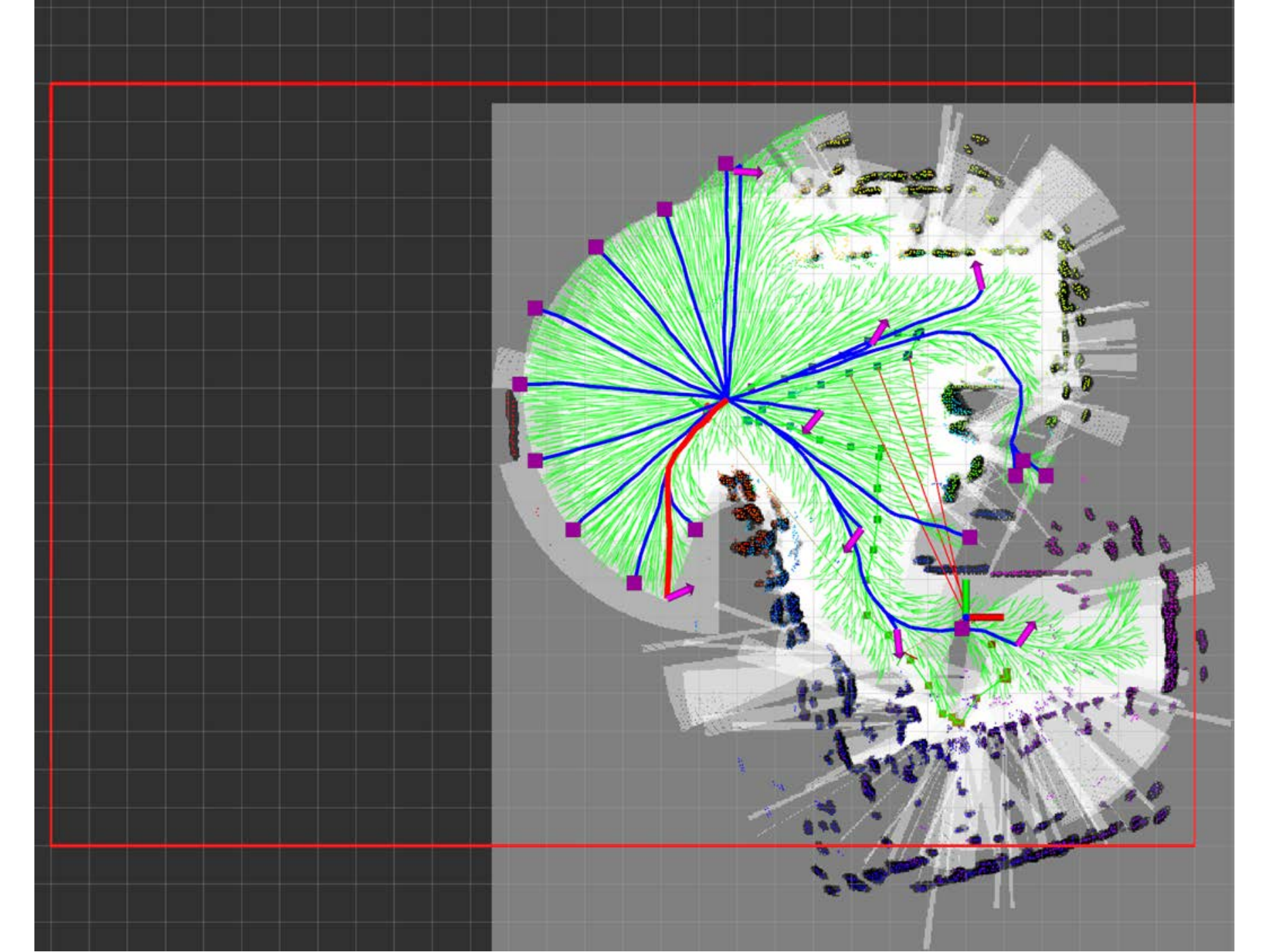}}\\
	\subfloat[Frame 7]{\includegraphics[width=0.33\textwidth]{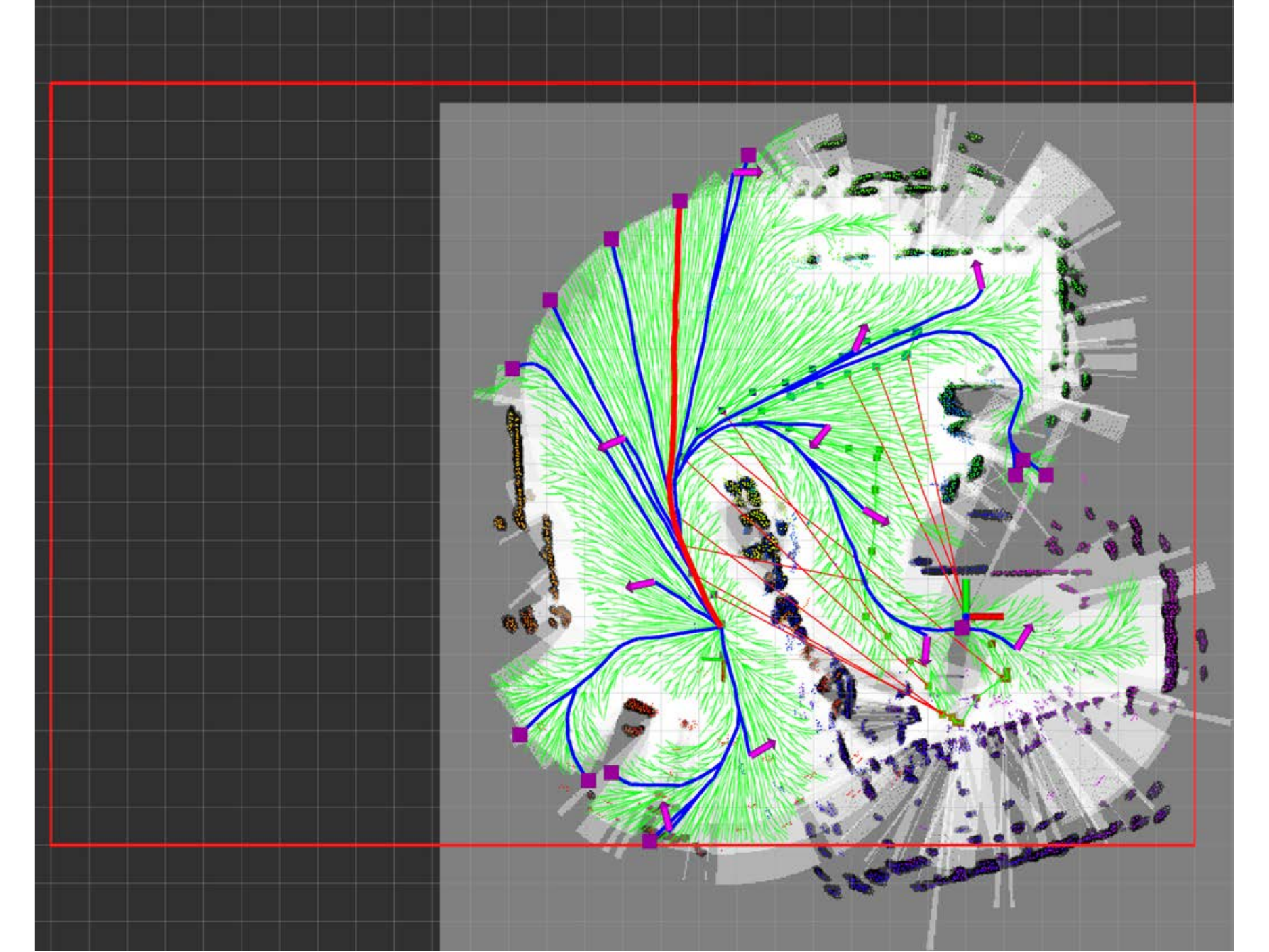}}
	\subfloat[Frame 8]{\includegraphics[width=0.33\textwidth]{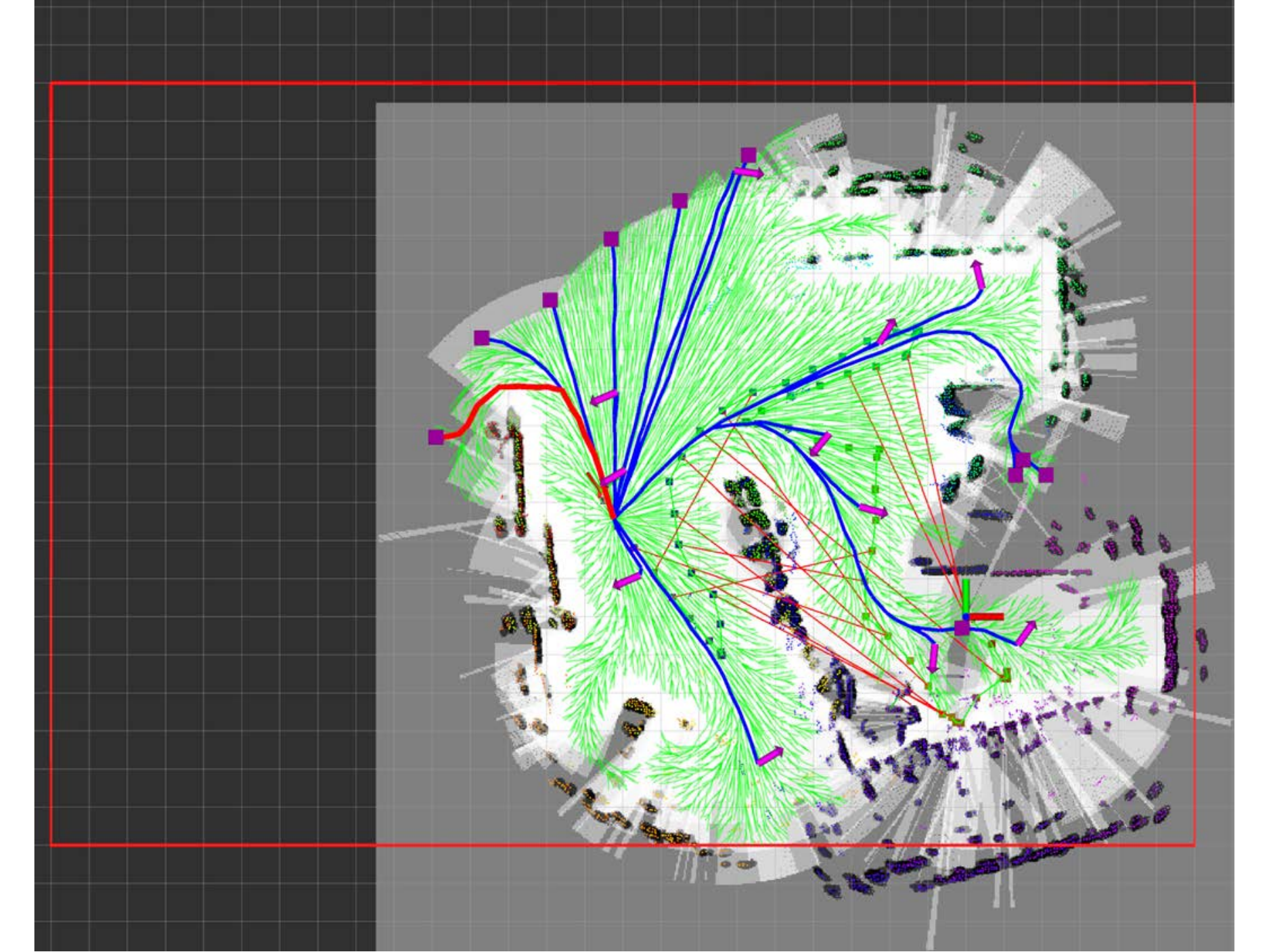}}
	\subfloat[Frame 9]{\includegraphics[width=0.33\textwidth]{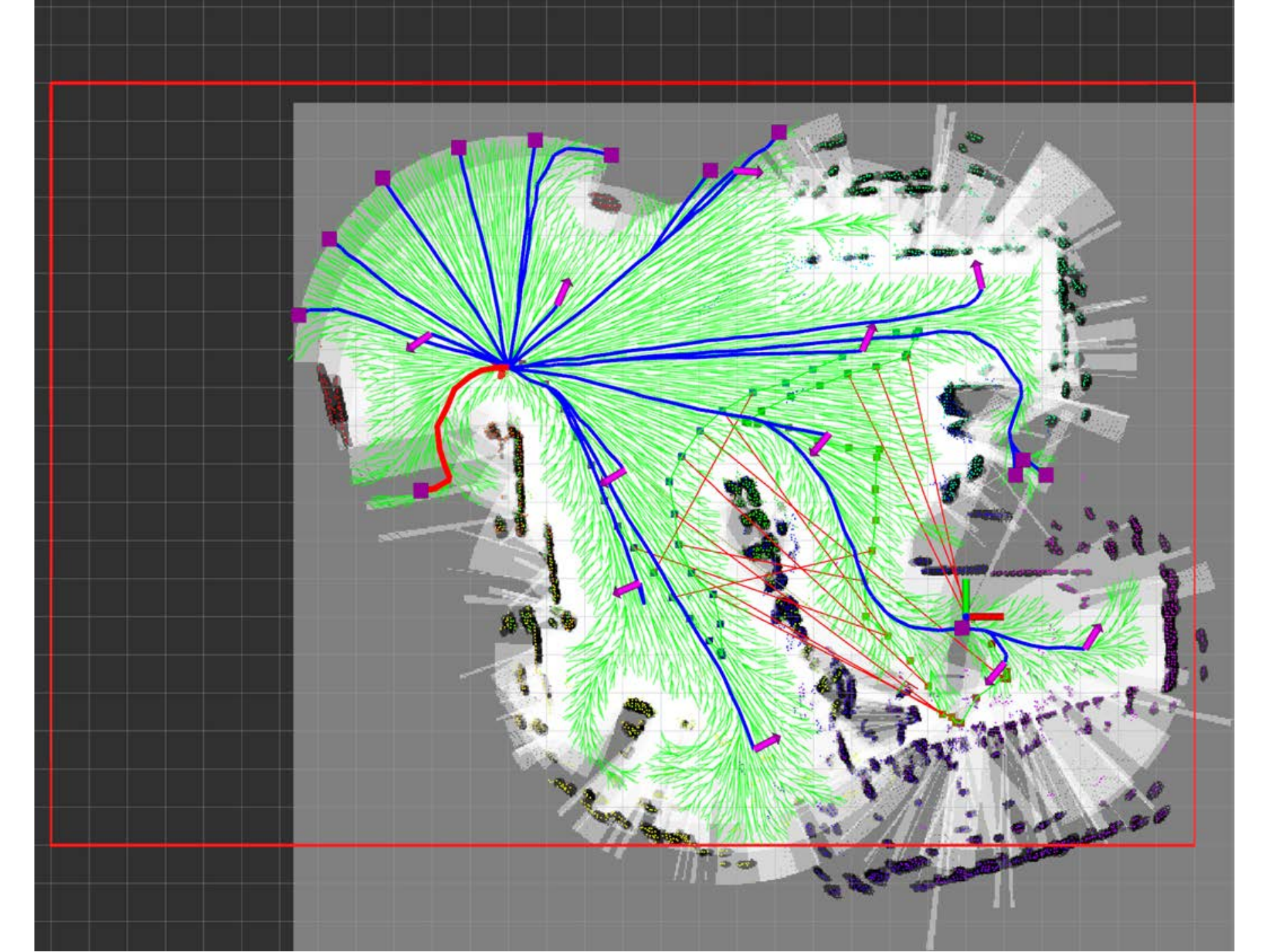}}
	\caption{Examples of the motion planning process used to support exploration. Purple squares are frontier candidates, purple arrows are place revisiting candidates. Blue lines are candidate trajectories and the green graph is used for global planning in conjunction with A*.}
	\label{fig:bruce-real-planning}
\end{figure*}

\begin{figure*}
	\centering
	\subfloat[Run 1 using NF]{\includegraphics[width=0.33\textwidth]{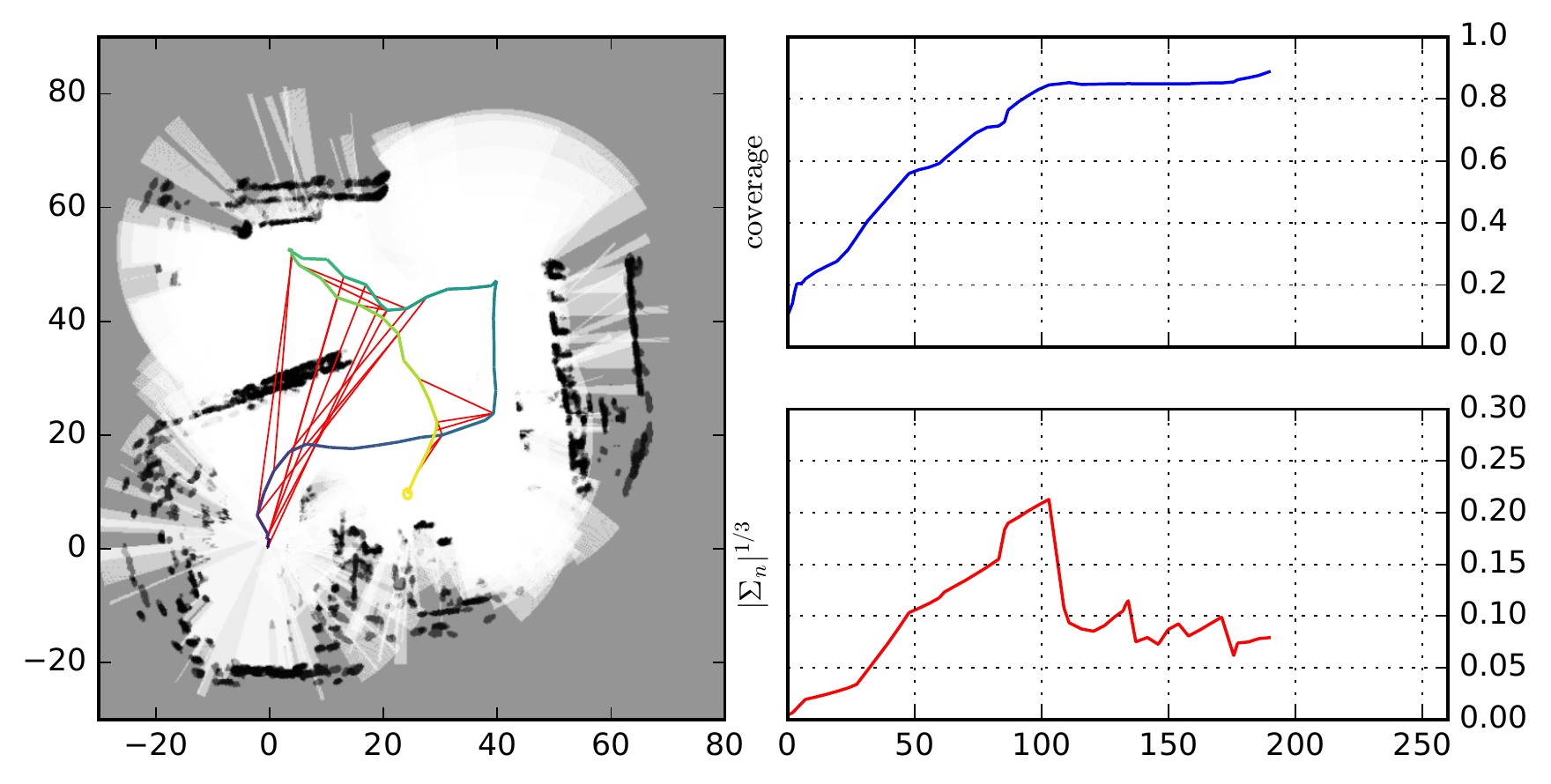}}
	\subfloat[Run 2 using NF]{\includegraphics[width=0.33\textwidth]{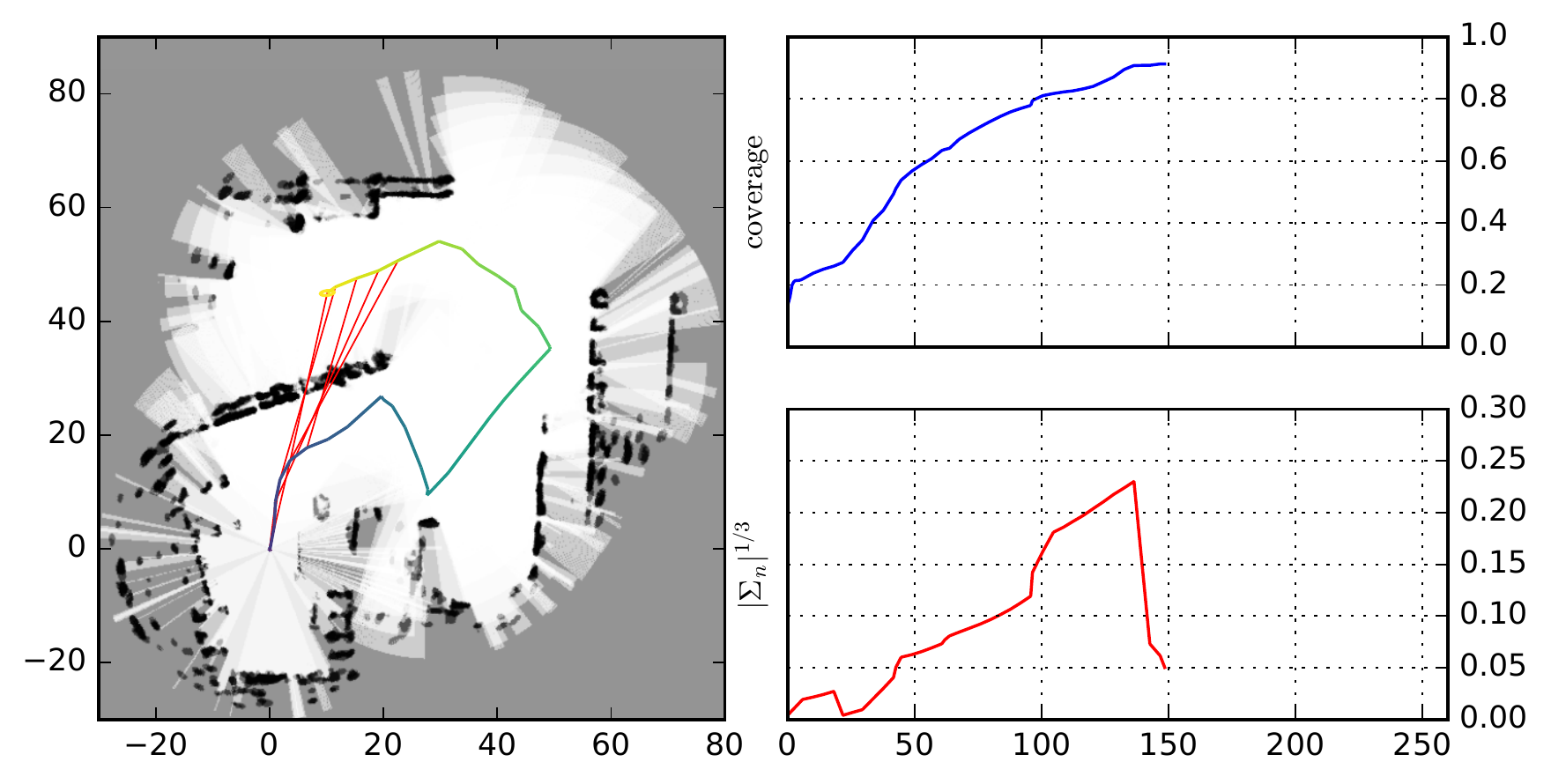}}
	\subfloat[Run 3 using NF]{\includegraphics[width=0.33\textwidth]{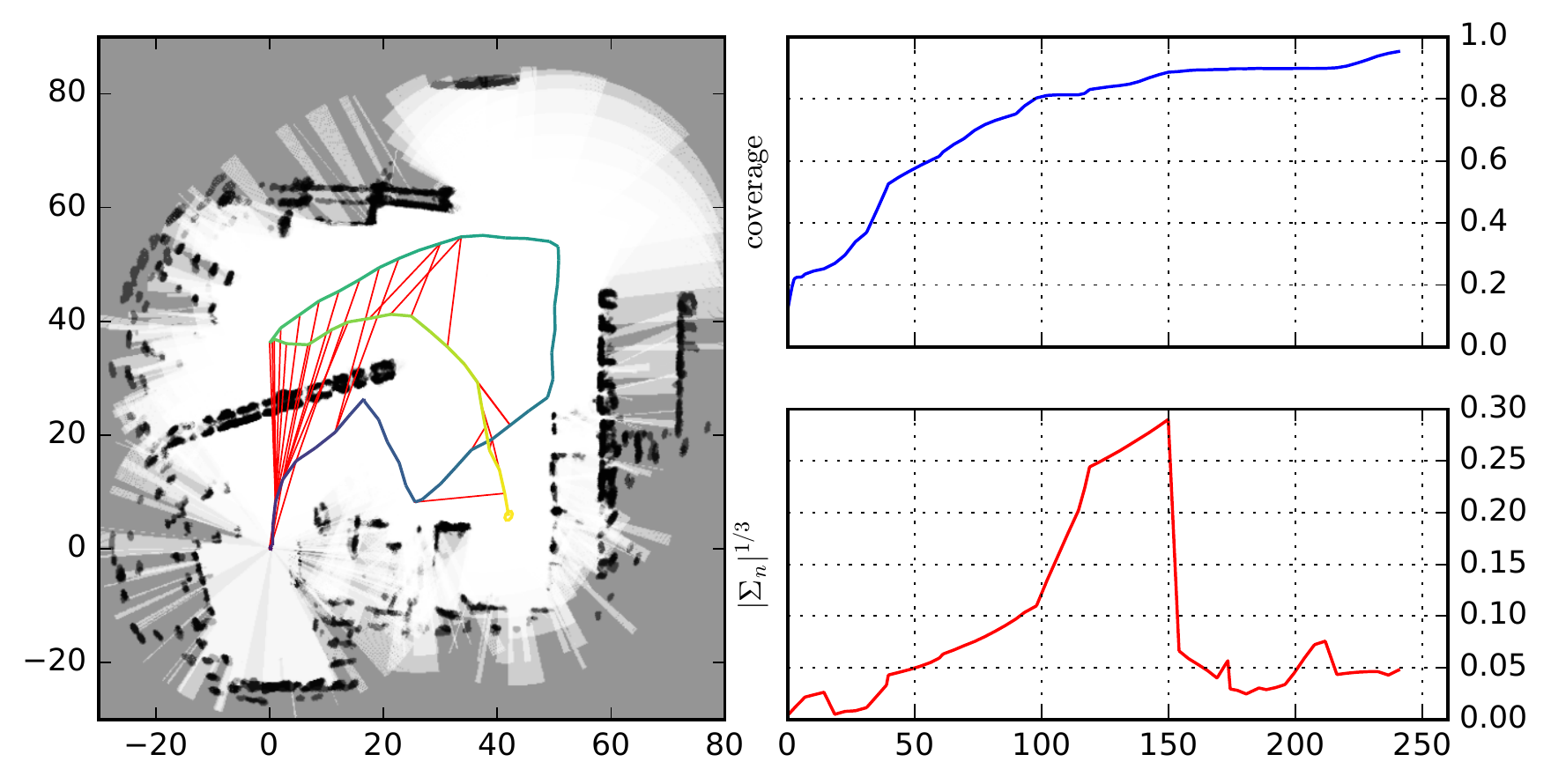}}\\
	\subfloat[Run 1 using NBV]{\includegraphics[width=0.33\textwidth]{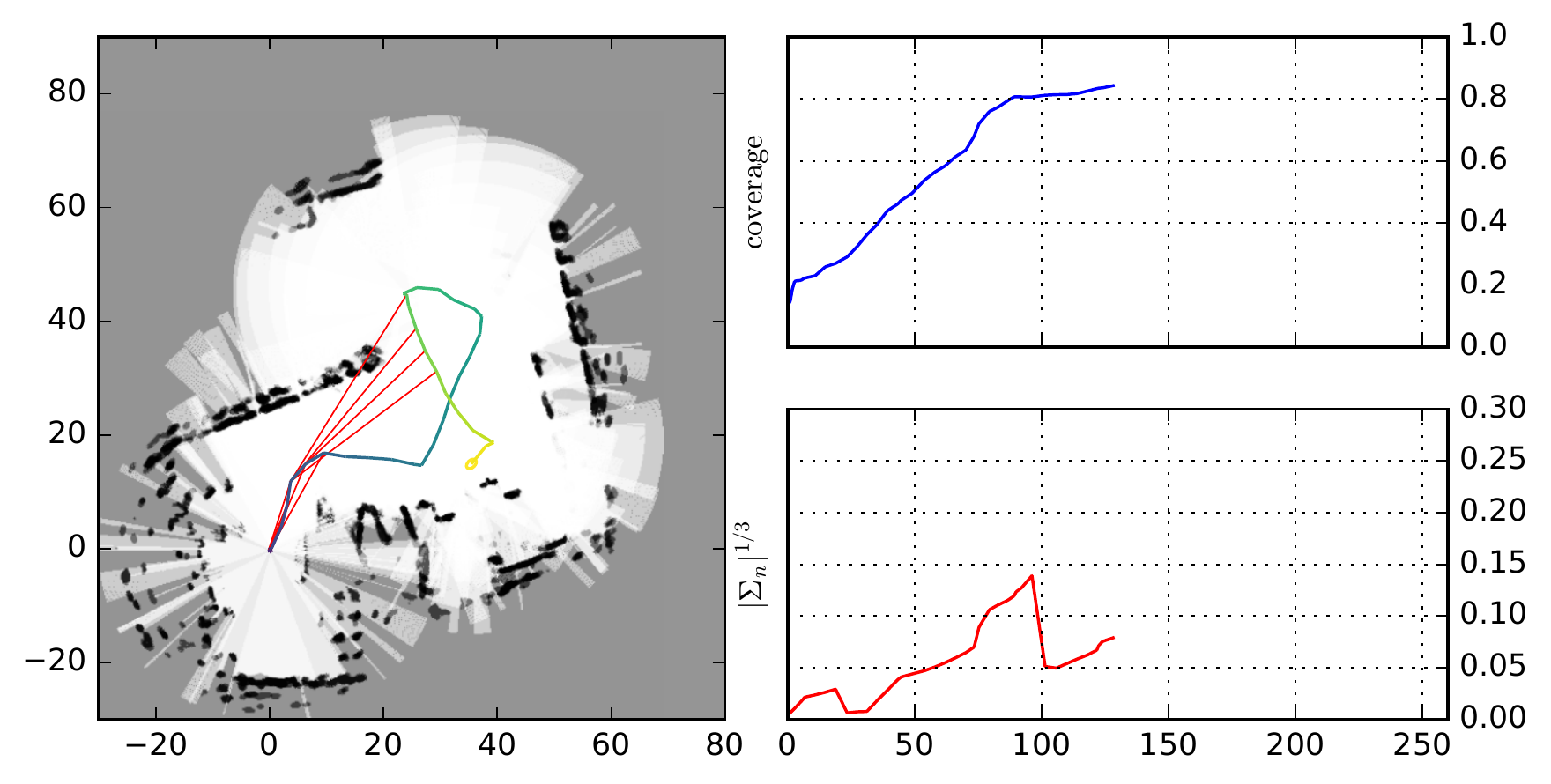}}
	\subfloat[Run 2 using NBV]{\includegraphics[width=0.33\textwidth]{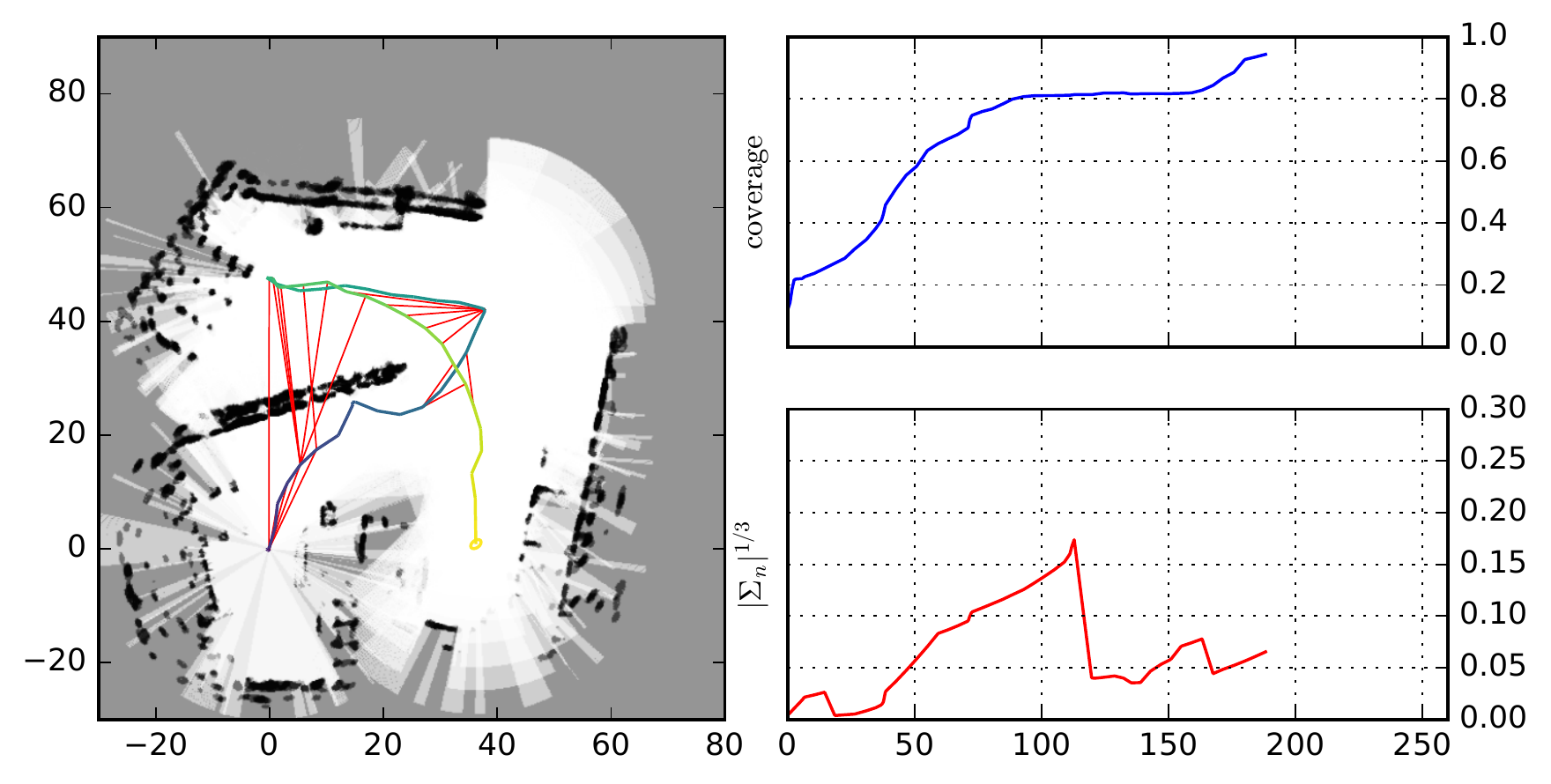}}
	\subfloat[Run 3 using NBV]{\includegraphics[width=0.33\textwidth]{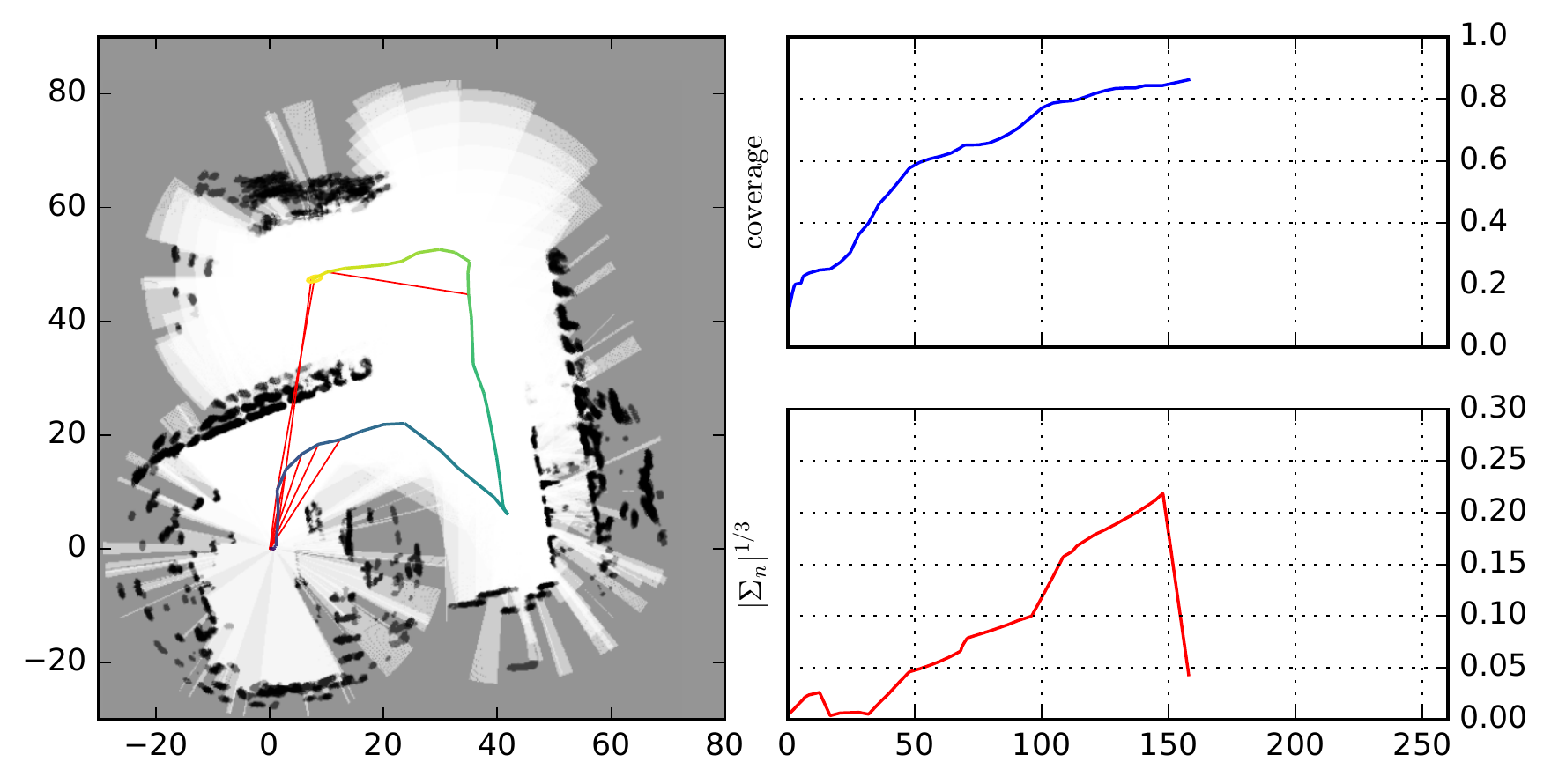}}\\
	\subfloat[Run 1 using Heuristic]{\includegraphics[width=0.33\textwidth]{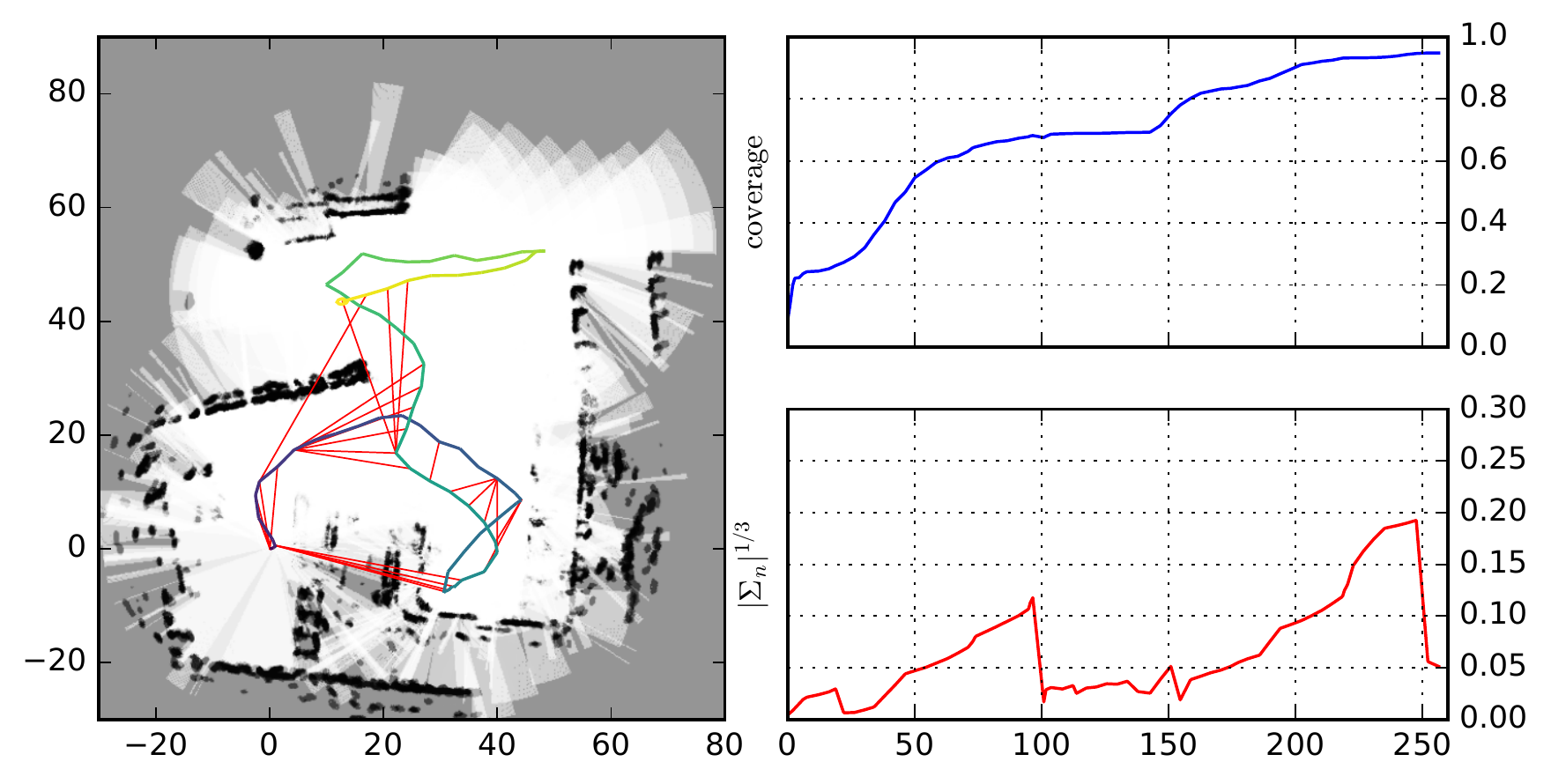}}
	\subfloat[Run 2 using Heuristic]{\includegraphics[width=0.33\textwidth]{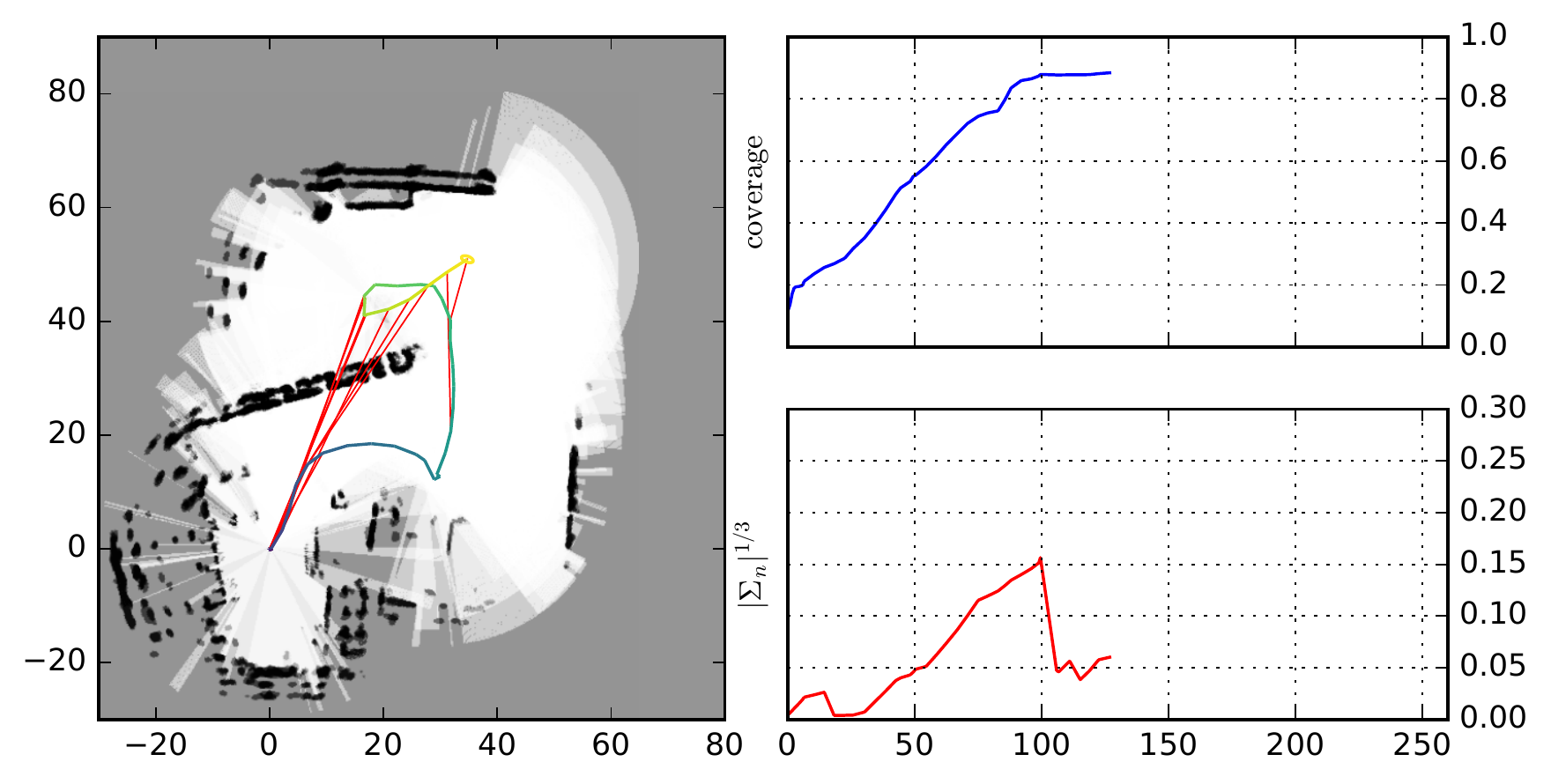}}
	\subfloat[Run 3 using Heuristic]{\includegraphics[width=0.33\textwidth]{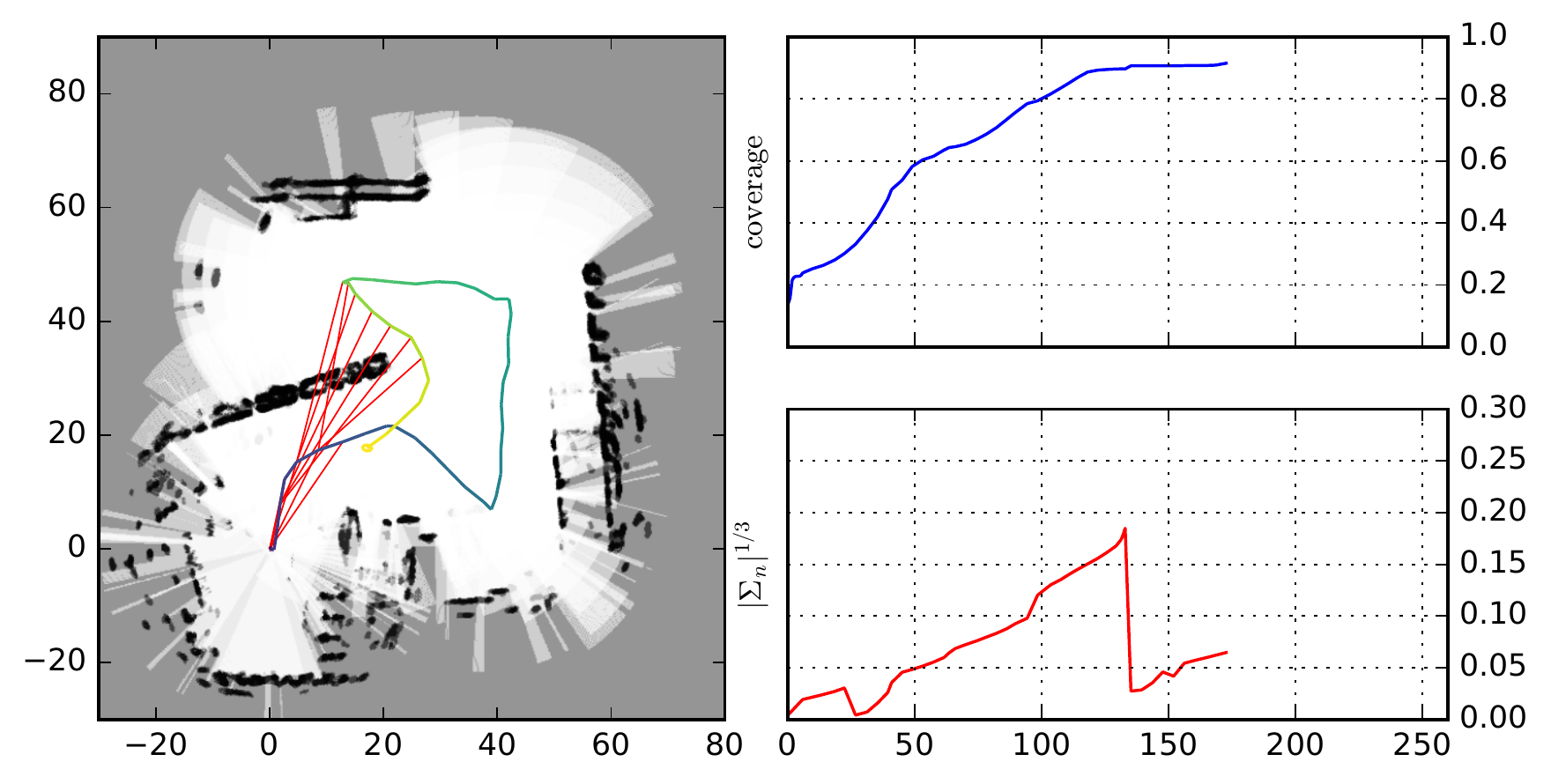}}
	\caption{Three runs of real-world robot exploration using the NF, NBV and Heuristic algorithms in the  environment depicted in Fig. \ref{fig:bruce-real-setup}. Each run shows the BlueROV's pose history and loop closure constraints over the occupancy map at left, with coverage and pose uncertainty shown at right, plotted against travel distance. }
	\label{fig:bruce-real-others}
\end{figure*}

\begin{figure*}
	\centering
	\subfloat[Run 1: step 10]{\includegraphics[width=0.31\textwidth]{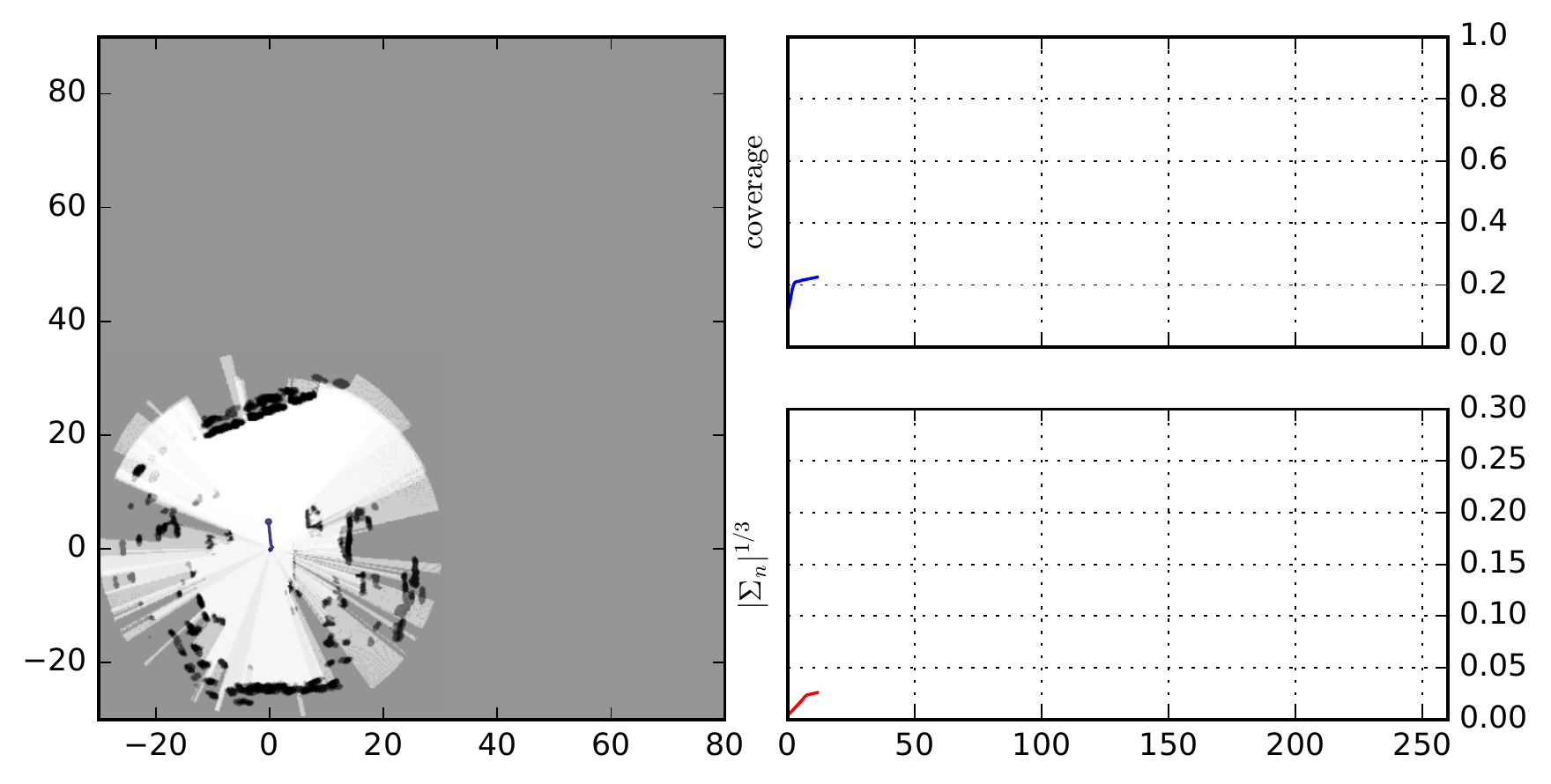}}
	\subfloat[Run 2: step 10]{\includegraphics[width=0.31\textwidth]{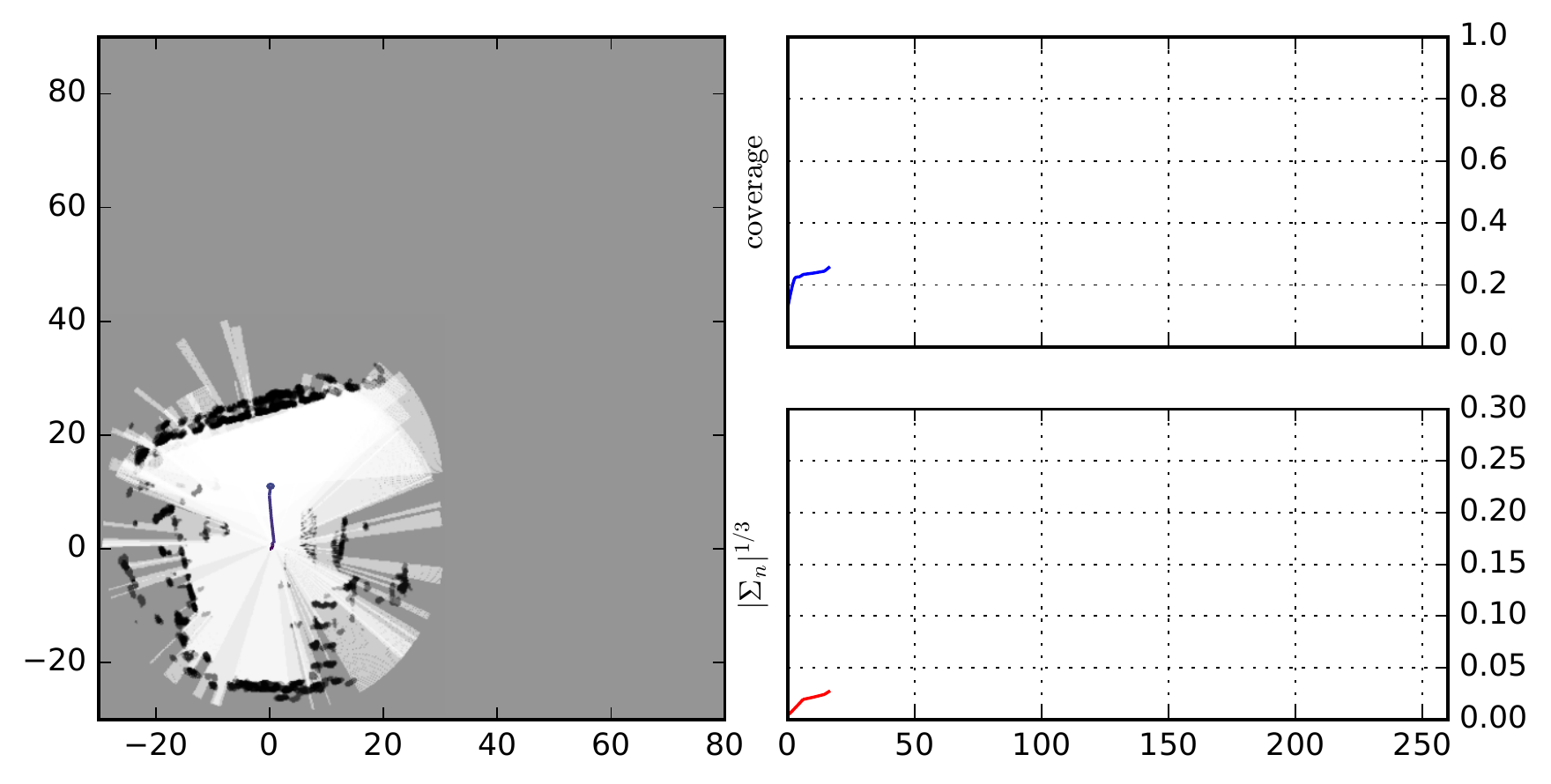}}
	\subfloat[Run 3: step 10]{\includegraphics[width=0.31\textwidth]{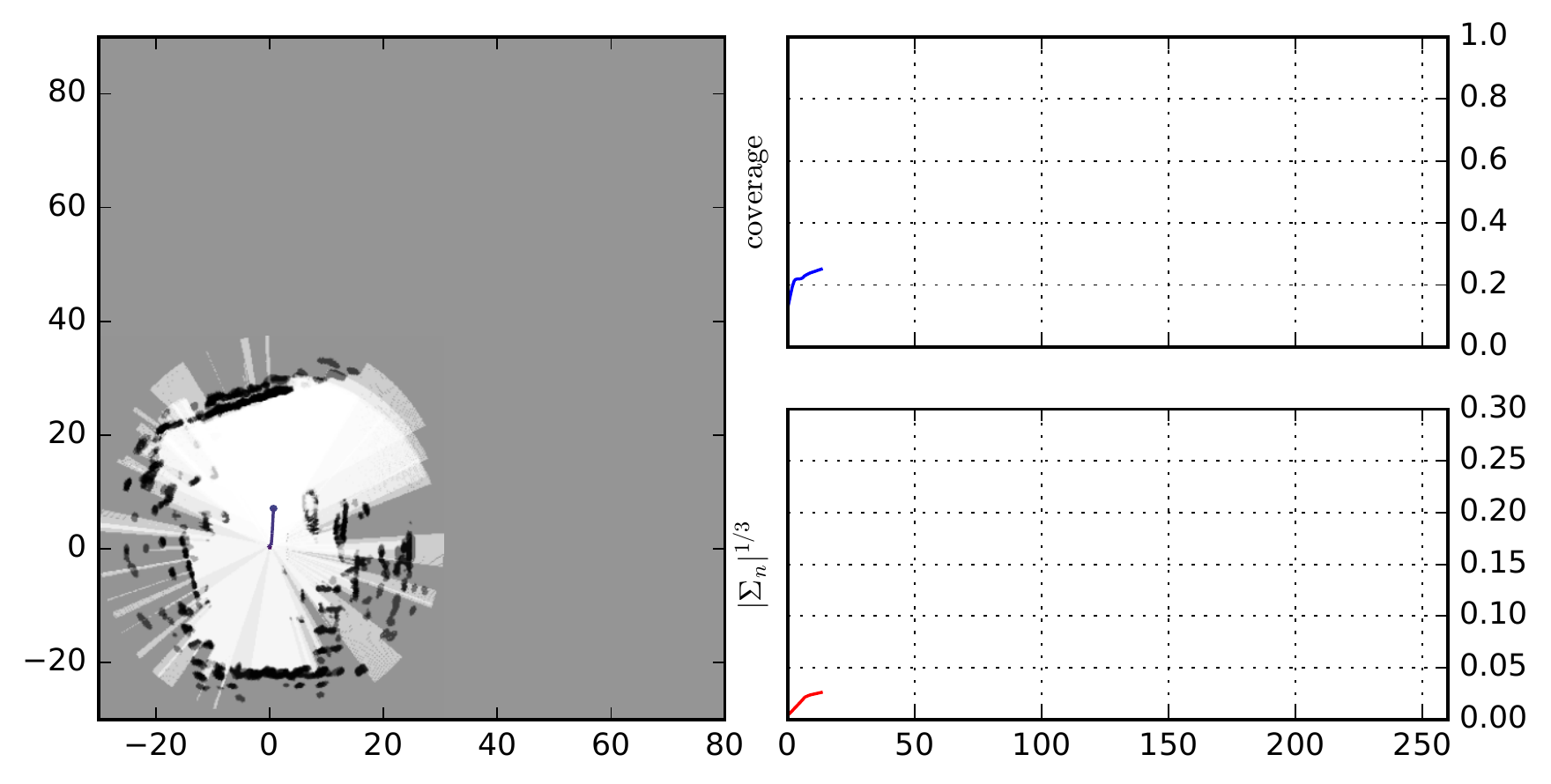}}\\
	\subfloat[Run 1: step 20]{\includegraphics[width=0.31\textwidth]{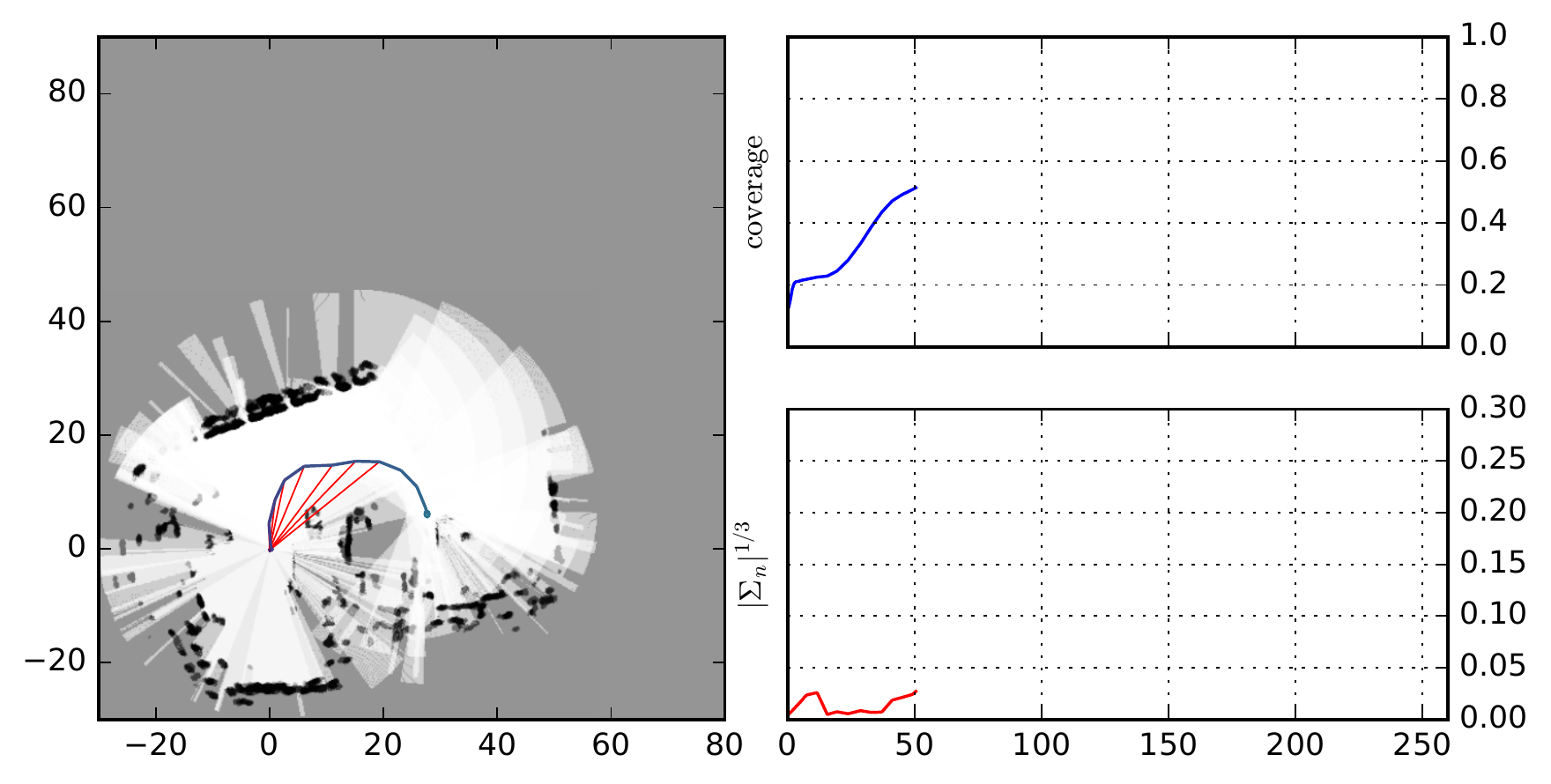}}
	\subfloat[Run 2: step 20]{\includegraphics[width=0.31\textwidth]{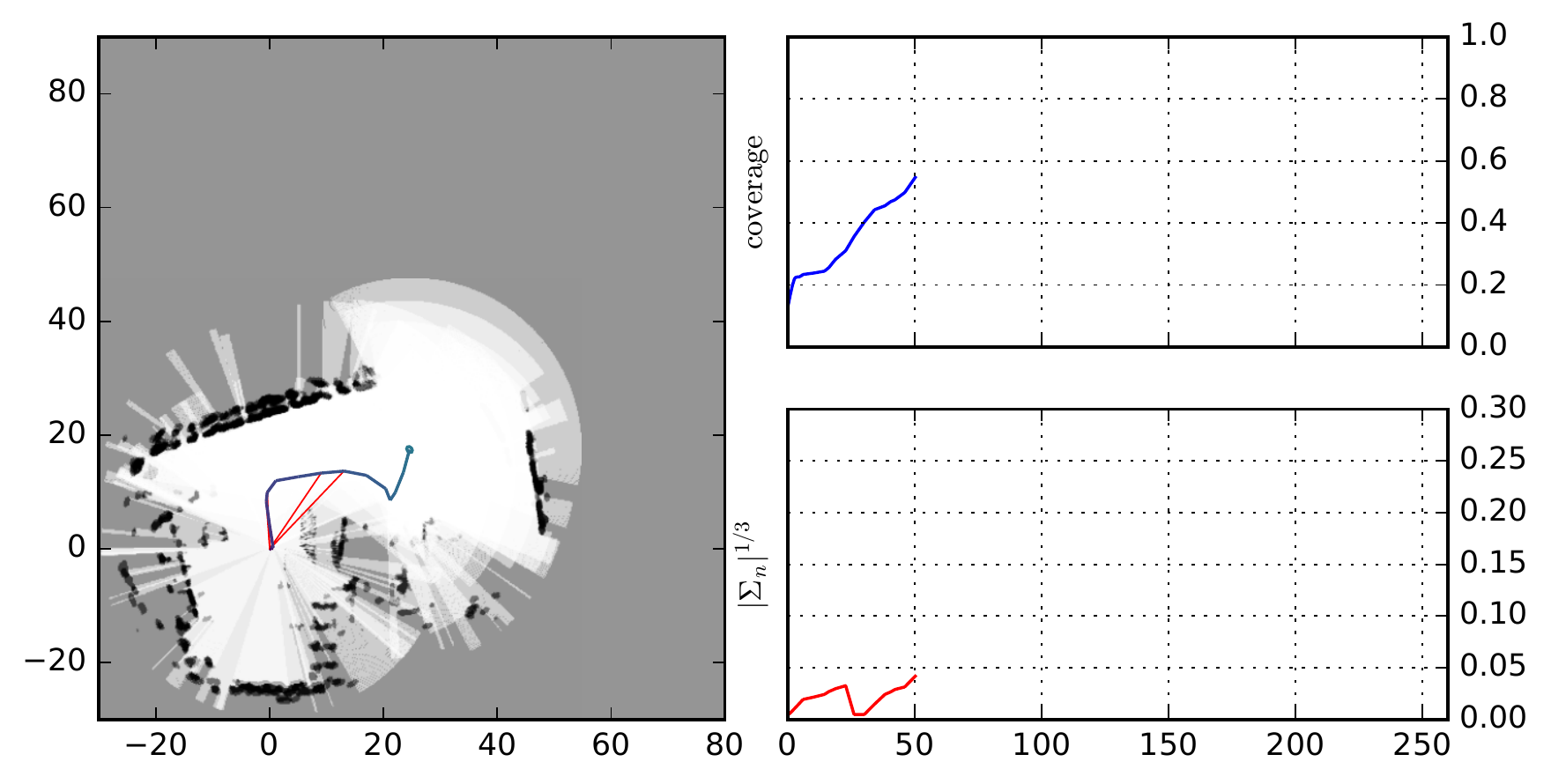}}
	\subfloat[Run 3: step 20]{\includegraphics[width=0.31\textwidth]{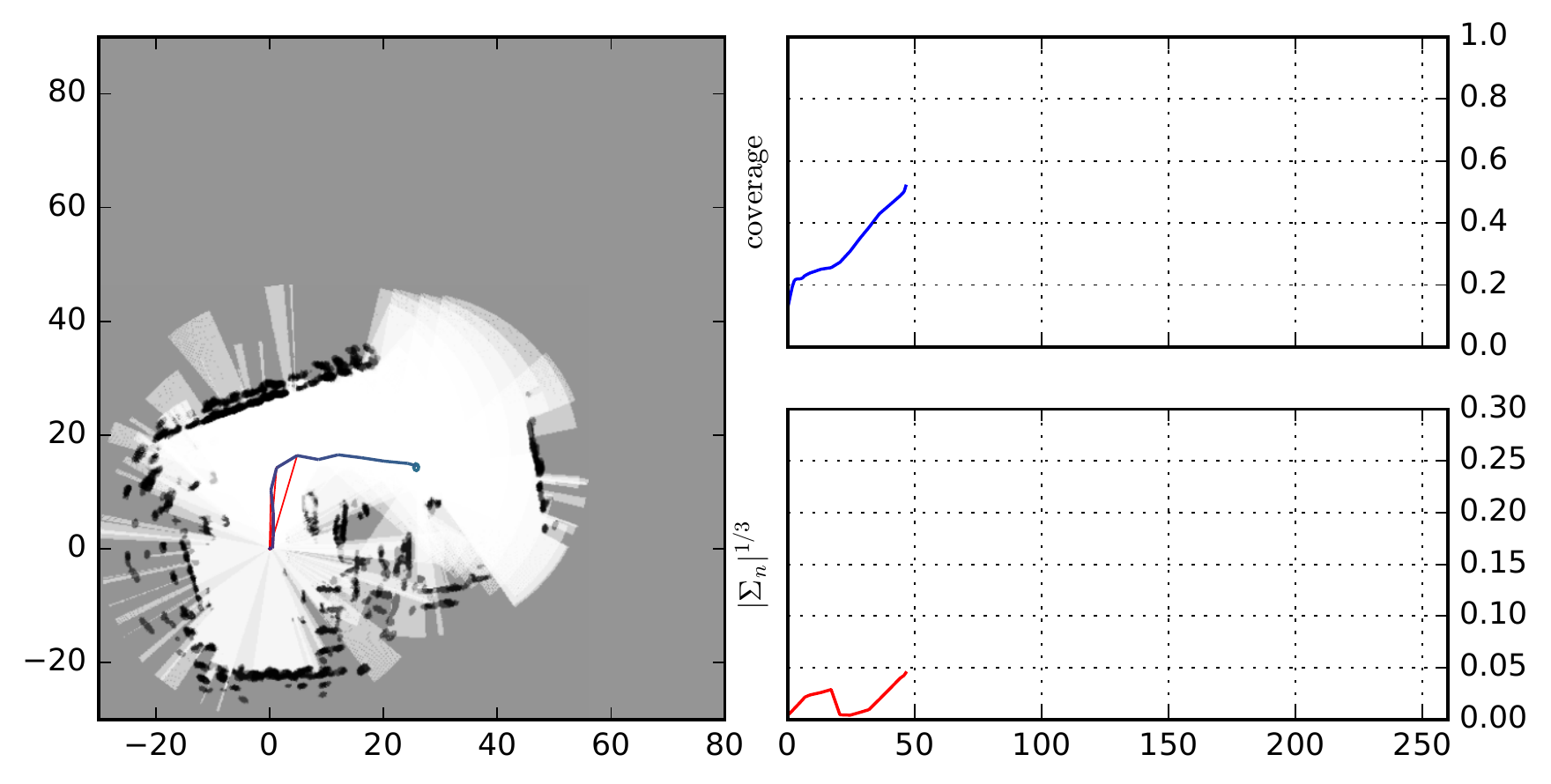}}\\
	\subfloat[Run 1: step 30]{\includegraphics[width=0.31\textwidth]{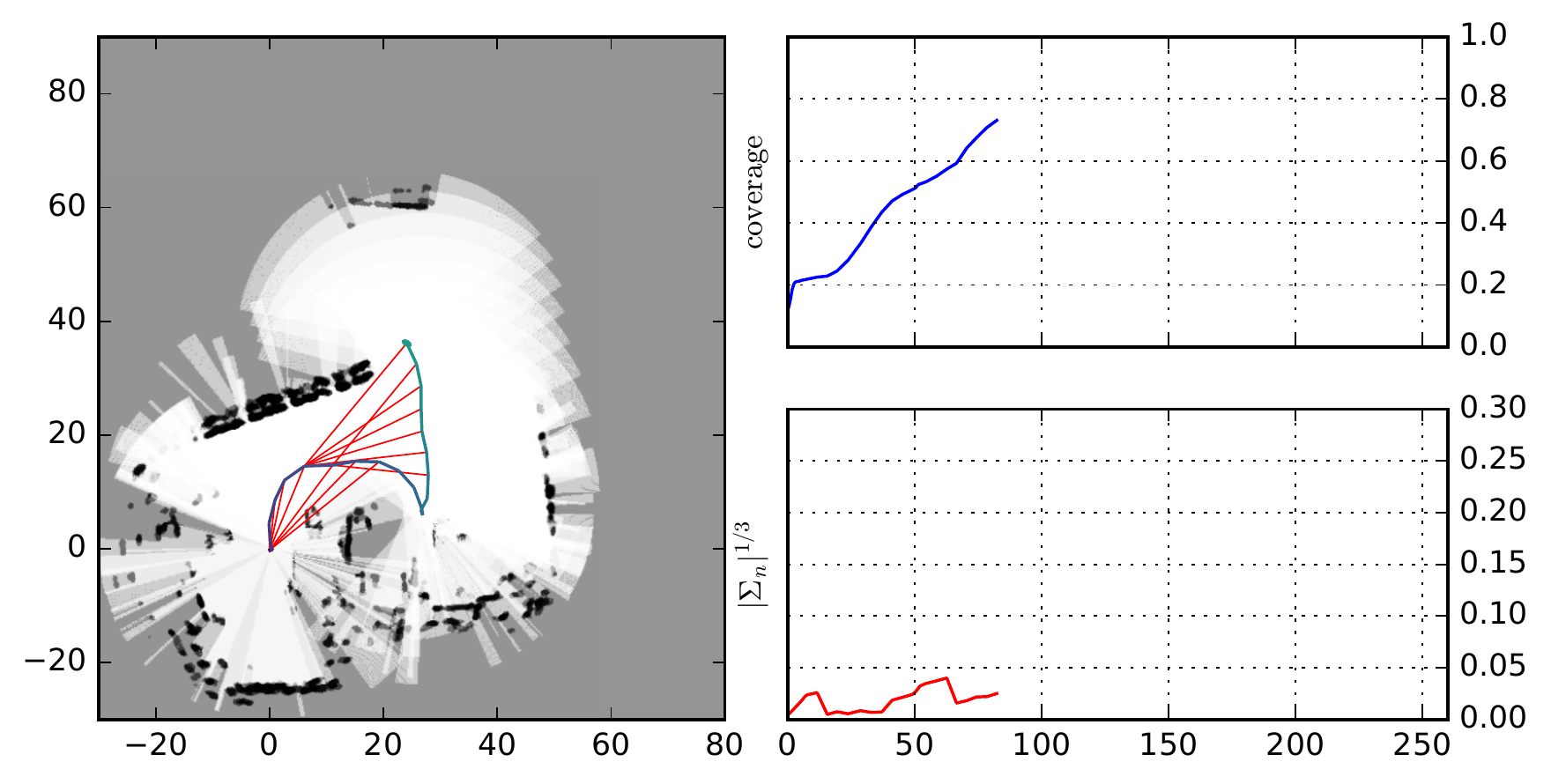}}
	\subfloat[Run 2: step 30]{\includegraphics[width=0.31\textwidth]{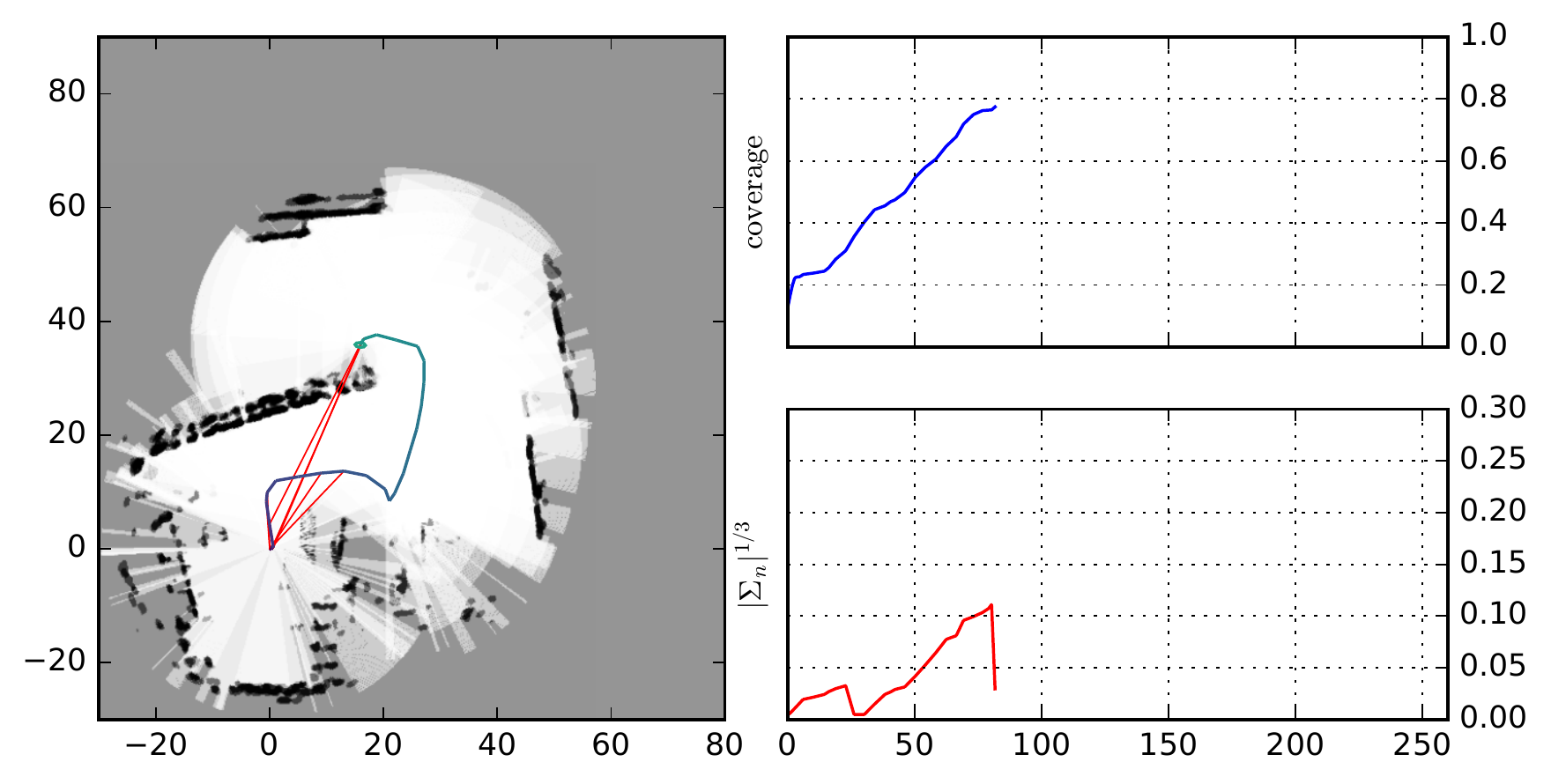}}
	\subfloat[Run 3: step 30]{\includegraphics[width=0.31\textwidth]{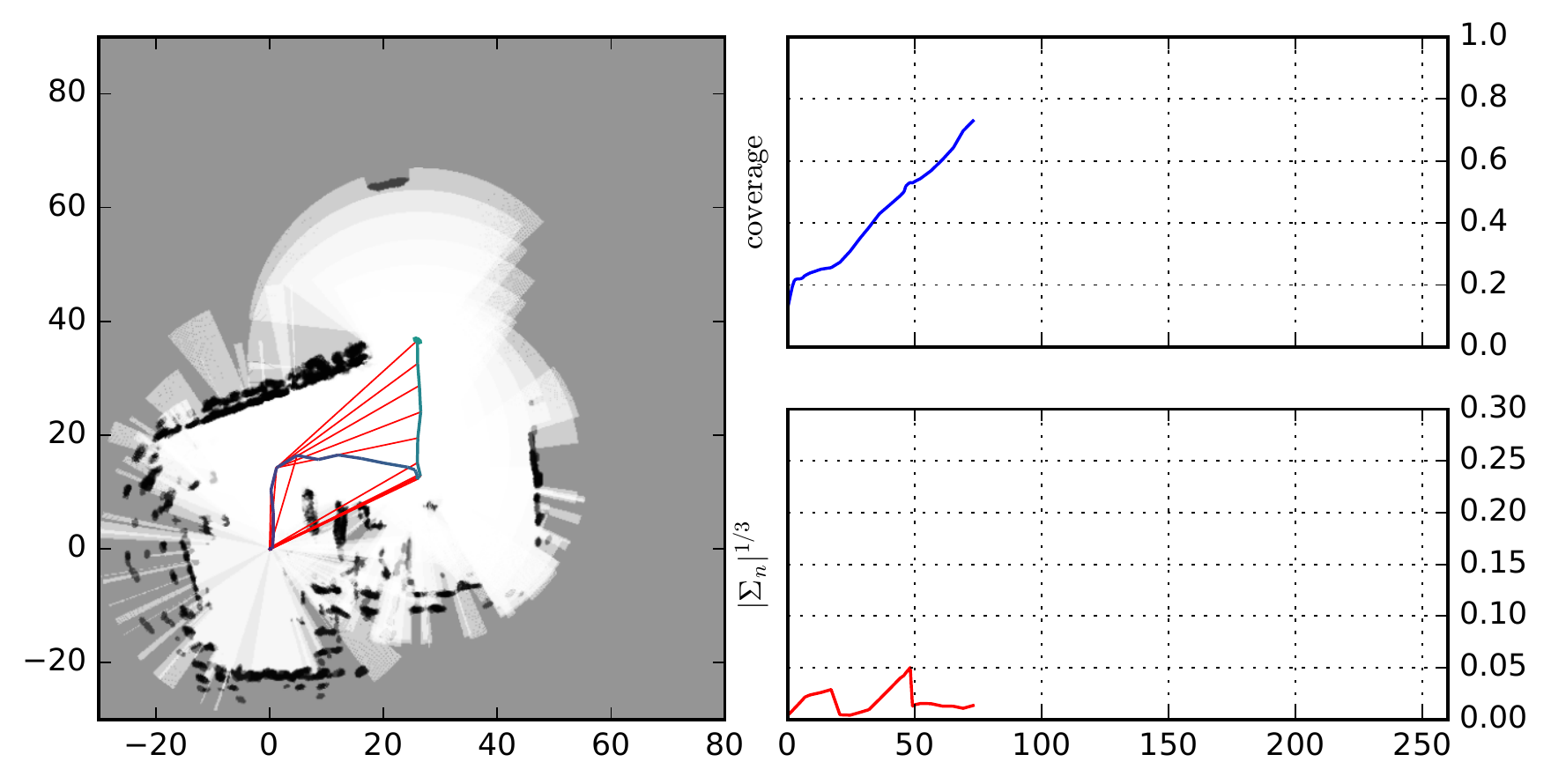}}\\
	\subfloat[Run 1: step 40]{\includegraphics[width=0.31\textwidth]{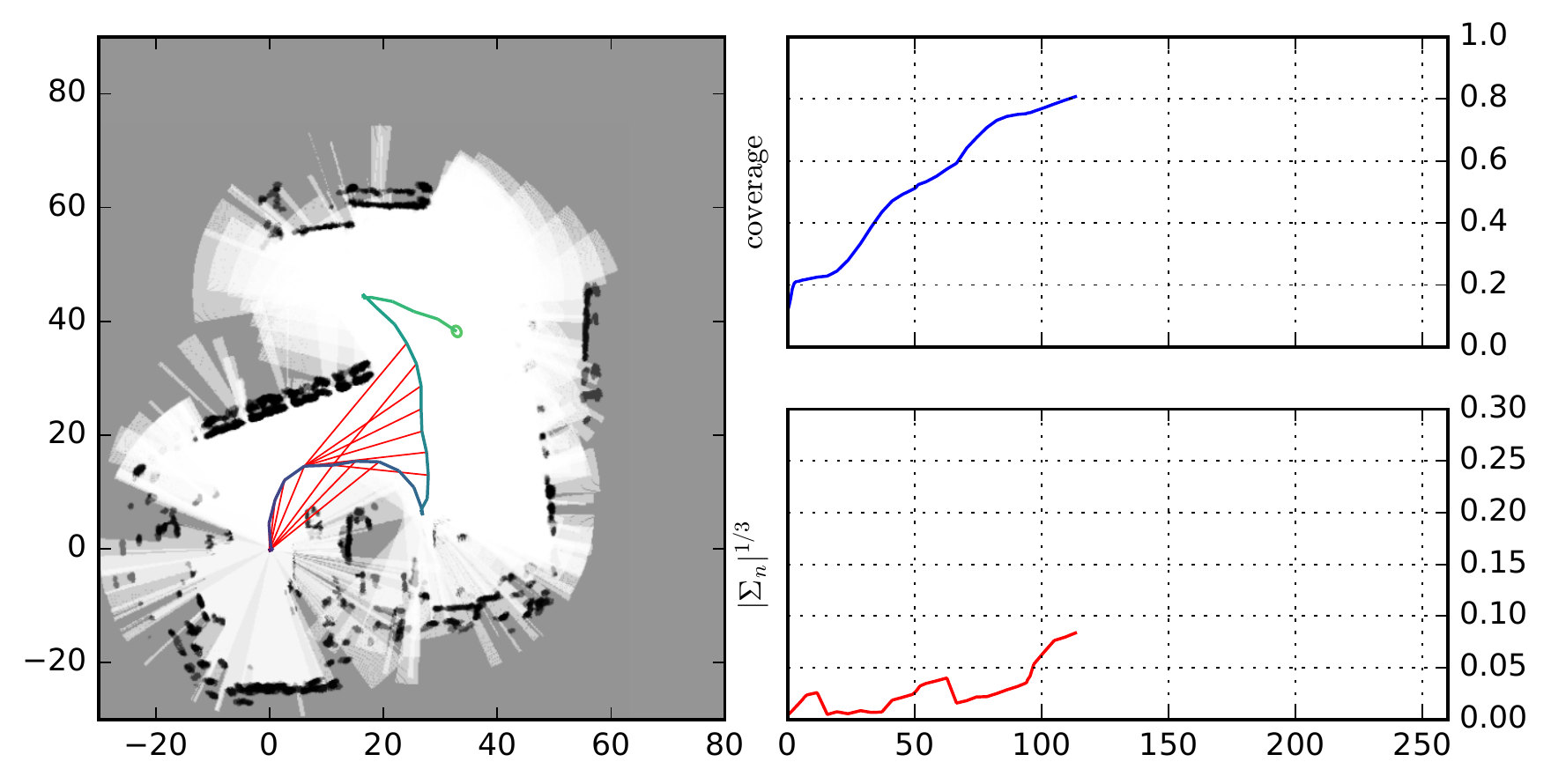}}
	\subfloat[Run 2: step 40]{\includegraphics[width=0.31\textwidth]{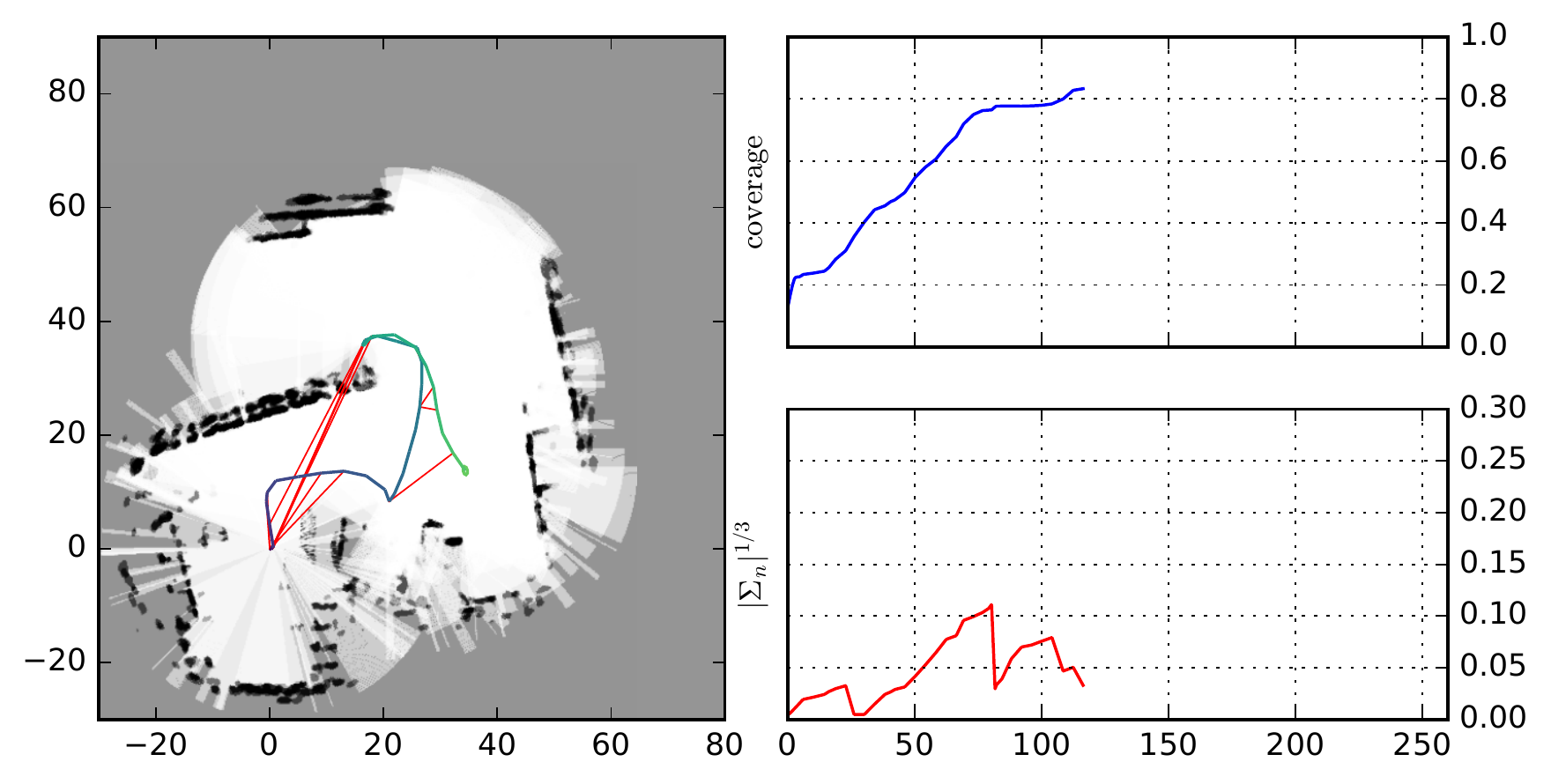}}
	\subfloat[Run 3: step 40]{\includegraphics[width=0.31\textwidth]{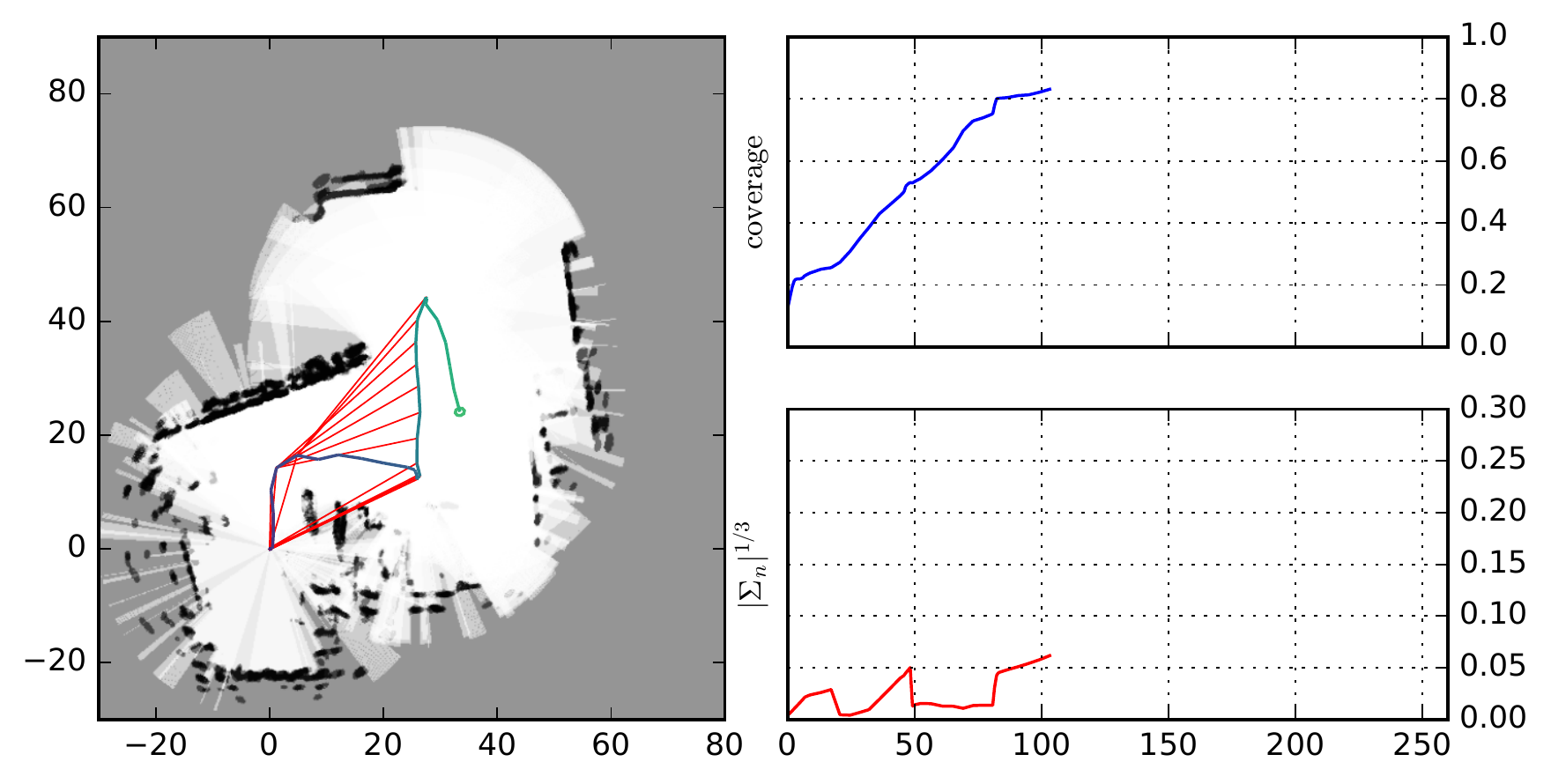}}\\
	\subfloat[Run 1: step 50]{\includegraphics[width=0.31\textwidth]{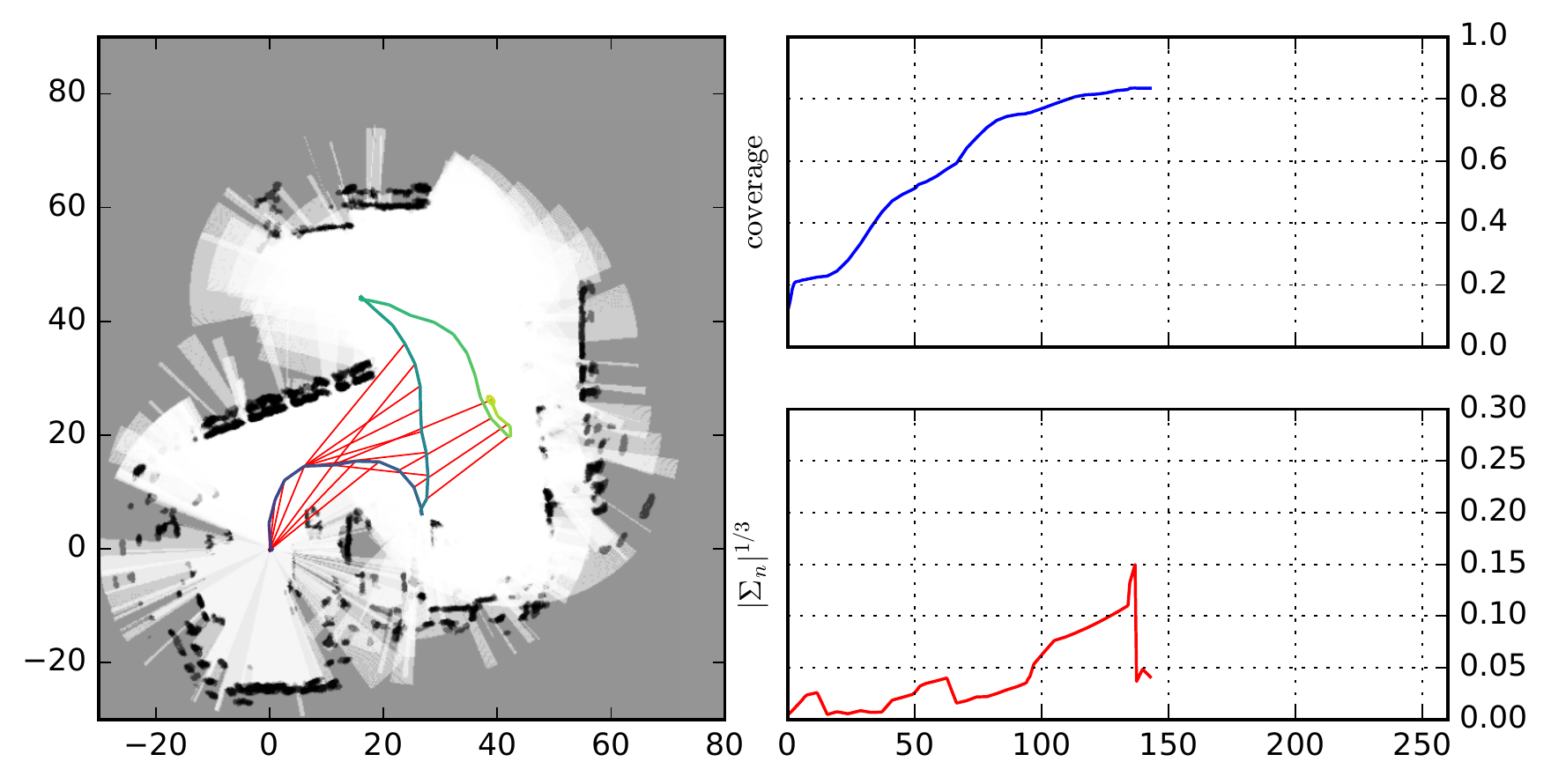}}
	\subfloat[Run 2: step 50]{\includegraphics[width=0.31\textwidth]{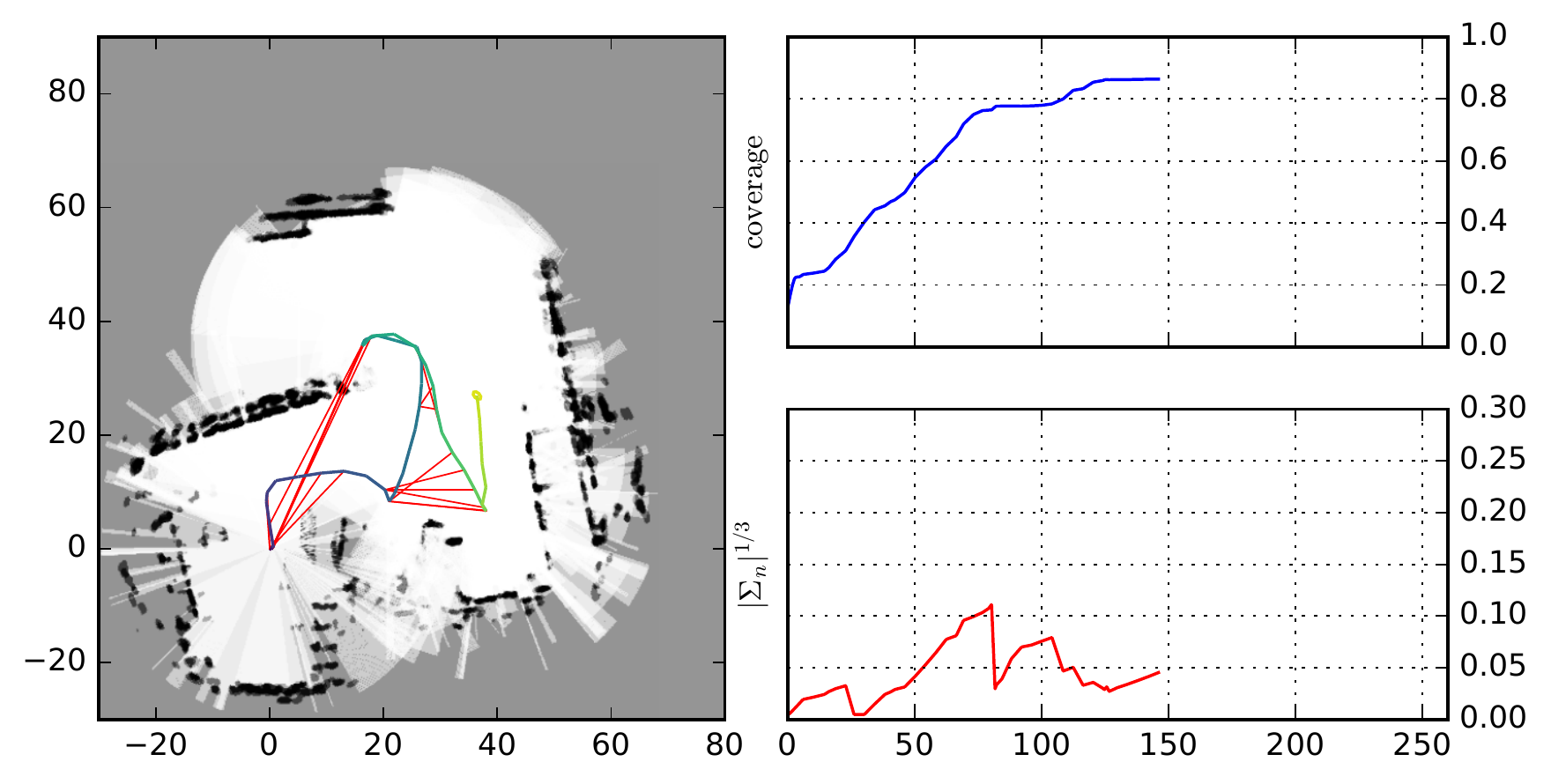}}
	\subfloat[Run 3: step 50]{\includegraphics[width=0.31\textwidth]{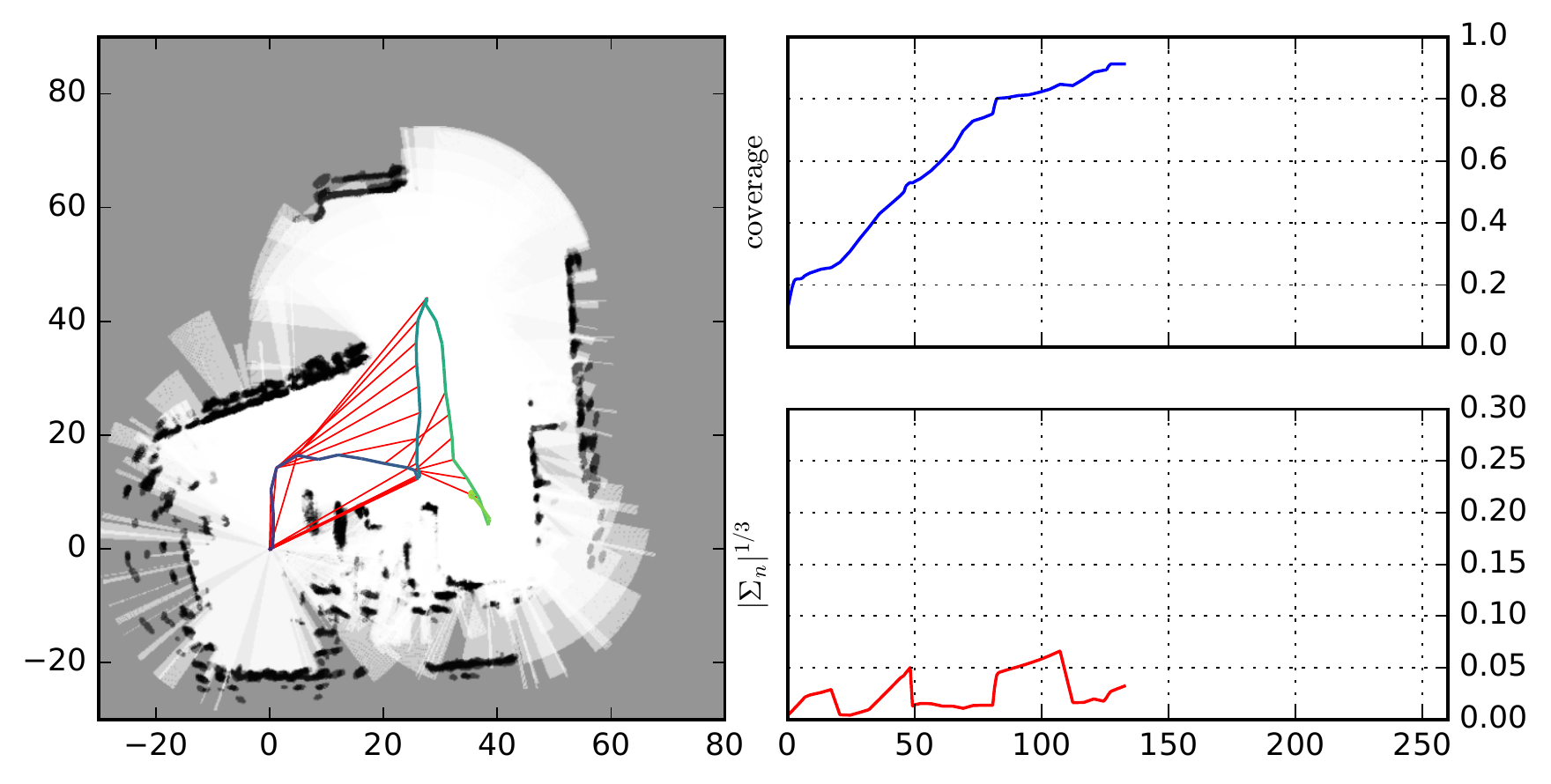}}\\
	\subfloat[Run 1: step 56]{\includegraphics[width=0.31\textwidth]{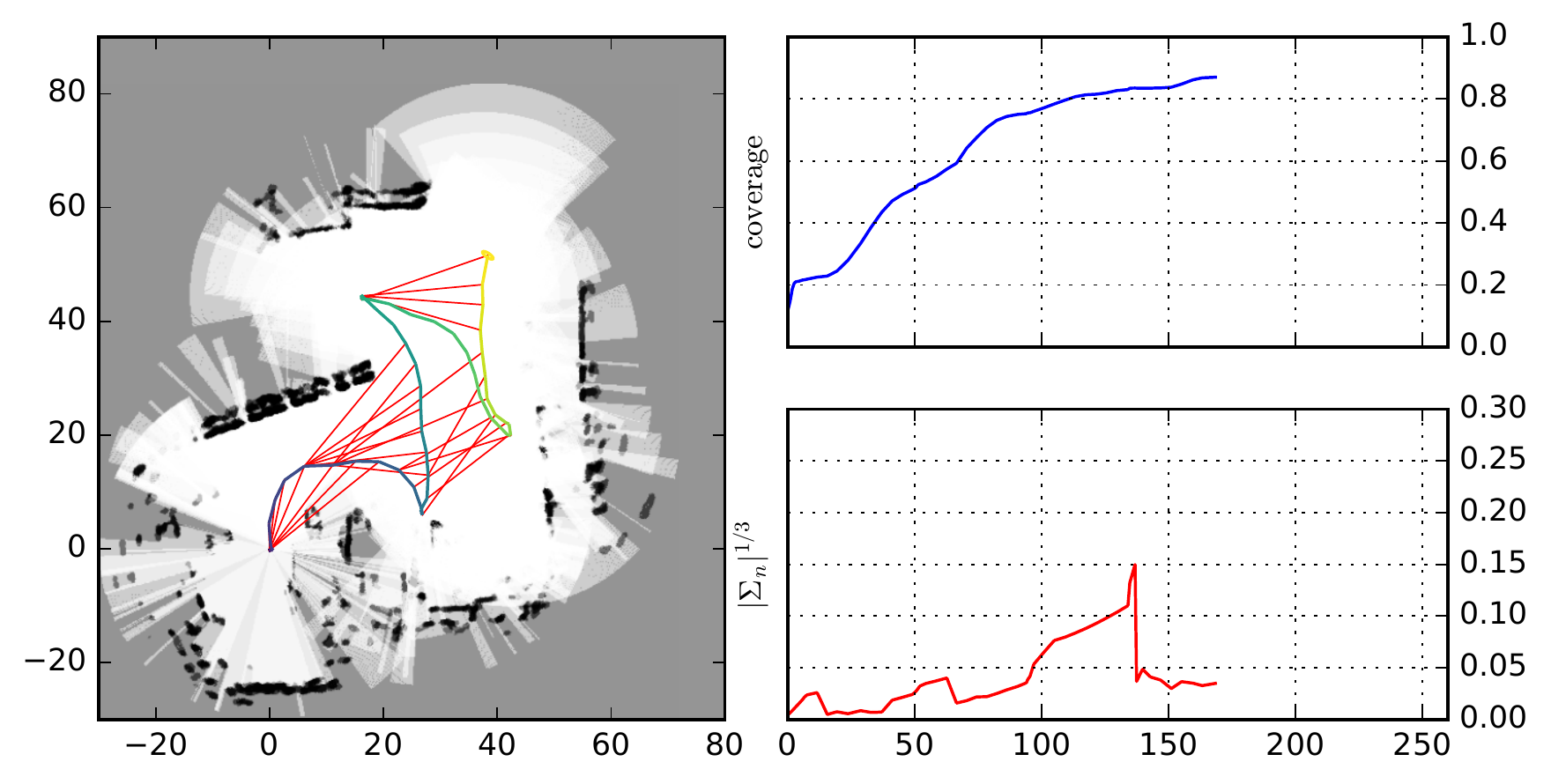}}
	\subfloat[Run 2: step 54]{\includegraphics[width=0.31\textwidth]{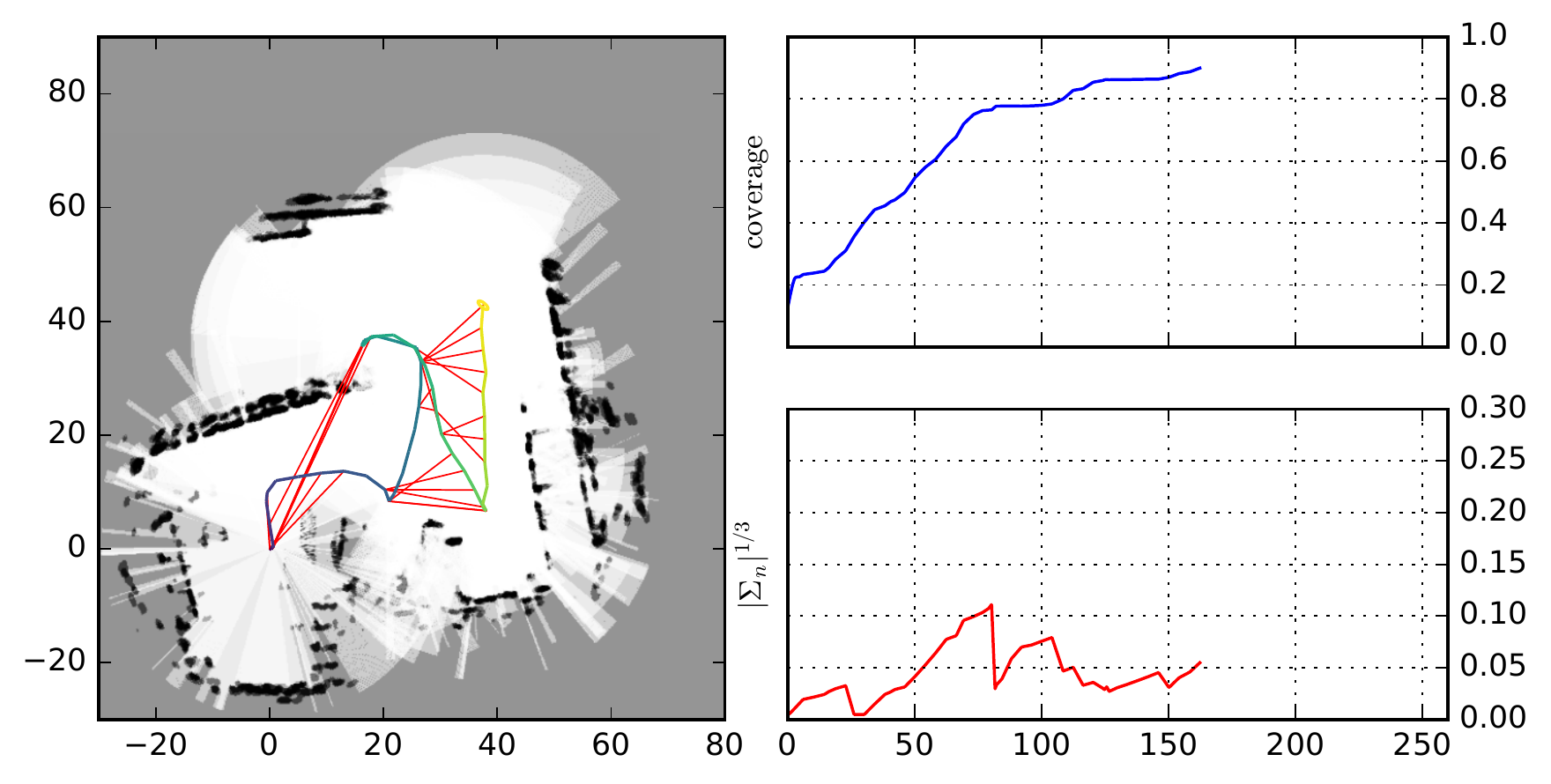}}
	\subfloat[Run 3: step 60]{\includegraphics[width=0.31\textwidth]{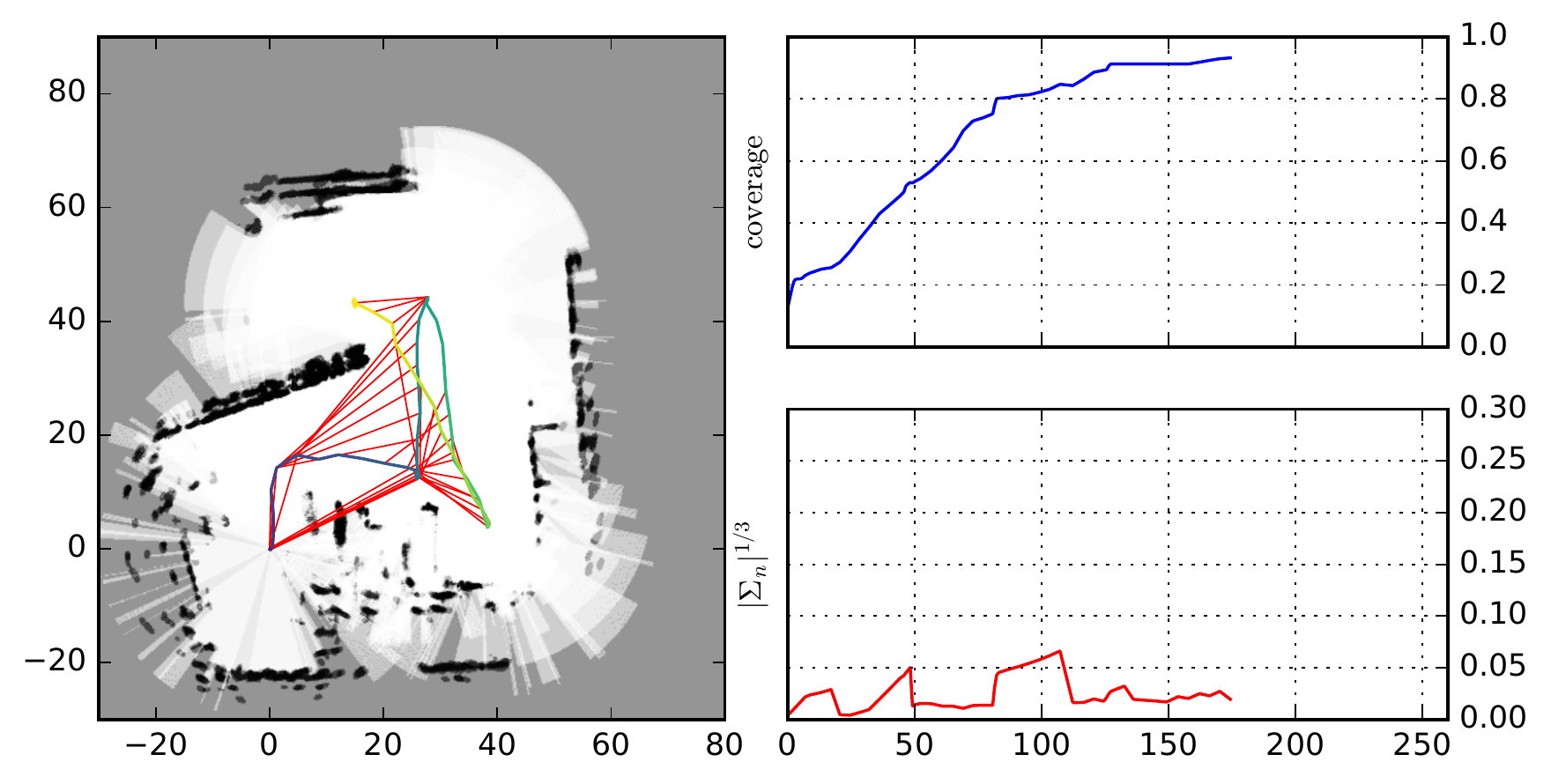}}
	\caption{Three runs of real-world robot exploration using the proposed EM algorithm in the environment depicted in Fig. \ref{fig:bruce-real-setup}. Each column shows the results from a single run, with each row showing a snapshot at a specific time-step. The final row shows the terminus of each run. Each subfigure shows the BlueROV's pose history and loop closure constraints over the occupancy map at left, with coverage and pose uncertainty shown at right, plotted against travel distance.}
	\label{fig:bruce-real-em}
\end{figure*}

\subsection{Exploration with a Real Underwater Robot in a Harbor Environment}

The hardware used to test this work is a modified BlueROV2, pictured in Fig. \ref{fig:bruce-real-setup}. This platform accommodates an Oculus M750d multi-beam imaging sonar, a Rowe SeaPilot DVL, a VectorNav VN100 IMU, and a Bar30 pressure sensor. Similar to our simulation, the sonar field of view is $r=[\SI{0}{m}, \SI{30}{m}]$ and $\theta=[\SI{-65}{\degree}, \SI{65}{\degree}]$, with a $\SI{20}{\degree}$ vertical aperture. Although the pitch angle of the sonar is adjustable, in the experiments to follow, it is set to $\SI{0}{\degree}$. The sonar is operated at \SI{750}{MHz} with 512 beams, \SI{4}{mm} range resolution and \SI{1}{\degree} angular resolution. The ROV is operated at a fixed depth of 1m throughout our experiments. Data is streamed via a tether to a surface laptop and processed in real-time on the laptop. The surface laptop is equipped with an Intel i7-4710 quad-core CPU, NVIDIA Quadro K1100M GPU, and 8GB of RAM.  All components of the software stack are handled by the surface computer, including image processing, the SLAM solution, belief propagation, motion planning, and issuing control inputs to the vehicle. The Robot Operating System (ROS) navigation stack is used for motion planning and control, with a customized global planner as described in Sec. \ref{sec:motion-planning}, and the timed elastic band local planner \cite{Rosman2017}. The GTSAM \cite{gtsam} implementation of iSAM2 \cite{Kaess2012} is used for pose graph optimization in our SLAM back end. No human intervention was required during the experiments, apart from reeling out and retracting tether cable as needed. Examples of the motion planning process during the ROV's exploration of an unknown environment are given in Figure \ref{fig:bruce-real-planning}.
%This is shown in Figure \ref{fig:bruce-real-planning} in green, a graph is constructed over the environment with A* used to generate candidate trajectories to candidate goals.  

 Real world experiments were carried out in a marina at the United States Merchant Marine Academy (USMMA). An overview of the experiment setup is shown in Fig. \ref{fig:bruce-real-setup}. Prior to launching the vehicle, it is manually commanded to a starting location aligned with the axes shown in the figure (with the sonar field of view centered along the red axis). In Fig. \ref{fig:bruce-real-setup}, the estimated map (represented by a time-colored point cloud), the robot's estimated trajectory,
 and loop closure measurements are visualized, overlaid with satellite imagery. We note that the satellite imagery is manually aligned with these results, serving only as a qualitative indicator of the discrepancy between ground truth and the estimated map. The goal is to explore and map the portions of the environment that lie within a bounding box of dimensions 130 m × 60 m. Three runs were performed for each of the four competing algorithms, and results are shown in Figs. \ref{fig:bruce-real-others} and \ref{fig:bruce-real-em}.
 Due to the lack of ground truth information and a limited number of repeated experiments (due to the length of each experiment and limited availability of the facility), exploration performance, evaluated with respect to map coverage and pose uncertainty plotted against traveled distance, is presented to better understand the workings of the four competing algorithms. Four intermittent steps and the final outcomes of exploration are plotted in Fig. \ref{fig:bruce-real-em}, and to save space, only the final state of exploration is plotted for the NF, NBV and heuristic algorithms in Fig. \ref{fig:bruce-real-others}. As a supplementary resource, all twelve trials can be viewed side-by-side in their entirety in our video attachment. 
 
 From Fig. \ref{fig:bruce-real-others}, it is evident that both the NF and NBV algorithms take actions that favor exploration, resulting in trajectories that drive toward the top-right corner of the map quickly. As a result, loop closures are added later than with the heuristic or EM algorithms, and the loop closures achieved are not intentional. The quality of the maps produced by NBV is the worst among the four algorithms, and although the final maps produced using NF appear accurate, the large deviations from this final estimate, and the levels of uncertainty incurred during exploration, are not reduced until approaching the completion of each session. 
 
 The heuristic algorithm similarly incurs growing uncertainty as it explores, although it does not reach the same level as in the NF and NBV algorithms, and uncertainty is reduced through the intentional selection of loop closures. This can be observed most clearly in run 1 of Figure \ref{fig:bruce-real-others}, where at distance 100 the robot intentionally closes the loop to reduce estimation uncertainty.
 %By contrast, the uncertainty-aware exploration algorithms, EM and heuristic, achieved loop closure constraints. 
 Different from the heuristic approach, which seeks loop closures only at highly uncertain states, the EM approach continuously takes map uncertainty into account, and loop closure constraints are intertwined throughout the entire trajectory in all three trials. %\Comment{JM: Maybe this is what he means above, EM has a dense network of loop closures throughout the entire trajectory rather then just one big one at the end.}
 We can see from the comparison that EM also maintains low uncertainty throughout all three trials, while the map coverage rate is competitive with other approaches. 

\section{Conclusion and Future Work}
\label{sect:conclusion}

In this paper we have presented and evaluated the concept of using \textit{virtual maps}, comprised of uniformly discretized virtual landmarks, to support the efficient autonomous exploration of unknown environments while managing state estimation and map uncertainty. Our proposed EM exploration algorithm, when used in conjunction with a sonar keyframe-based pose SLAM architecture, equips an underwater robot with the ability to autonomously explore a cluttered underwater environment while maintaining an accurate map, pose estimate, and trajectory estimate, in addition to achieving a time-efficient coverage rate. The advantages of the proposed framework are demonstrated in simulation, and a qualitative proof of concept is presented over several real-world underwater exploration experiments. These experiments represent, to our knowledge, the first instance of autonomous exploration of a real-world, obstacle-filled outdoor environment in which an underwater robot (a tethered BlueROV we employ as a proof-of-concept system) directly incorporates its SLAM process and predictions based on that process into its decision-making. Future work entails the expansion of our framework to support three-dimensional sonar-based occupancy mapping.  

%The first instance, to our knowledge, of autonomous exploration of a real-world, obstacle-filled outdoor environment, in which an underwater robot (a tethered BlueROV we employ as a proof-of-concept system) directly incorporates its SLAM process and predictions based on that process into its decision-making.

\section*{Acknowledgments}
We would like to thank Arnaud Croux, Timothy Osedach, Sepand Ossia and Stephane Vannuffelen of Schlumberger-Doll Research Center for guidance and support that was instrumental to the success of the field experiments described in this paper. We also thank Prof. Carolyn Hunter of the U.S. Merchant Marine Academy for facilitating access to the marina used in the experiments.

\ifCLASSOPTIONcaptionsoff
  \newpage
\fi



%\bibliographystyle{IEEEtran}

%\begin{thebibliography}{99}
%
%\bibitem{a}
\bibliographystyle{IEEEtran}
\bibliography{refs}
%
%\end{thebibliography}

% \subsection{Proof of Proposition 1}

%\begin{thebibliography}{1}
%\end{thebibliography}



 \begin{IEEEbiography}[{\includegraphics[width=1in,height=1.25in,clip,keepaspectratio]{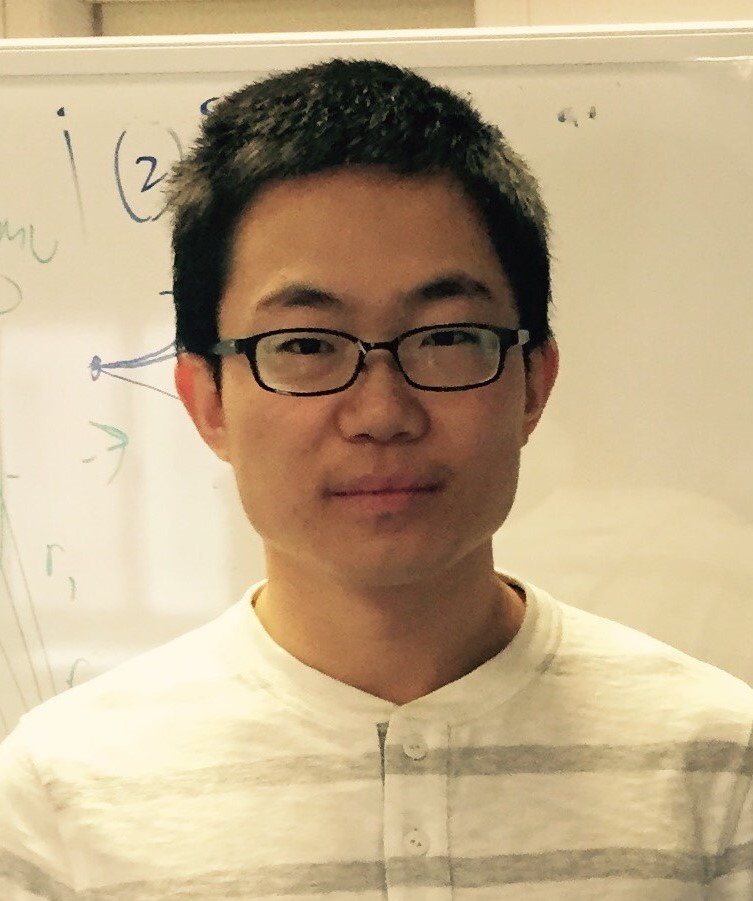}}]{Jinkun Wang}
 received a Bachelor of Science in Mechanical Engineering from the University of Science and Technology of China, Hefei, China, in 2014, and M.Eng. and Ph.D. degrees in Mechanical Engineering from Stevens Institute of Technology, Hoboken, NJ, USA, in 2016 and 2020, respectively. He is currently an Applied Scientist at Amazon Robotics in Louisville, CO, USA.
 \end{IEEEbiography}

 \begin{IEEEbiography}[{\includegraphics[width=1in,height=1.25in,clip,keepaspectratio]{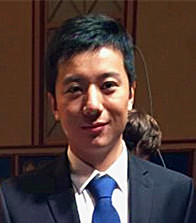}}]{Fanfei Chen}
 received a Bachelor's Degree in Mechanical Design, Manufacturing and Automation from Tongji Zhejiang College, Jiaxing, China, in 2014, and M.Eng. and Ph.D. degrees in Mechanical Engineering from Stevens Institute of Technology, Hoboken, NJ, USA, in 2016 and 2021, respectively. He is currently a Senior Robotics Engineer at Exyn Technologies in Philadelphia, PA, USA.
 %a Master of Engineering in Mechanical Engineering from Stevens Institute of Technology, Hoboken, NJ, USA, in 2015. He is currently working toward the Ph.D. degree with the Department of Mechanical Engineering, Stevens Institute of Technology, NJ, USA.
 \end{IEEEbiography}

 \begin{IEEEbiography}[{\includegraphics[width=1in,height=1.25in,clip,keepaspectratio]{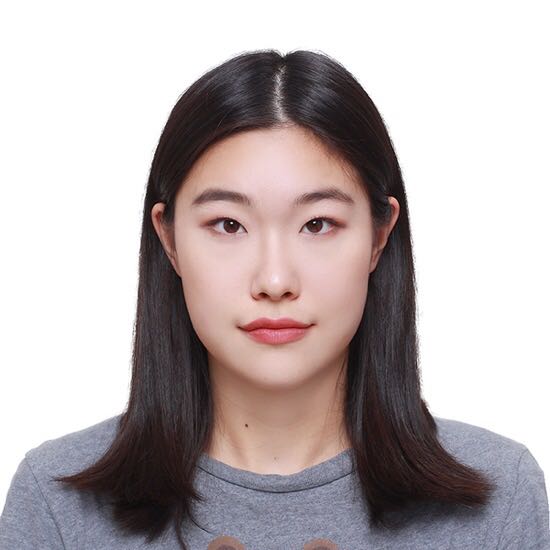}}]{Yewei Huang}
 received a Bachelor's Degree in Geographic Information Systems and a Master's Degree in Surveying and Mapping from Tongji University, Shanghai, China, in 2016 and 2019, respectively. She is currently working toward the Ph.D. degree with the Department of Mechanical Engineering, Stevens Institute of Technology, Hoboken, NJ, USA.
 \end{IEEEbiography}

 \begin{IEEEbiography}[{\includegraphics[width=1in,height=1.25in,clip,keepaspectratio]{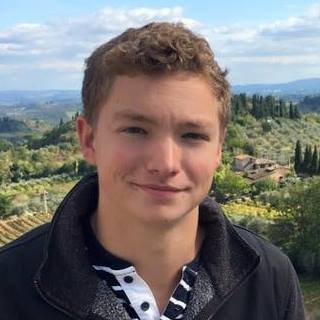}}]{John McConnell}
 received a Bachelor's Degree in Mechanical Engineering from SUNY Maritime College, NY, USA, in 2015, and a Master of Engineering in Mechanical Engineering from Stevens Institute of Technology, Hoboken, NJ, USA, in 2020. He is currently working toward the Ph.D. degree with the Department of Mechanical Engineering, Stevens Institute of Technology, Hoboken, NJ, USA. Prior to joining Stevens, he worked in engineering roles with the American Bureau of Shipping, ExxonMobil, and Duro UAS.
 \end{IEEEbiography}

 \begin{IEEEbiography}[{\includegraphics[width=1in,height=1.25in,clip,keepaspectratio]{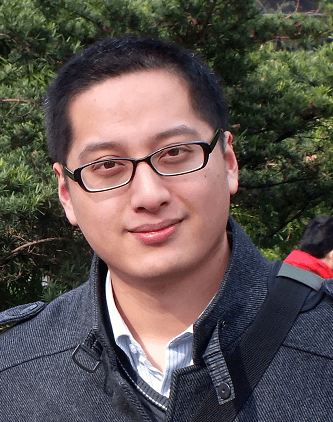}}]{Tixiao Shan}
 received a Bachelor of Science in Mechanical Engineering and Automation from Qingdao University, Qingdao, China, in 2011, a Master's Degree in Mechatronic Engineering from Shanghai University, Shanghai, China, in 2014, and a Ph.D. in Mechanical Engineering from Stevens Institute of Technology, Hoboken, NJ, USA, in 2019. During 2019 to 2021, he was a Postdoctoral Associate %affiliated jointly with the Computer Science and Artificial Intelligence Laboratory (CSAIL) and the Department of Urban Studies and Planning 
 at Massachusetts Institute of Technology, Cambridge, MA, USA. He is currently a Computer Scientist at SRI International, Princeton, NJ, USA.
 \end{IEEEbiography}

%\newpage

 \begin{IEEEbiography}[{\includegraphics[width=1in,height=1.25in,clip,keepaspectratio]{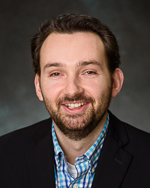}}]{Brendan Englot}
 (S'11--M'13--SM'20) received S.B., S.M., and Ph.D. degrees in Mechanical Engineering from Massachusetts Institute of Technology, Cambridge, MA, USA, in 2007, 2009, and 2012, respectively.

 He was a research scientist with United Technologies Research Center, East Hartford, CT, USA, from 2012 to 2014. He is currently an Associate Professor with the Department of Mechanical Engineering, Stevens Institute of Technology, Hoboken, NJ, USA. His research interests include motion planning, localization, and mapping for mobile robots, learning-aided autonomous navigation, and marine robotics. He is the recipient of a 2017 National Science Foundation CAREER award and a 2020 Office of Naval Research Young Investigator Award. 
 \end{IEEEbiography}


\vfill


\end{document}